\documentclass{article}
\PassOptionsToPackage{table}{xcolor}
\usepackage{iclr2027_conference,times}
\setcitestyle{authoryear,round,citesep={;},aysep={,},yysep={,}}

\usepackage[utf8]{inputenc}
\usepackage[T1]{fontenc}
\usepackage{newtxtext}
\usepackage{mathtools}
\usepackage{hyperref}
\usepackage{url}
\usepackage{longtable}
\usepackage{wrapfig}
\usepackage{capt-of}
\usepackage{amsfonts}
\usepackage{nicefrac}
\usepackage{xcolor}
\usepackage{booktabs}
\usepackage{tikz}
\usetikzlibrary{positioning}

\tikzset{
box/.style={
    draw,
    rounded corners,
    align=center,
    minimum width=2.8cm,
    minimum height=0.9cm
},
arrow/.style={->,thick}
}

\usepackage{amsmath, amssymb}
\usepackage{algorithm}
\usepackage{algpseudocode}
\usepackage{graphicx}
\usepackage{float}
\usepackage{enumitem}

\usepackage{amsthm}
\newtheorem{assumption}{Assumption}

\newtheorem{proposition}{Proposition}
\newtheorem{lemma}{Lemma}

\newtheorem{theorem}{Theorem}

\usepackage{multirow}
\usepackage{titletoc}
\usepackage[most]{tcolorbox}

\newtcolorbox{leftrulebox}{
 colback=gray!12,
 colframe=gray!50,
 boxrule=0.35pt,
 arc=2pt,
 left=4pt,right=4pt,top=2pt,bottom=2pt,
 boxsep=1pt
}

\newcommand{\Oc}{\mathcal{O}}
\newcommand{\bx}{\mathbf{x}}
\newcommand{\by}{\mathbf{y}}
\newcommand{\bw}{\mathbf{w}}
\newcommand{\bz}{\mathbf{z}}
\newcommand{\bv}{\mathbf{v}}
\newcommand{\X}{\mathcal{X}}
\newcommand{\FPF}{{\mathcal{F}_{\rm PF}}}
\newcommand{\f}{\mathbf{f}}
\newcommand{\fPF}{{\mathbf{f}_{\rm PF}}}
\DeclareMathOperator*{\argmin}{arg min}

\title{SURFACE: Preference Distribution Matching on Pareto Fronts via Scalarization Steering}

\author{%
 Liuyuan Jiang$^{1}$,\quad Prasanna Sattigeri$^{2}$,\quad Lisha Chen$^{1}$ \\
 $^{1}$University of Rochester \quad
 $^{2}$IBM Research
}

\makeatletter
\def\@maketitle{\vbox{\hsize\textwidth
{\LARGE\sc \@title\par}
    \lhead{Preprint}%
    \def\And{\end{tabular}\hfil\linebreak[0]\hfil
            \begin{tabular}[t]{l}\bf\rule{\z@}{24pt}\ignorespaces}%
  \def\AND{\end{tabular}\hfil\linebreak[4]\hfil
            \begin{tabular}[t]{l}\bf\rule{\z@}{24pt}\ignorespaces}%
    \begin{tabular}[t]{l}\bf\rule{\z@}{24pt}\@author\end{tabular}%
\vskip 0.3in minus 0.1in}}
\makeatother
\ClearShipoutPicture

\begin{document}
\maketitle

\begin{abstract}
Constructing a finite Pareto representation requires allocating a limited number of solutions across the Pareto front (PF). A desired preference distribution specifies this allocation by assigning more solutions to some PF regions and fewer to others. 
Scalarization generates Pareto solutions by varying objective weights and solving single-objective problems. However, a distribution of scalarization weights generally does not translate into the same distribution of Pareto solutions along the PF, because the weight-to-solution map is nonlinear and unknown in advance.
To account for this distortion, we introduce SURFACE (\emph{\textbf{S}calarization \textbf{U}pdates for \textbf{R}edistributing \textbf{F}inite \textbf{A}pproximations with \textbf{C}hart \textbf{E}stimation}), which adaptively updates scalarization weights to match a desired PF distribution. SURFACE estimates how each region of the weight grid corresponds to an area of the PF from the current scalarized solutions, then redistributes the weights accordingly. The method supports parallel, warm-started, inexact optimization with standard single-objective solvers.
Theoretically, we establish that, under the stated regularity conditions, SURFACE achieves linear convergence of the preference distribution mismatch to a finite-resolution error floor of $\mathcal{O}(N^{-2/(M-1)})$, where $N$ is the number of adaptive weights and $M$ is the number of objectives.
Empirically, experiments across multiple benchmarks demonstrate improved target-mass balance and PF coverage compared with established baselines.
\end{abstract}

\section{Introduction}\label{sec:introduction}

\vspace{-0.1cm}
Multi-objective optimization (MOO) is increasingly important in modern artificial intelligence (AI), including multimodal learning~\citep{wei2024innocent,saha2025f3ocus}, large language model (LLM) alignment~\citep{shi2024decoding}, and agentic AI systems~\citep{zhang2026evoroute}.
Let $\bx$ be a decision variable in the feasible domain $\X\subseteq\mathbb R^n$. MOO minimizes a vector of objective functions $f_m:\X\to\mathbb R$:
\begin{align}
\min_{\bx\in\X} \f(\bx):=\left[ f_1(\bx),f_2(\bx),\ldots,f_M(\bx)\right]^\top.
\label{eq:moo_problem}
\end{align}

\vspace{-0.2cm}
\noindent
In MOO, there is generally no single solution that minimizes all objectives simultaneously.
Instead, the goal of~\eqref{eq:moo_problem} is to find a \emph{Pareto-optimal} solution $\bx^\star\in\X$, for which no $\bx\in\X$
satisfies $f_m(\bx)\le f_m(\bx^\star)$ for all $m$, with strict inequality for at least one $m$.
\label{def:po_pf}
The collection of such solutions is called the \emph{Pareto set} $\X_{\rm PF}$. Its image $\FPF:=\f(\X_{\rm PF})$ forms the \emph{Pareto front} (PF)~\citep{miettinen1999nonlinear}.

Preference information is essential in MOO because Pareto optimality alone does not determine which tradeoffs matter most to a decision maker~\citep{pajasmaa2025group}. Under standard regularity conditions, the PF forms an $(M{-}1)$-dimensional manifold of tradeoff solutions~\citep{hillermeier2001generalized}. This motivates two classical paradigms. \emph{A priori} methods incorporate preferences before optimization to target preferred Pareto solutions~\citep{chen2024ferero,chen2025efficient}, but require preferences to be specified in advance. \emph{A posteriori} methods instead construct a representative PF approximation for later selection~\citep{zhang2024umod,boehmer2026picking}, but must represent a continuous PF with finitely many solutions.
This raises a fundamental question of how to allocate these solutions across the PF so that the resulting approximation reflects the desired preference distribution:
\begin{center}
\vspace{-0.1cm}
\emph{(Q1) How to generate Pareto solutions whose distribution matches a desired preference on the PF?}
\vspace{-0.1cm}
\end{center}
To specify the desired distribution of Pareto solutions, we prescribe a positive (unnormalized) \emph{preference profile} $\vartheta:\f(\X)\to(0,\infty)$ over attainable objective values~\citep{wang2017preference,chen2024ferero}. On the PF $\FPF\subseteq\f(\X)$, larger values of $\vartheta$ indicate regions where more Pareto solutions are desired, with $\vartheta\equiv1$ corresponding to uniform surface-area coverage.
Existing methods seek Pareto stationarity, specified tradeoffs, or well-spread representations, but offer limited control over aligning a finite Pareto representation with $\vartheta$~\citep{desideri2012multiple,lin2019pareto,pereyra2013equispaced,zhang2024umod}.
To address this challenge, we build on scalarization, which generates Pareto solutions by solving single-objective problems parameterized by weights $\bw$ in the $(M-1)$-dimensional simplex $\Delta_M$~\citep{ehrgott2008improved,marler2010weighted}. For example, weighted-sum scalarization yields $\bx^\star(\bw)\in\argmin_{\bx\in\X}\sum_{m=1}^M w_m f_m(\bx)$. Weighted Tchebycheff scalarization (TS)~\citep{steuer1983interactive} can further recover Pareto solutions on nonconvex fronts~\citep{ehrgott2005multicriteria,lin2024smooth}, as detailed in Section~\ref{sec:geometry}. Under suitable regularity, varying $\bw$ induces the weight-to-PF map~\citep{miettinen1999nonlinear,ehrgott2005multicriteria,hillermeier2001generalized}:
\begin{align}
\bw\mapsto\bx^\star(\bw)\mapsto\fPF(\bw):=\f(\bx^\star(\bw))\in\FPF,\quad\bw\in\Delta_M.
\label{eq:weight_pf_map}
\end{align}

\begin{wrapfigure}{r}{0.38\textwidth}
    \centering
    \vspace{-0.6cm}
    \includegraphics[width=\linewidth]{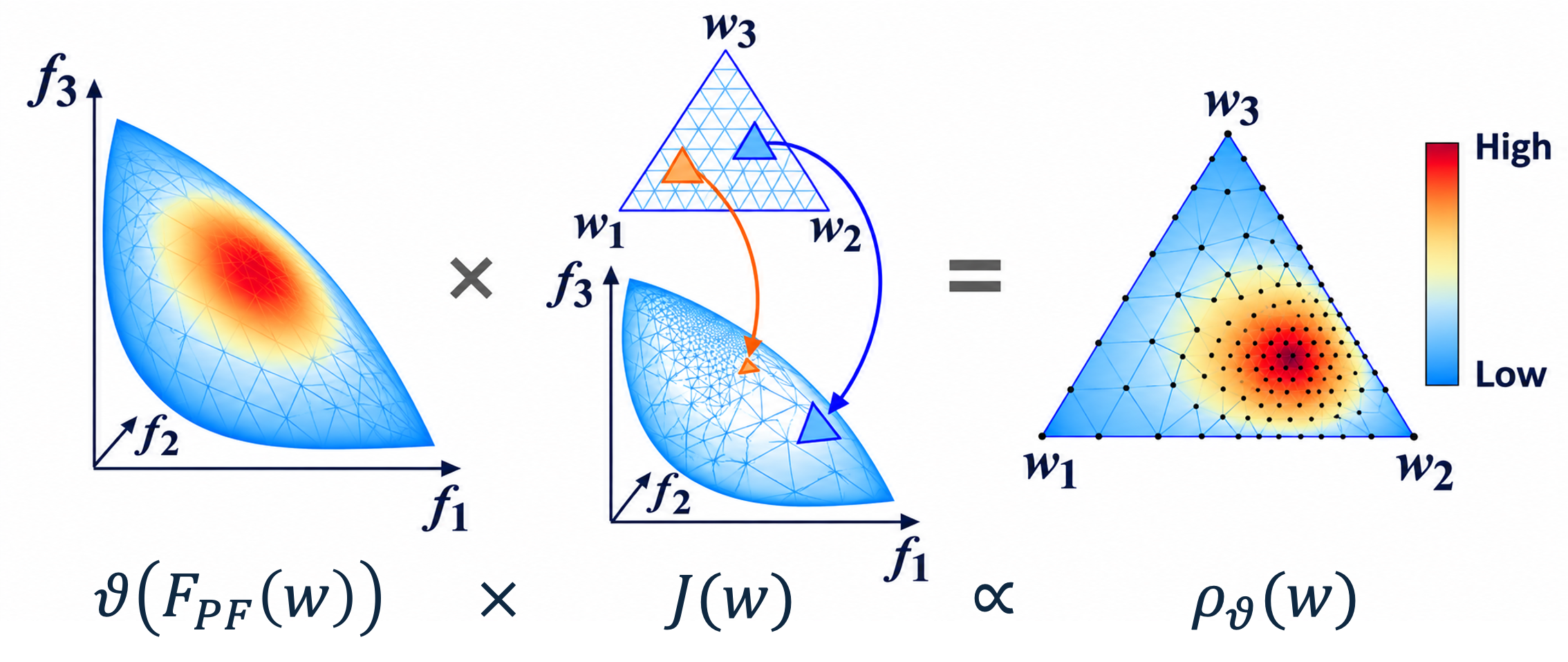}
    \vspace{-0.8cm}
    \caption{Preference $\vartheta(\fPF(\bw))$ and the factor $J(\bw)$ measuring PF area scaling jointly determine the target weight density $\rho_\vartheta(\bw)$. 
    }
    \label{fig:rho_illustration}
    \vspace{-0.8cm}
\end{wrapfigure}
Although the weight simplex and the PF are both $(M-1)$-dimensional, equally sized regions in weight space can correspond to regions of different surface areas on the PF~\citep{jiang2026surf}. We characterize this local area scaling under~\eqref{eq:weight_pf_map} by the factor $J(\bw)$. 
As illustrated in Figure~\ref{fig:rho_illustration}, the change-of-variables formula combines $J(\bw)$ with the prescribed preference $\vartheta$ to yield the \emph{target weight density}: 
\begin{align}
\small
~~~~\underbrace{\rho_\vartheta(\bw)}_{\mathclap{\text{\scriptsize ~~~~~~ target weight density~~~~~}}}
~\propto 
\underbrace{\vartheta(\fPF(\bw))}_{\text{\scriptsize ~~~~PF preference~}}
\underbrace{\sqrt{\det\bigl(D\fPF(\bw)^\top D\fPF(\bw)\bigr)}}_{\text{\scriptsize generalized PF traversal speed }\propto J(\bw)}.
\label{eq:intro_target_density}
\end{align}
Here $D\fPF(\bw)$ denotes the Jacobian of the weight-to-PF map~\eqref{eq:weight_pf_map}. The square-root determinant generalizes the PF traversal speed in SURF~\citep{jiang2026surf}. Section~\ref{sec:geometry} provides the formal change-of-variables derivation and exact definitions of $J(\bw)$ and the normalized $\rho_\vartheta(\bw)$. Realizing this distribution with finitely many scalarization solves motivates our second question:
\begin{center}
\emph{(Q2) How can we use finitely many scalarization weights \\
to approximate the desired PF distribution under a limited computational budget?}
\end{center}

\begin{wrapfigure}{r}{0.66\textwidth}
\vspace{-0.3cm}
\centering
\includegraphics[width=\linewidth]{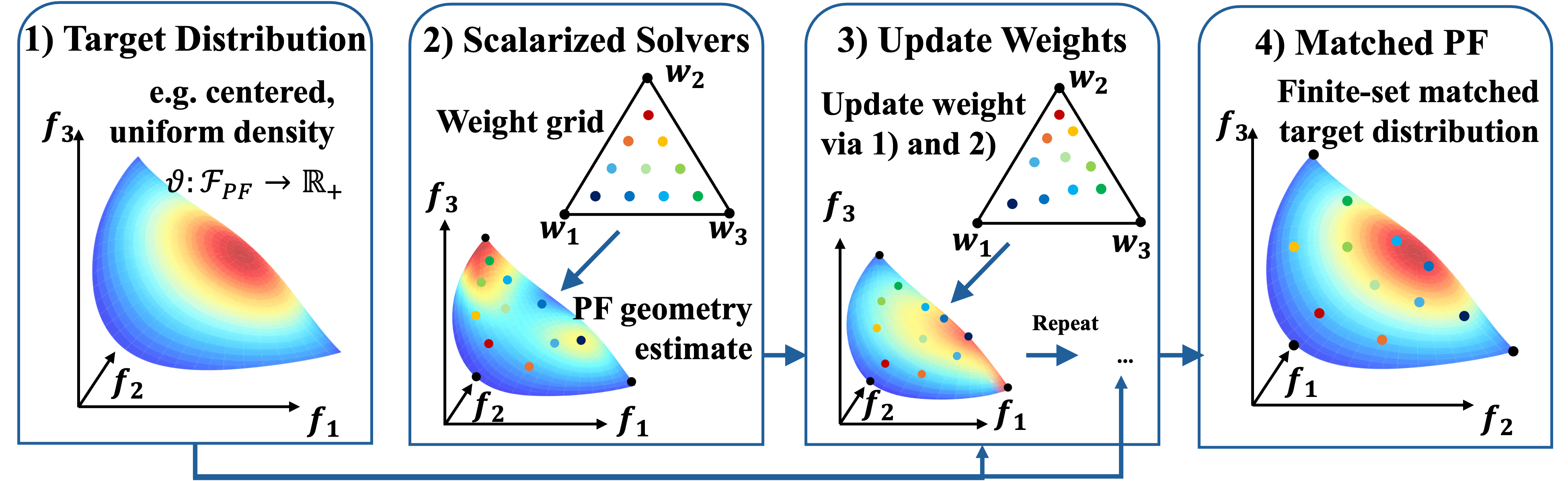}
\vspace{-0.7cm}
\caption{An overview of SURFACE. \textcolor{blue}{Cooler} regions indicate lower target density, while \textcolor{orange}{warmer} regions indicate higher density and a denser allocation of Pareto solutions. SURFACE estimates the weight-to-PF geometry and updates the scalarization weights to match this target distribution.}
\label{fig:surface_overview}
\vspace{-0.2cm}
\end{wrapfigure}
We answer this question with SURFACE (\emph{\textbf{S}calarization \textbf{U}pdates for \textbf{R}edistributing \textbf{F}inite \textbf{A}pproximations with \textbf{C}hart \textbf{E}stimation}), a geometry-aware framework for matching finite Pareto representations to a desired preference distribution.
As illustrated in Figure~\ref{fig:surface_overview}, SURFACE estimates the weight-to-PF geometry from a finite grid of scalarization weights and adaptively redistributes them toward the target weight density $\rho_\vartheta$ in~\eqref{eq:intro_target_density}.
Specifically, it estimates the PF distribution induced by the weights by measuring the area of the PF patch corresponding to each cell of the weight grid. It then updates the weights so that the resulting Pareto solutions move toward the desired preference distribution. 

In summary, our main contributions are:
\begin{enumerate}[label=\textbf{C\arabic*}),leftmargin=*,noitemsep]

\item In Section~\ref{sec:geometry}, we show how a desired preference distribution over the PF maps to a distribution of scalarization weights via the weight-to-PF geometry, and give sufficient conditions for Tchebycheff scalarization to yield a smooth, nondegenerate PF parameterization.

\item In Section~\ref{sec:algorithm}, we introduce SURFACE, which adaptively updates scalarization weights to approximate a target preference distribution with a fixed number of Pareto solutions.
It supports off-the-shelf optimizers, warm-started inexact solves, and parallel subproblem solves.

\item In Section~\ref{sec:theory}, we provide a rigorous convergence analysis, showing that the weight updates contract the preference mismatch linearly down to a finite-resolution floor of $\Oc(N^{-2/(M-1)})$.

\item In Section~\ref{sec:experiments}, we empirically corroborate the effectiveness of SURFACE and verify our theoretical claims: across 9 baselines and benchmarks including ZDT/DTLZ and MO-Gymnasium, SURFACE consistently achieves strong, steady performance.
\end{enumerate}

\vspace{-0.2cm}
\paragraph{Related work.}
Existing approaches to finite PF representation mainly target well-spread solutions through different design criteria. Adaptive scalarization adjusts parameters toward more even objective-space spacing~\citep{eichfelder2009adaptive,zhang2006min,saini2025efficient}, but may require local sensitivity information~\citep{eichfelder2009adaptive} or sequential parameter updates~\citep{zhang2006min,saini2025efficient}. Geometric methods construct additional objective-space directions or constraints~\citep{das1998normal,messac2003normalized,erfani2011directed}, while continuation methods trace the PF sequentially using local derivative or sensitivity information~\citep{hillermeier2001generalized,pereyra2013equispaced,bolten2021tracing}. Evolutionary methods adapt weights or reference points according to sparsity and diversity~\citep{zhang2007moead,qi2014moea,deb2014evolutionary,cheng2016reference}, and finite-design methods use covering, packing, subset-selection, or repulsion objectives~\citep{blank2020generating,borghi2023repulsion,zhang2024umod,shang2023benchmarking}. Preference-conditioned models instead learn mappings from preferences to Pareto solutions~\citep{navon2021learning,lin2022pareto,ye2024pareto}.  SURF~\citep{jiang2026surf} achieves uniform spacing on bi-objective PFs by inverting a cumulative distribution function (CDF) that measures the fraction of PF arc length reached as the weight increases, with density proportional to PF traversal speed.
SURFACE extends this idea to higher-dimensional PFs and user-specified preferences by estimating PF geometry from parallel scalarized solves and adjusting weights accordingly.
Appendix~\ref{app:related_work} provides detailed comparisons with existing methods.

\vspace{-0.2cm}
\paragraph{Notation.}
Let $\Delta_M:=\{\bw\in\mathbb R^M:w_m\ge0,\ \sum_{m=1}^M w_m=1\}$ denote the weight simplex. We write $D$ for the Jacobian with respect to the displayed parameter coordinates, $\|\cdot\|$ for the Euclidean norm, and $\|\cdot\|_\infty$ for the uniform norm. For $k\ge0$, $C^{k,1}$ denotes functions with $k$ continuous derivatives and Lipschitz $k$th derivative, with $\|\cdot\|_{C^k}$ the supremum norm through order $k$. We use $\operatorname{Vol}_{M-1}$ for intrinsic weight-space volume. Recurring symbols are collected in Appendix~\ref{app:notation}.

\vspace{-0.15cm}

\section{Pareto-Front Geometry from Scalarization}\label{sec:geometry}
\vspace{-0.15cm}

Under suitable regularity, the weight-to-PF map~\eqref{eq:weight_pf_map} parameterizes the corresponding PF region~\citep{miettinen1999nonlinear,ehrgott2005multicriteria,hillermeier2001generalized}. Sufficient conditions are established for TS in Section~\ref{sec:tchebycheff_geometry} and weighted-sum in Appendix~\ref{sec:linear_specialization}.
 This allows us to answer \emph{(Q1)} by finding \emph{a distribution of scalarization weights whose induced Pareto outcomes $\fPF(\bw)$ follow the target PF preference $\vartheta$}. The key is to characterize how variations in $\bw$ translate into changes on the PF.

\begin{wrapfigure}{r}{0.45\textwidth}
    \centering
    \vspace{-0.55cm}
    \includegraphics[width=0.85\linewidth]{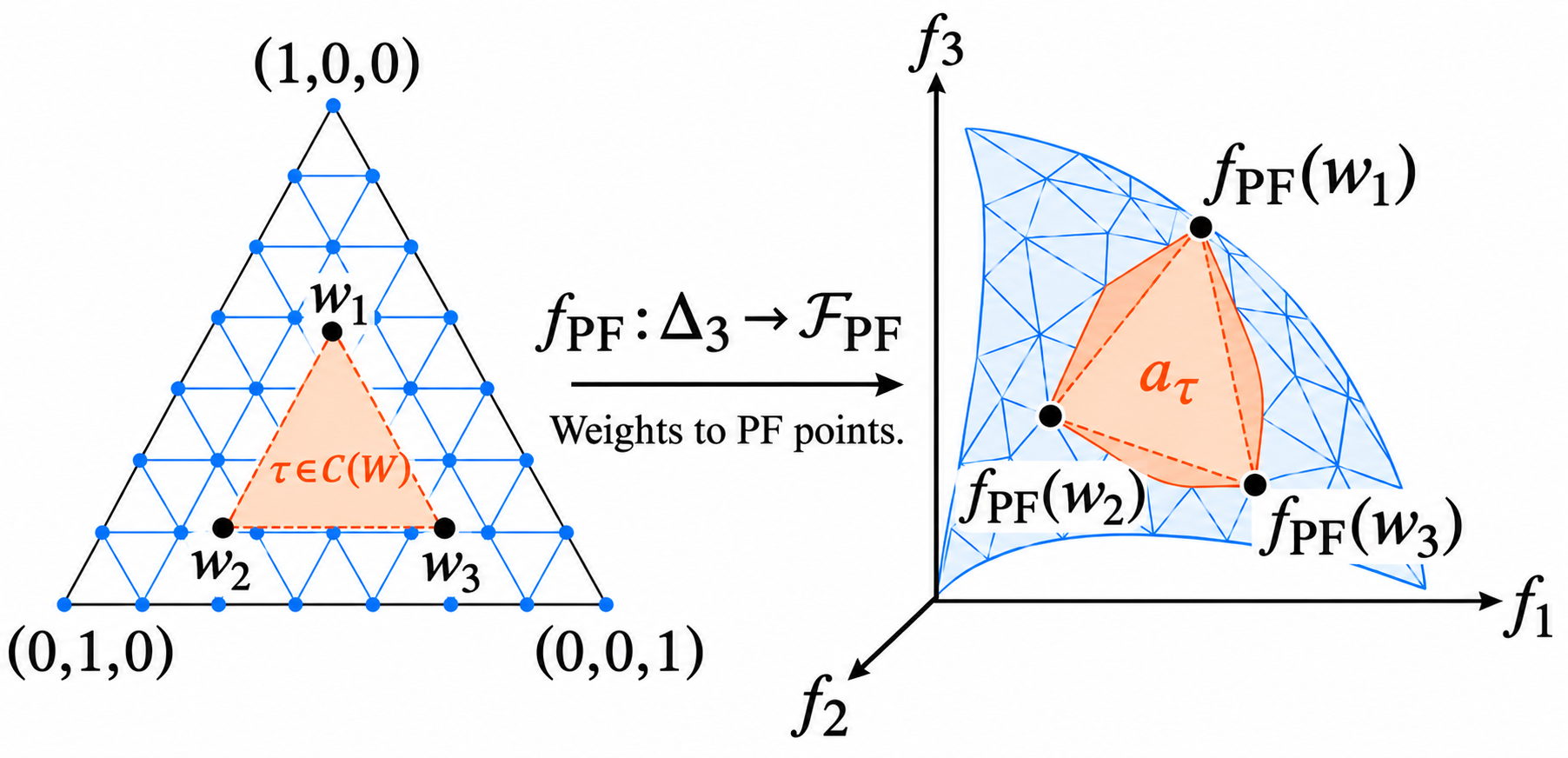}
    \vspace{-0.45cm}
\caption{Weight-to-PF geometry. A simplex cell $\tau$ maps to a PF region $\fPF(\tau)$. The factor $J(\bw)$ generalizes PF traversal speed to local surface-area scaling, with $\int_\tau J(\bw)\,\mathrm d\bw$ giving the area of the corresponding PF region.}
\label{fig:finite_pf_area_proxy}
\vspace{-0.65cm}
\end{wrapfigure}
To characterize how changes in $\bw$ affect the PF, we use the surface-area change-of-variables formula. Following SURF~\citep[Appendix C.3]{jiang2026surf}, we define the \emph{PF-area factor} that measures this local geometric scaling:
\begin{align}
\underbrace{J(\bw)}_{\mathclap{\text{\scriptsize PF-area factor}}}
&:=\tfrac{1}{\sqrt M}
\underbrace{\sqrt{\det\bigl(D\fPF(\bw)^\top D\fPF(\bw)\bigr)}}_{\mathclap{\text{\scriptsize generalized PF traversal speed}}}.
\label{eq:pf_area_density}
\end{align}
Here $D\fPF$ is the Jacobian in $M-1$ independent simplex coordinates, $\mathrm d\bw$ denotes intrinsic $(M-1)$-dimensional simplex volume, and $1/\sqrt M$ accounts for the corresponding volume conversion. A weight-space volume element $\mathrm d\bw$ thus corresponds to PF surface area $J(\bw)\,\mathrm d\bw$, so a simplex cell $\tau\subseteq\Delta_M$ represents area $\int_\tau J(\bw)\,\mathrm d\bw$, as illustrated in Figure~\ref{fig:finite_pf_area_proxy}. For $M=2$, this recovers the PF traversal-speed measure in SURF.
To realize the target preference $\vartheta$, the probability assigned to each weight-space region $\tau$ must be proportional to the preference-weighted PF area $\int_\tau\vartheta(\fPF(\bw))J(\bw)\,\mathrm d\bw$. Normalization yields the \emph{target weight density}
\begin{align}
\underbrace{\rho_\vartheta(\bw)}_{\mathclap{\text{\scriptsize target weight density}}}
:=\frac{\vartheta(\fPF(\bw))J(\bw)}{\int_{\Delta_M}\vartheta(\fPF(\bv))J(\bv)\,\mathrm d\bv}.
\label{eq:preference_pullback_density}
\end{align}
Thus, sampling weights from $\rho_\vartheta$ induces the prescribed PF distribution. In particular, $\vartheta\equiv1$ yields $\rho_1\propto J$, recovering uniform PF surface-area coverage.
To prepare for \emph{(Q2)}, since evaluating $J$ may require costly sensitivity information, we use finitely many sampled weight--PF pairs $\{(\bw_n,\fPF(\bw_n))\}_{n=1}^N$ to estimate the represented PF areas. The following lemma characterizes preference matching across their associated weight-space cells.
\begin{lemma}[Preference Matching]
\label{lem:preference_pullback}
Suppose $\fPF$ is injective and has full tangent rank.
Let 
$\mathcal C(W)$ be a finite measurable partition of $\Delta_M$ associated with the weights $W=\{\bw_n\}_{n=1}^N\subseteq\Delta_M$, with cells of positive volume.
For each $\tau\in\mathcal C(W)$, let
$a_\tau:=\fPF(\tau)$
denote the corresponding PF patch, and define
$\widehat{\mathcal R}_\vartheta(W):=
\max_{\tau\in\mathcal C(W)}
\left| |\mathcal C(W)|\int_{\tau}\rho_\vartheta(\bw)\,\mathrm d\bw-1\right|$.
Then $\widehat{\mathcal R}_\vartheta(W)=0$ if and only if all PF patches
$\{a_\tau\}_{\tau\in\mathcal C(W)}$ carry equal preference-weighted area.
\end{lemma}
Appendix~\ref{app:prelim} provides the proof of Lemma~\ref{lem:preference_pullback}. The residual measures preference-mass imbalance across $\mathcal C(W)$. The separate weight mesh used to reconstruct the PF is introduced in Section~\ref{sec:algorithm}. For $\vartheta\equiv1$, this residual measures PF surface-area imbalance.

\subsection{Regular PF Charts from TS}
\label{sec:tchebycheff_geometry}
\vspace{-0.1cm}
In this section, we specialize the weight-to-PF map to TS and establish its PF coverage and regular chart properties. Appendix~\ref{sec:linear_specialization} provides an alternative analysis for weighted-sum scalarization. 

TS is especially useful for PF construction because it can recover Pareto solutions on nonconvex fronts that some other scalarizations may miss~\citep{ehrgott2005multicriteria,lin2024smooth}.
The conventional weighted TS takes the form $\min_{\bx\in\X}\max_m\lambda_m(f_m(\bx)-z_m)$, where $\boldsymbol{\lambda}\in\operatorname{relint}(\Delta_M)$ and $z_m$ is a reference point chosen strictly below the minimum value of each objective. Equivalently, under the reciprocal reparameterization $\lambda_m=\frac{w_m^{-1}}{\sum_{j=1}^M w_j^{-1}}$~\citep{lin2022pareto}, TS can be written as
\begin{align}
\bx_{\rm TS}^\star(\bw)
\in
\argmin_{\bx\in\X}
\max_{m=1,\ldots,M}
\frac{f_m(\bx)-z_m}{w_m},
\quad
z_m < \min_{\bx\in\X}f_m(\bx),
\quad
\bw\in\operatorname{relint}(\Delta_M).
\label{eq:standard_tchebycheff}
\end{align}
We adopt this formulation because $\bw$ directly parameterizes reference directions in objective space, facilitating the geometric analysis of the weight-to-PF map.
For $\bw\in\operatorname{relint}(\Delta_M)$, TS in~\eqref{eq:standard_tchebycheff} 
admits the equivalent epigraph formulation~\eqref{eq:ray_epigraph}, which lets us state the conditions for a smooth solution path.

\begin{assumption}[Regular full-ray TS regime]
\label{ass:tch_route}
Fix any interior weight margin $0<\delta<1/M$ and define $\Delta_M^\delta:=\{\bw\in\Delta_M:w_m\ge\delta,\ m=1,\ldots,M\}$. Suppose $\X$ is compact and $\bz$ is utopian, i.e., $z_m<\min_{\bx\in\X}f_m(\bx)$ for every $m$. For every $\bw\in\Delta_M^\delta$, the problem
\begin{align}
\min_{\bx\in\X, s\in\mathbb R}\quad &s
\quad\text{s.t.}\quad
f_m(\bx)-z_m\le s w_m,\quad m=1,\ldots,M,
\label{eq:ray_epigraph}
\end{align}
has a unique solution $(\bx_{\rm TS}^\star(\bw),s^\star(\bw))$. At this solution, all $M$ ray inequalities are active, the linear independence constraint qualification (LICQ) holds, every active inequality has a positive multiplier, and the second-order sufficient condition (SOSC) holds on the critical cone.
\end{assumption}

Compactness and a utopian $\bz$ are standard in TS~\citep{ehrgott2005multicriteria}. Full activity, LICQ, positive multipliers, and SOSC are checkable from the ray residuals, active-constraint Jacobian, multiplier signs, and reduced Hessian~\citep{fiacco1976sensitivity,bonnans2013perturbation}. Appendix~\ref{app:tch_sufficient_regimes} gives concrete sufficient regimes for uniqueness and these regularity conditions. 
When the standard Tchebycheff formulation is nonsmooth, smooth Tchebycheff scalarizations provide a related alternative~\citep{lin2024smooth}. These conditions ensure a smooth, nonfolding TS parameterization over $\Delta_M^\delta$.

\begin{proposition}[PF coverage and injectivity of TS]
\label{prop:tch_coverage}
Under Assumption~\ref{ass:tch_route}, TS maps $\Delta_M^\delta$ injectively onto the corresponding PF patch $\fPF(\Delta_M^\delta)$. Moreover, every Pareto point whose normalized positive-gap ray lies in $\Delta_M^\delta$ is represented.
\end{proposition}

Injectivity follows from full ray activity in Assumption~\ref{ass:tch_route}, which gives
$\fPF(\bw)=\bz+s^\star(\bw)\bw$ with $s^\star(\bw)>0$, so each represented PF point uniquely determines its weight.
Coverage follows from classical TS completeness~\citep{miettinen1999nonlinear,ehrgott2005multicriteria}, which applies to the represented positive-gap rays without requiring PF convexity.
The following standard condition~\citep{fiacco1976sensitivity,bonnans2013perturbation} further ensures the smoothness and regularity of this parameterization.

\begin{assumption}[Feasibility and Smoothness]
\label{ass:common_model}
The feasible set $\X=\{\bx:\mathbf c_1(\bx)\le\mathbf0,\ \mathbf c_2(\bx)=\mathbf0\}$ is nonempty and closed. Every $f_m$ and every component of $\mathbf c_1$ is $C^{3,1}$ on a common open convex neighborhood of $\X$, each component of $\mathbf c_1$ is convex there, and $\mathbf c_2$ is affine.
\end{assumption}

Together, these assumptions yield the following regularity properties of the TS-induced PF chart.

\begin{proposition}[Regular PF chart from TS]
\label{prop:pf_geometry}
Suppose Assumptions~\ref{ass:tch_route} and~\ref{ass:common_model} hold. Then the TS-induced PF chart has the following properties. (i) \textbf{Smoothness.} $\bx_{\rm TS}^\star$ is uniformly Lipschitz and face-wise $C^{2,1}$ on $\Delta_M^\delta$, with compatible one-sided extensions, and $\fPF=\f\circ\bx_{\rm TS}^\star$ is face-wise $C^{2,1}$. (ii) \textbf{Nonfolding.} $\fPF$ is injective, has full tangent rank, and there exist $0<c_P\le C_P<\infty$ such that
\begin{align}
c_P\|\bw-\bv\|
\le \|\fPF(\bw)-\fPF(\bv)\|
\le C_P\|\bw-\bv\|,
\quad \bw,\bv\in\Delta_M^\delta.
\label{eq:pf_chart_bilipschitz}
\end{align}
(iii) \textbf{Nondegenerate area.} $J$ is face-wise $C^{1,1}$ with compatible one-sided traces and satisfies $\underline J\le J(\bw)\le\overline J$ for some $0<\underline J\le\overline J<\infty$ and all $\bw\in\Delta_M^\delta$.
\end{proposition}

The proof of Proposition~\ref{prop:pf_geometry} is given in Appendix~\ref{app:proof_pf_geometry}. Thus TS induces a smooth, nonfolding PF chart with uniformly nondegenerate area distortion on each $\Delta_M^\delta$. These properties extend to $\operatorname{relint}(\Delta_M)$ when the full-ray regularity conditions hold throughout the interior, allowing an ideal reference point. For example, DTLZ2~\citep{deb2002scalable} with $\bz=\mathbf0$ admits $\fPF(\bw)=\bw/\|\bw\|_2$, whose geometric regularity bounds remain uniform as $\delta\downarrow0$. Combined with Lemma~\ref{lem:preference_pullback}, these properties yield $\rho_\vartheta(\bw)\propto\vartheta(\fPF(\bw))J(\bw)$, with $\rho_1(\bw)\propto J(\bw)$ recovering uniform PF surface-area coverage.

\section{The SURFACE Algorithm}
\label{sec:algorithm}

\begin{algorithm}[t]
\caption{SURFACE}
\label{alg:surface}
\begin{algorithmic}[1]
\Require Preference $\vartheta$, number of adaptive weights $N$, chosen weight domain (e.g. $\Delta_M^\delta$) and its $M$ (fixed) corner samples $\{(\bw_m^{\delta},\tilde{\f}_m)\}_{m=1}^M$, initial solver states $\{\tilde\bx_n^{(-1)}\}_{n=1}^N$, inner-step budget $K$, damping $\alpha\in(0,1]$, and outer rounds $T$
\For{$t=0,\ldots,T-1$}
    \State $\{\bw_n^{(t)}\}_{n=1}^N\leftarrow\operatorname{MEQ}_{N}(\rho_t)$ via Appendix~\ref{app:finite_meq}\Comment{Initialize $\rho_0$ uniformly on $\Delta_M^\delta$}
    \State $\tilde\bx_n^{(t)}\leftarrow\textsc{InnerSolver}_K\left(\bw_n^{(t)}, \bx_{\rm init}=\tilde\bx_n^{(t-1)}\right)$ in parallel for $n\in[N]$
    \State \textbf{Optional:} refine $\tilde{\f}_m$ towards $\f(\bx^\star_{\rm TS}(\bw_m^{\delta}))$ in parallel for $m\in[M]$
    \State Interpolate $P_t$ from $\{(\bw_n^{(t)},\f(\tilde\bx_n^{(t)}))\}_{n=1}^N\cup \{(\bw_m^\delta,\tilde{\f}_m)\}_{m=1}^M$ via Appendix~\ref{app:reconstruction_proof}
    \State Update $\rho_{t+1}$ by~\eqref{eq:density_update}
\EndFor
\Ensure $\{(\bw_n^{(T-1)},\f(\tilde\bx_n^{(T-1)}))\}_{n=1}^N$ and corner samples $\{(\bw_m^{\delta},\tilde{\f}_m)\}_{m=1}^M$
\end{algorithmic}
\end{algorithm}

Building on the target weight density $\rho_\vartheta\propto(\vartheta\circ\fPF)J$ from Section~\ref{sec:geometry}, we propose SURFACE, a general framework for adaptive weight selection under different scalarizations.
SURFACE combines parallel scalarized solves, PF reconstruction, and mass equidistribution (MEQ) to match the prescribed preference.
In this section, we use TS on $\Delta_M^\delta$ as the running specialization, with fixed corners $\bw_m^\delta:=\delta\mathbf1+(1-M\delta)\mathbf e_m$.
When the PF chart extends to the full simplex $\Delta_M$, the fixed corners become the vertices $\{\mathbf e_m\}_{m=1}^M$, as illustrated by DTLZ2~\citep{deb2002scalable} with an ideal reference point (cf. Appendix~\ref{app:tch_sufficient_regimes}).
Appendix~\ref{sec:linear_specialization} gives the weighted-sum specialization on $\Delta_M$.

SURFACE retains $M$ fixed TS chart-corner samples $\tilde{\f}_m\approx\fPF(\bw_m^\delta)$ alongside the $N$ adaptive samples $\{\bw_n^{(t)}\}_{n=1}^N$. When $\fPF$ extends continuously to $\Delta_M$, the chart corners approach $\fPF(\mathbf e_m)$ as $\delta\downarrow0$.
At outer round $t$, SURFACE applies finite $K$-step \textsc{InnerSolver} in parallel on the scalarized objectives to obtain
\begin{align}
\tilde\bx_n^{(t)} = \textsc{InnerSolver}_K \left(\bw_n^{(t)}, \bx_{\rm init}=\tilde\bx_n^{(t-1)}\right)\approx \bx^\star_{\rm TS}(\bw_n^{(t)}),
\quad n=1,\ldots,N.
\label{eq:surface_inner_solve}
\end{align}
For differentiable objectives, scalarization can be applied to the losses, and \textsc{InnerSolver} may use Adam~\citep{kingma2015adam}. For reinforcement learning, scalarization is applied to the rewards, and \textsc{InnerSolver} may use Proximal Policy Optimization (PPO)~\citep{schulman2017proximal} or Group Relative Policy Optimization (GRPO)~\citep{shao2024deepseekmath}.
Warm-starting from the preceding round with $\bx_{\rm init}=\tilde\bx_n^{(t-1)}$ helps $\textsc{InnerSolver}_K$ track the updated scalarized solutions with inexact solves, as detailed in the next section. The corner references $\{\tilde{\f}_m\}_{m=1}^M$ may be held fixed when accurate estimates are available or refined similarly in parallel. 

Using the $N$ weight-objective pairs $\{(\bw_n^{(t)},\f(\tilde\bx_n^{(t)}))\}_{n=1}^N$ and the $M$ corner pairs $\{(\bw_m^\delta,\tilde{\f}_m)\}_{m=1}^M$, SURFACE constructs a smooth PF approximation $P_t\approx \fPF$ through interpolation. 
One practical choice fits local radial basis function (RBF) models with quadratic terms and blends them smoothly~\citep{wendland2005scattered}. Appendix~\ref{app:reconstruction_proof} gives the interpolation details.
For problems such as DTLZ2~\citep{deb2002scalable} with $\bz=\mathbf0$, this reconstruction also extends to $\Delta_M$ as $\delta\downarrow0$, with the simplex vertices as fixed corners.
Replacing $\fPF$ by $P_t$ in~\eqref{eq:pf_area_density} gives the estimated area factor $J_{P_t}(\bw):=\tfrac{1}{\sqrt M}\sqrt{\det\bigl(DP_t(\bw)^\top DP_t(\bw)\bigr)}$.
SURFACE updates the weight density using this geometry and the prescribed PF preference, with damping factor $\alpha\in(0,1]$:
\begin{align}
\rho_{t+1}(\bw)
:=
(1-\alpha)\rho_t(\bw)
+
\alpha
\frac{\vartheta(P_t(\bw))J_{P_t}(\bw)}
{\int_{\Delta_M^\delta} \vartheta(P_t(\bv))J_{P_t}(\bv) \mathrm d\bv}, \quad \bw \in \Delta_M^\delta .
\label{eq:density_update}
\end{align}
For $\vartheta=1$, it targets uniform surface-area coverage of the corresponding PF patch.

Finally, SURFACE converts $\rho_{t+1}$ into $N$ adaptive weights for the next round through MEQ~\citep{budd2009moving}. By Lemma~\ref{lem:preference_pullback}, MEQ seeks weights whose associated cells carry equal $\rho_{t+1}$-mass.
Let $W=\{\bw_n\}_{n=1}^N\cup\{\bw_m^\delta\}_{m=1}^M$ include the adaptive weights and fixed corners, and let $\mathcal C(W)$ be an associated cell partition of $\Delta_M^\delta$. MEQ redistributes the weights as follows:
\begin{align}
\{\bw_n^{(t+1)}\}_{n=1}^N
&=\operatorname{MEQ}_N(\rho_{t+1})
\approx
\argmin_{\{\bw_n\}_{n=1}^N}
\max_{\tau\in\mathcal C(W)}
\left|
|\mathcal C(W)|
\int_\tau\rho_{t+1}(\bw)\,\mathrm d\bw-1
\right|.
\label{eq:meq_target}
\end{align}
SURFACE realizes this objective by transporting an initially equal-volume tessellation, producing smaller cells in regions of higher density.
For $M=2$, MEQ can be implemented by inverse cumulative-distribution transport~\citep{jiang2026surf}.
For $M\ge3$, Appendix~\ref{app:finite_meq} establishes two realizations: Moser transport~\citep{bruveris2018moser} and equal-capacity minimization, where equal capacity means equal probability mass per cell.

Algorithm~\ref{alg:surface} summarizes the iteration, which requires $\Oc(TN)$ scalarized solves. The dominant cost is typically \textsc{InnerSolver}, since reconstruction and MEQ operate only in the $(M- 1)$-dimensional weight space and require no additional scalarized solves.

\section{Convergence Analysis}
\label{sec:theory}

This section establishes the convergence of SURFACE by tracking the three approximations in Section~\ref{sec:algorithm}: finite PF reconstruction via interpolation, numerical MEQ, and inexact inner scalarized solves. 
We use TS on a fixed $\Delta_M^\delta$ from Proposition~\ref{prop:pf_geometry} and normalize~\eqref{eq:preference_pullback_density} over this domain, targeting the represented PF patch $\fPF(\Delta_M^\delta)$. We impose the following condition on the preference density.

\begin{assumption}[Regular preference]
\label{ass:preference_profile}
The preference density $\vartheta:\FPF\to\mathbb R_+$ extends to a $C^{2,1}$ function on an open neighborhood containing $\FPF$ and $P_t(\Delta_M^\delta)$ for all $t$, with $0<\underline{\vartheta}\le\vartheta(\by)\le\overline{\vartheta}<\infty$ and $\|\vartheta\|_{C^{2,1}}\le B_\vartheta$ for some constants $\underline{\vartheta},\overline{\vartheta},B_\vartheta$.
\end{assumption}
To characterize the target distortion, define the preference-weighted condition number $\kappa_\vartheta:=\overline\vartheta\,\overline J/(\underline\vartheta\,\underline J)$. 
For finite PF reconstruction, RBF interpolation~\citep{wendland2005scattered} gives second-order $C^1$ accuracy, while inexact scalarized solves introduce an additional reconstruction-error term. 

\begin{lemma}[Reconstruction and density accuracy]
\label{lem:main_reconstruction}
Suppose Assumptions~\ref{ass:tch_route}--\ref{ass:preference_profile} hold. Then there exists an augmented RBF interpolation $P_t$ (cf. Appendix~\ref{app:reconstruction_proof}) such that
\begin{align}
\Big\|
\tfrac{\vartheta(P_t)J_{P_t}}
{\int \vartheta(P_t)J_{P_t}}
-\rho_\vartheta
\Big\|_\infty
\le
{C C_t N^{-\frac{2}{M-1}}}
+
C N^{\frac{1}{M-1}}
\max_{n\in[N]}
\big\|
\tilde\bx_n^{(t)}
-\bx^\star_{\rm TS}(\bw_n^{(t)})
\big\|,
\label{eq:main_reconstruction_error}
\end{align}
where $C$ and the round-dependent coefficient $C_t$ are bounded uniformly over the rounds.
\end{lemma}

The interpolation details and its proof are provided in Appendix~\ref{app:reconstruction_proof}.
With exact scalarized solves, Lemma~\ref{lem:main_reconstruction} gives a density error of order $C C_t N^{-\frac{2}{M-1}}$, where $C$ is independent of $\kappa_\vartheta$ under uniform regularity conditions.
The coefficient $C_t$ captures the round-dependent reconstruction geometry and may be large when $\kappa_\vartheta$ is large.
As MEQ redistributes weights toward the target preference, it can improve the reconstruction geometry and reduce $C_t$.
For inexact solves, the second term captures the amplification of inner optimization error by the inverse mesh scale.
We next analyze numerical MEQ error, where both realizations balance the $\rho_{t+1}$-mass of their complete represented cells.

\begin{proposition}[Finite MEQ stability]
\label{prop:main_meq}
Suppose Assumptions~\ref{ass:tch_route}--\ref{ass:preference_profile} hold.
Then there exist numerical MEQ realizations (cf. Appendix~\ref{app:finite_meq}) that achieve $\Oc(N^{-2/(M-1)})$ cell-mass error on complete represented cells while preserving mesh regularity for stable reconstruction in the next round.
\end{proposition}
Thus numerical MEQ has the same $\Oc(N^{-\frac{2}{M-1}})$ resolution as PF reconstruction while preserving the mesh regularity needed by Lemma~\ref{lem:main_reconstruction}. Appendix~\ref{app:finite_meq} gives the construction and proof. 

Finally, we quantify the effect of solving each scalarized problem for $K$ inner steps.
Following local convergence and sensitivity theory~\citep{bonnans2013perturbation} and the finite-step analysis of SURF~\citep{jiang2026surf}, let $\eta_K$ denote a $K$-step contraction factor such that $\|\bx_K-\bx_{\rm TS}^\star(\bw)\|\le\eta_K\|\bx_0-\bx_{\rm TS}^\star(\bw)\|$.
For linearly convergent methods, $\eta_K$ decreases geometrically in $K$; a sublinear rate $\eta_K=\Oc(K^{-p})$ applies when the solver satisfies the corresponding relative decision-error contraction.
Since damping limits the motion of successive weights, $\alpha$ trades outer progress against the required inner accuracy. 
We therefore define the coupled density-solver error
\begin{align}
\mathcal E_t
:=
\|\rho_t-\rho_\vartheta\|_\infty
+
C_0\left(1+\alpha N^{1/(M-1)}\right)
\max_{n\in[N]}
\left\|
\tilde\bx_n^{(t)}
-\bx^\star_{\rm TS}(\bw_n^{(t)})
\right\|.
\label{eq:coupled_metric}
\end{align}
The following theorem establishes SURFACE's joint convergence and preference-mass guarantees.

\begin{theorem}[SURFACE convergence]
\label{thm:surface_inexact}
Suppose Assumptions~\ref{ass:tch_route}--\ref{ass:preference_profile} hold.
Choose $N=\Omega(\kappa_\vartheta^{3(M-1)/2})$ sufficiently large, $0<\alpha\le\bar\alpha<1$ with $\bar\alpha=\Omega(N^{-3/(2(M-1))})$, and $K$ for the inner solver with $\eta_K$ satisfying $\bar\alpha^2\eta_K=\Oc(N^{-3/(M-1)})$. Initialize the inner solves to accuracy $\Oc(N^{-3/(M-1)})$.
Then there exist an interpolation (cf. Appendix~\ref{app:reconstruction_proof}) and a moving-weight Moser MEQ realization (cf. Appendix~\ref{app:finite_meq}) of Algorithm~\ref{alg:surface} such that, for $t=0,\ldots,T-1$,
\begin{align}
\mathcal E_t
&\le (1-\alpha/4)^t\mathcal E_0
+\Oc\left(N^{-\frac{2}{M-1}}\right),
\quad
\widehat{\mathcal R}_\vartheta(W^{(t)})
\le C\mathcal E_t
+\Oc\left(N^{-\frac{2}{M-1}}\right).
\label{eq:coupled_rate}
\end{align}
Here, $W^{(t)}:=\{\bw_n^{(t)}\}_{n=1}^N\cup\{\bw_m^\delta\}_{m=1}^M$ is the weight mesh at round $t$, and $\widehat{\mathcal R}_\vartheta(W^{(t)})$ is the target-mass residual in Lemma~\ref{lem:preference_pullback} evaluated on complete represented cells of the selected MEQ realization.
\end{theorem}

\begin{table}[t]
\centering
\small
\begin{tabular}{@{}c|c|c|c|c@{}}
\toprule
\textbf{Damping $\bar\alpha$} & \textbf{Contraction $\eta_K$} & \textbf{Inner $K$} & \textbf{Outer $T$} & \textbf{When to Use}\\
\midrule
$\Theta(1)$ & $\Oc(N^{-3/(M-1)})$ & $\Oc(\log N)$ & $\Oc(\log N)$ & Linear/superlinear or $p>2$\\
$\Theta(N^{-1/(M-1)})$ & $\Oc(N^{-1/(M-1)})$ & $\Oc(\log N)$ & $\Oc(N^{1/(M-1)}\log N)$ & $p=2$ / moderate $K$\\
$\Theta(N^{-3/(2(M-1))})$ & $\Oc(1)$ & $\Oc(1)$ & $\Oc(N^{3/(2(M-1))}\log N)$ & $p<2$ / fixed $K$\\
\bottomrule
\end{tabular}
\vspace{-0.4cm}

\caption{Outer--inner iteration trade-offs under $\bar\alpha^2\eta_K=\Oc(N^{-3/(M-1)})$. The $K$ and $T$ columns assume linear inner convergence and $\alpha=\Theta(\bar\alpha)$, respectively. The ``When to Use'' column identifies preferred damping regimes for sublinear inner solvers with $\eta_K=\Oc(K^{-p})$.}

\label{tab:outer_inner_regimes}
\end{table}
The convergence proof is given in Appendix~\ref{app:proofs_convergence}. With $N=\Omega(\kappa_\vartheta^{3(M-1)/2})$ sufficiently large for the regularity thresholds, MEQ keeps $C_t=\Oc(1)$ across rounds and yields an $\Oc(N^{-2/(M-1)})$ error floor for each fixed regular problem.
Additionally, Table~\ref{tab:outer_inner_regimes} summarizes the inner–outer trade-off: larger $\bar\alpha$ requires more accurate inner solves but fewer outer rounds, whereas smaller $\bar\alpha$ permits cheaper inner solves at the cost of more outer rounds. For sublinear inner convergence $\eta_K=\Oc(K^{-p})$, the cases $p>2$, $p=2$, and $p<2$ require inner cost $KT=\widetilde{\Oc}(\epsilon^{-3/(2p)})$, $\widetilde{\Oc}(\epsilon^{-1/2-1/(2p)})$, and $\widetilde{\Oc}(\epsilon^{-3/4})$, respectively. 
Thus, for repeated inner work, larger damping steps are preferable for $p>2$ and smaller steps for $p<2$. The initial states $\{\tilde\bx_n^{(-1)}\}_{n=1}^N$ can be obtained through a one-time $\textsc{InnerSolver}_{K_0}$ warmup to meet the stated initialization accuracy.

\section{Experiments}
\label{sec:experiments}

We empirically evaluate SURFACE against fixed Uniform weighting and adaptive/stochastic baselines with matched returned-solution counts, comparing computational budgets through matched solver work or wall-clock evaluations. We adopt $\widehat{\mathcal R}_\vartheta$ ($\downarrow$) from Lemma~\ref{lem:preference_pullback} as our primary metric of preference distribution matching, using a consistent evaluation partitioning rule for cross-method comparisons. We also report hypervolume (HV, $\uparrow$) and its normalized gap ($\Delta\mathrm{HV}$, $\downarrow$) for objective-space quality. We emphasize uniform PF coverage ($\vartheta\equiv1$), an important \emph{a posteriori} MOO task, evaluating PF-area uniformity through the target-mass residual and average and worst-case coverage through inverted generational distance (IGD, $\downarrow$) and fill distance ($\downarrow$), respectively~\citep{ishibuchi2015study,zhang2024umod}.
Section~\ref{sec:exp_controlled} examines our theoretical guarantees, Section~\ref{sec:exp_finite_coverage} benchmarks PF coverage under TS on ZDT/DTLZ~\citep{zitzler2000comparison,deb2002scalable} problems, and Section~\ref{sec:exp_policy} demonstrates applicability beyond TS using weighted-sum scalarization in multi-objective reinforcement learning. Further experimental protocols, metric definitions, implementation details, and additional experiments are in Appendix~\ref{app:benchmarks}.

\subsection{Controlled Geometry and Preference Matching}
\label{sec:exp_controlled}

\begin{wrapfigure}{r}{0.38\textwidth}
\centering
\vspace{-0.6cm}
\includegraphics[width=0.94\linewidth]{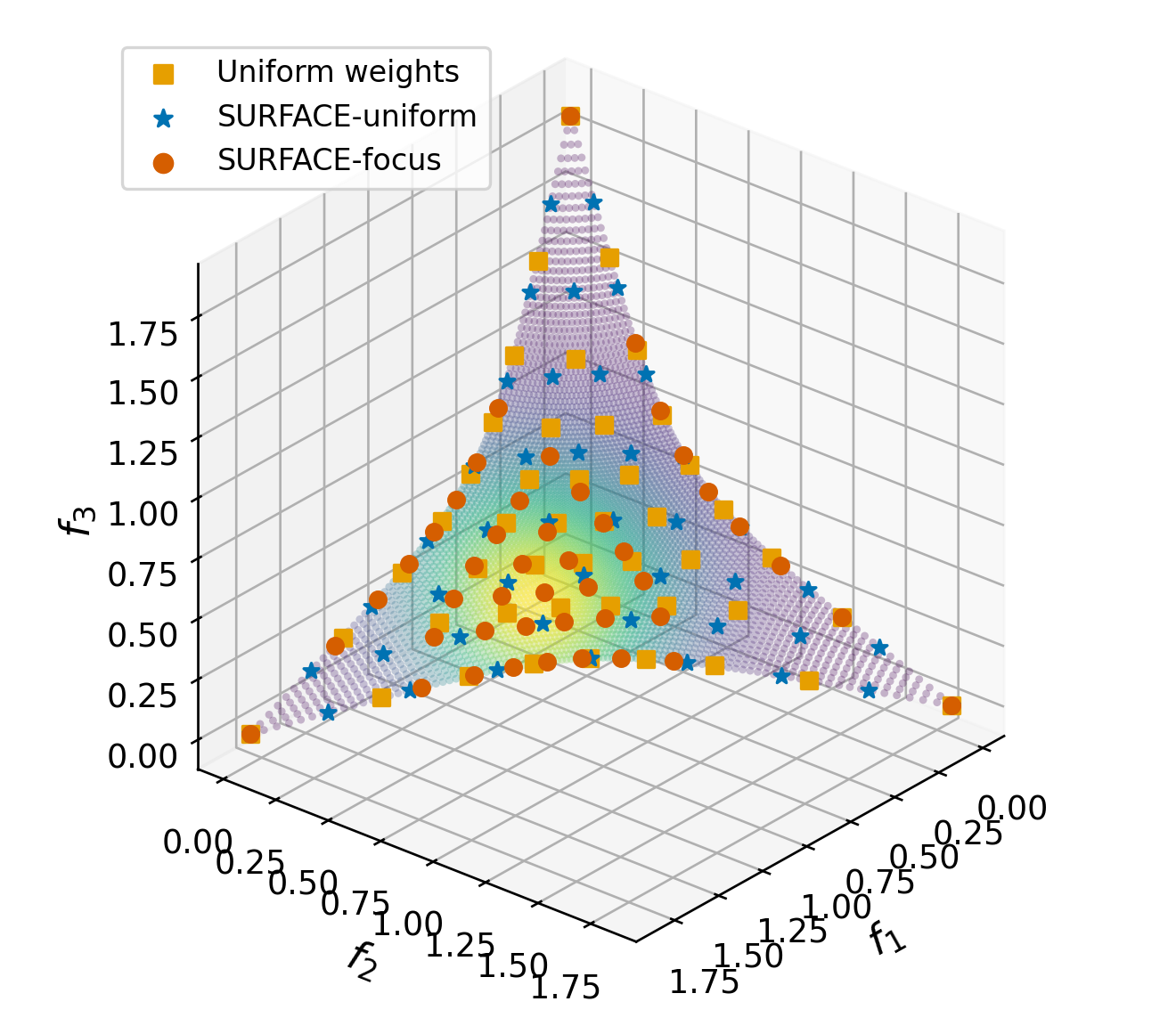}
\vspace{-0.4cm}
\caption{Uniform weighting versus SURFACE with $\vartheta= 1,\vartheta_{\rm focus}$. Warmer colors indicate higher $\vartheta_{\rm focus}$.}
\label{fig:general_preference_m3}
\vspace{0.2cm}
\includegraphics[width=0.8\linewidth]{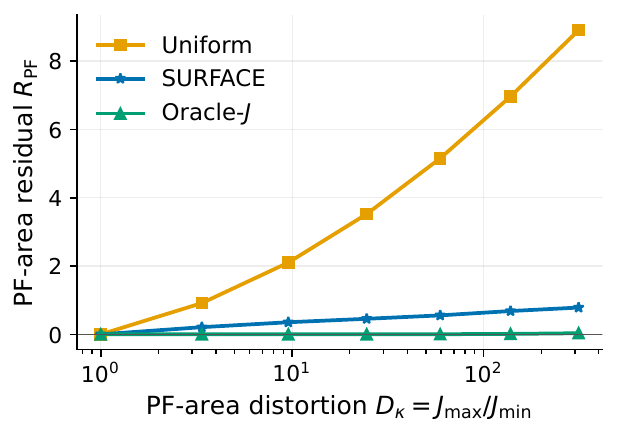}
\vspace{-0.4cm}
\caption{Residual $\widehat{\mathcal R}_\vartheta(W_T)$ versus PF distortion. SURFACE (Exact $J$) uses the analytic density.}
\label{fig:controlled_radial_distortion}
\vspace{-.8cm}
\end{wrapfigure}

This section considers an analytic three-objective problem:
\begin{align}
\min_{\bx\in\Delta_3}
\f(\bx)
=
\exp\left(
\kappa\left\|\bx-\tfrac13\mathbf1\right\|_2^2
\right)\bx,
\quad
0\leq \kappa\leq 3.
\label{eq:radial_moo_problem}
\end{align}
TS with $\bz=\mathbf0$ yields $\fPF(\bw)=\exp(\kappa\|\bw-\tfrac13\mathbf1\|_2^2)\bw$. Its exact area-distortion ratio over $\Delta_3$ is $\overline J/\underline J=e^{4\kappa/3}\sqrt{1+8\kappa/3+8\kappa^2/3}$, which increases with $\kappa$.
Setting $\kappa=1$, we first illustrate how SURFACE controls \emph{where} finite PF resolution is allocated through $\vartheta$. We consider the uniform preference $\vartheta = 1$ and a focused preference
$\vartheta  =  \vartheta_{\rm focus}(\by)=\exp \left(-\tfrac{\|\by-\fPF([0.55,0.30,0.15]^\top)\|_2^2}{2\sigma^2}\right)$,
where $\sigma$ is $15\%$ of the PF diameter. 
As shown in Figure~\ref{fig:general_preference_m3},  using $N=63$, $\alpha=1$, and $T=5$ on \eqref{eq:radial_moo_problem} with $\kappa=1$, SURFACE under the focused preference concentrates Pareto solutions around the target region, while SURFACE with $\vartheta = 1$ redistributes samples toward uniform PF surface-area coverage. In contrast, fixed Uniform weighting inherits the uncontrolled weight-to-PF distortion and concentrates toward the PF center.

\begin{figure}[t]
\centering

\begin{minipage}[t]{0.58\linewidth}
\vspace{0pt}
\centering
\begin{minipage}[t]{0.48\linewidth}
    \centering
    \includegraphics[width=\linewidth]{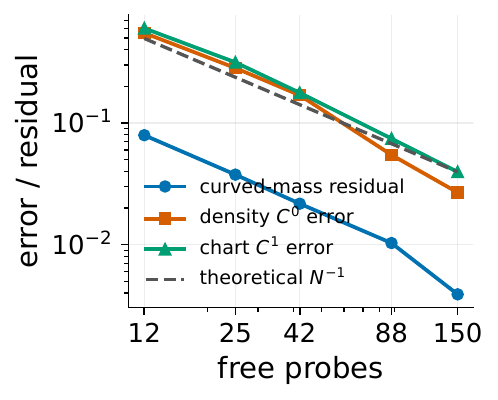}
\end{minipage}
\hfill
\begin{minipage}[t]{0.48\linewidth}
    \centering
    \includegraphics[width=\linewidth]{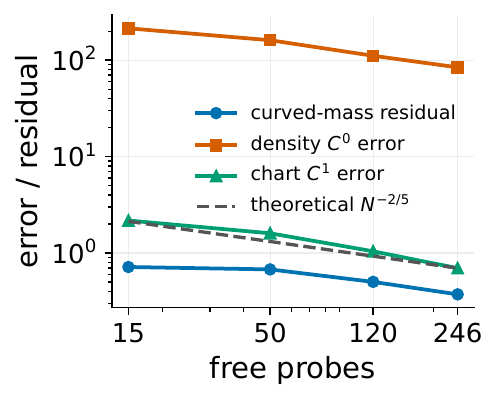}
\end{minipage}

\vspace{-0.3cm}
\captionof{figure}{Finite-resolution errors versus $N$ for $M=3$ (left) and $M=6$ (right). The decay is consistent with the predicted $\Oc(N^{-1})$ and $\Oc(N^{-2/5})$ rates.}
\label{fig:controlled_rates}
\end{minipage}
\hfill
\begin{minipage}[t]{0.39\linewidth}
\vspace{-0.1cm}
\centering
\includegraphics[width=0.76\linewidth]{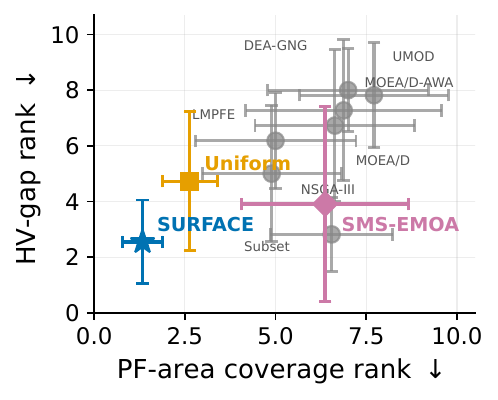}

\vspace{-0.455cm}
\captionof{figure}{Mean ranks on PF coverage and HV over eleven ZDT/DTLZ tasks. Error bars show standard deviations. Complete results are in Appendix~\ref{app:exp_complete_tables}.}
\label{fig:synthetic_rank_tradeoff_rewrite}
\vspace{-0.3cm}
\end{minipage}

\end{figure}

We fix $\vartheta = 1$ and vary $\kappa\in\{0,0.5,1,1.5,2,2.5,3\}$ to examine increasingly distorted PFs. As shown in Figure~\ref{fig:controlled_radial_distortion}, with $N=63$, $\alpha=0.2$, and $T=5$ rounds, the measured distortion ratio $\overline J/\underline J$ increases from $1$ to $299.53$.
At $\kappa=3$, Uniform weighting has measured residual $\widehat{\mathcal R}_\vartheta=8.901$, whereas SURFACE remains below $0.785$ throughout the sweep.
For reference, SURFACE (Exact $J$) replaces the reconstructed area density with the analytic $J$ while keeping the same transport procedure and has residual $0.0337$ at $\kappa=3$, {providing an analytic-density reference.} 
Figure~\ref{fig:controlled_rates} further shows that the finite-resolution errors decrease with $N$ at rates consistent with the theoretical $\Oc(N^{-1})$ for $M=3$ and $\Oc(N^{-2/5})$ for $M=6$ in Theorem~\ref{thm:surface_inexact}.

\begin{table}[t]
\centering
\scriptsize
\renewcommand{\arraystretch}{1.02}
\resizebox{\linewidth}{!}{%
\begin{tabular}{@{}l|r|r|r|r|r|r|r@{}}
\toprule
Metric
& SMS-EMOA
& NSGA-III
& LMPFE
& Subset
& UMOD
& Uniform
& SURFACE \\

\midrule
\multicolumn{8}{@{}l}{\textit{ZDT2, $M=2$, $N=9$}} \\
$\hat{\mathcal R}_{\vartheta}\downarrow$ & $.480 \pm .018$ & $.589 \pm .000$ & $.881 \pm .374$ & $.491 \pm .000$ & $.812 \pm .047$ & $.514$ & $\mathbf{.397}$\\
$\mathrm{IGD}\downarrow$ & $.039 \pm .000$ & $.038 \pm .000$ & $.048 \pm .013$ & $.039 \pm .000$ & $.041 \pm .000$ & $.037$ & $\mathbf{.036}$\\
Fill$\downarrow$ & $.126 \pm .003$ & $.089 \pm .000$ & $.156 \pm .068$ & $.122 \pm .000$ & $.101 \pm .003$ & $.087$ & $\mathbf{.073}$\\
$\Delta\mathrm{HV}\downarrow$ & $\mathbf{.069 \pm .000}$ & $.076 \pm .000$ & $.097 \pm .025$ & $.075 \pm .000$ & $.084 \pm .001$ & $.078$ & $.074$\\

\midrule
\multicolumn{8}{@{}l}{\textit{ZDT4, $M=2$, $N=9$}} \\
$\hat{\mathcal R}_{\vartheta}\downarrow$ & $.418 \pm .006$ & $.611 \pm .003$ & $.572 \pm .033$ & $.760 \pm .005$ & $2.124 \pm .972$ & $.505$ & $\mathbf{.316}$\\
$\mathrm{IGD}\downarrow$ & $.037 \pm .000$ & $.043 \pm .000$ & $.038 \pm .000$ & $.040 \pm .000$ & $.108 \pm .060$ & $.037$ & $\mathbf{.035}$\\
Fill$\downarrow$ & $.090 \pm .001$ & $.150 \pm .000$ & $\mathbf{.088 \pm .004}$ & $.173 \pm .001$ & $.324 \pm .155$ & $.106$ & $.096$\\
$\Delta\mathrm{HV}\downarrow$ & $.049 \pm .000$ & $.056 \pm .001$ & $.053 \pm .002$ & $.050 \pm .000$ & $.196 \pm .122$ & $.048$ & $\mathbf{.045}$\\

\midrule
\multicolumn{8}{@{}l}{\textit{DTLZ3, $M=2$, $N=9$}} \\
$\hat{\mathcal R}_{\vartheta}\downarrow$ & $.509 \pm .017$ & $.616 \pm .006$ & $.492 \pm .016$ & $.624 \pm .001$ & $1.471 \pm .399$ & $.543$ & $\mathbf{.402}$\\
$\mathrm{IGD}\downarrow$ & $.046 \pm .000$ & $.041 \pm .000$ & $.041 \pm .001$ & $.048 \pm .001$ & $.085 \pm .022$ & $.040$ & $\mathbf{.038}$\\
Fill$\downarrow$ & $.143 \pm .001$ & $.099 \pm .000$ & $.096 \pm .006$ & $.148 \pm .001$ & $.233 \pm .052$ & $.097$ & $\mathbf{.078}$\\
$\Delta\mathrm{HV}\downarrow$ & $\mathbf{.070 \pm .006}$ & $.088 \pm .003$ & $.092 \pm .007$ & $.079 \pm .013$ & $.289 \pm .113$ & $.085$ & $.074$\\

\midrule
\multicolumn{8}{@{}l}{\textit{DTLZ3, $M=3$, $N=18$}} \\
$\hat{\mathcal R}_{\vartheta}\downarrow$ & $1.490 \pm .402$ & $1.179 \pm .001$ & $2.546 \pm 1.447$ & $1.588 \pm .001$ & $3.120 \pm .202$ & $1.078$ & $\mathbf{.693}$\\
$\mathrm{IGD}\downarrow$ & $.166 \pm .005$ & $.137 \pm .000$ & $.210 \pm .062$ & $.162 \pm .000$ & $.204 \pm .022$ & $.132$ & $\mathbf{.129}$\\
Fill$\downarrow$ & $.342 \pm .020$ & $.272 \pm .000$ & $.479 \pm .137$ & $.333 \pm .000$ & $.476 \pm .050$ & $.265$ & $\mathbf{.227}$\\
$\Delta\mathrm{HV}\downarrow$ & $\mathbf{.134 \pm .001}$ & $.162 \pm .000$ & $.258 \pm .071$ & $.139 \pm .002$ & $.266 \pm .021$ & $.175$ & $.162$\\

\midrule
\multicolumn{8}{@{}l}{\textit{DTLZ4, $M=3$, $N=18$}} \\
$\hat{\mathcal R}_{\vartheta}\downarrow$ & $20.000 \pm .000$ & $20.000 \pm .000$ & $4.698 \pm 8.556$ & $8.954 \pm 10.083$ & $20.000 \pm .000$ & $1.078$ & $\mathbf{.693}$\\
$\mathrm{IGD}\downarrow$ & $.943 \pm .000$ & $.943 \pm .000$ & $.296 \pm .361$ & $.475 \pm .427$ & $.943 \pm .000$ & $.132$ & $\mathbf{.129}$\\
Fill$\downarrow$ & $1.414 \pm .000$ & $1.414 \pm .000$ & $.523 \pm .498$ & $.765 \pm .592$ & $1.414 \pm .000$ & $.265$ & $\mathbf{.227}$\\
$\Delta\mathrm{HV}\downarrow$ & $.848 \pm .000$ & $.848 \pm .000$ & $.307 \pm .302$ & $.422 \pm .389$ & $.848 \pm .000$ & $.175$ & $\mathbf{.162}$\\

\bottomrule
\end{tabular}%
}
\vspace{-0.3cm}
\caption{Benchmark results for representative baselines. Entries for stochastic baselines are mean $\pm$ standard deviation over five seeds. SURFACE and Uniform are deterministic. Complete results for all ten methods are reported in Appendix~\ref{app:exp_complete_tables}.}
\label{tab:synthetic_main}
\end{table}

\subsection{Uniform Pareto-Front Coverage on ZDT and DTLZ Benchmarks}
\label{sec:exp_finite_coverage}
We further evaluate SURFACE on ZDT1/2/4 and DTLZ2/3/4~\citep{zitzler2000comparison,deb2002scalable} with the uniform PF coverage preference $\vartheta = 1$. We compare SURFACE with \emph{Uniform weighting}~\citep{zhang2007moead}, \emph{UMOD}~\citep{zhang2024umod}, \emph{LMPFE}~\citep{tian2022local}, \emph{DEA-GNG}~\citep{liu2019adapting}, \emph{Subset selection}~\citep{shang2023benchmarking}, \emph{NSGA-III}~\citep{deb2014evolutionary}, \emph{SMS-EMOA}~\citep{beume2007smsemoa}, \emph{MOEA/D}~\citep{zhang2007moead}, and \emph{MOEA/D-AWA}~\citep{qi2014moea}, with the same number of returned solutions. SURFACE and Uniform use the same scalarized optimization procedure and differ only in their scalarization weights, allowing us to isolate the effect of adaptive weight redistribution.
We evaluate the remaining baselines end-to-end, including their optimization and solution-selection procedures. We use target-mass residuals as the primary distribution-matching metrics, together with area-weighted IGD~\citep{ishibuchi2015study}, fill distance~\citep{zhang2024umod}, and hypervolume~\citep{beume2007smsemoa,shang2020survey}.

\begin{figure}[t]
\centering
\includegraphics[width=0.24\linewidth]{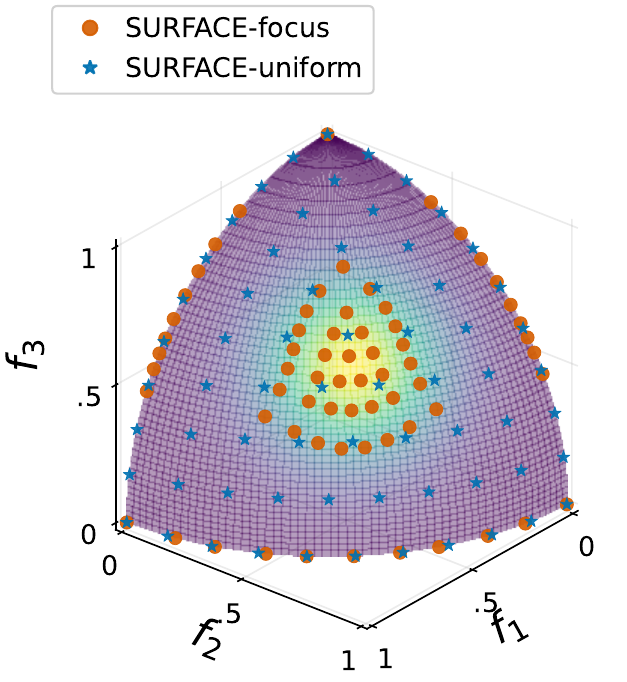}
\includegraphics[width=0.235\linewidth]{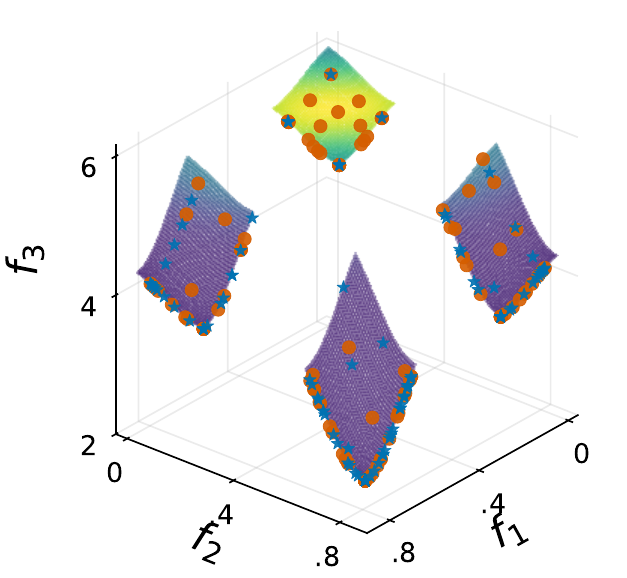}
\includegraphics[width=0.215\linewidth]{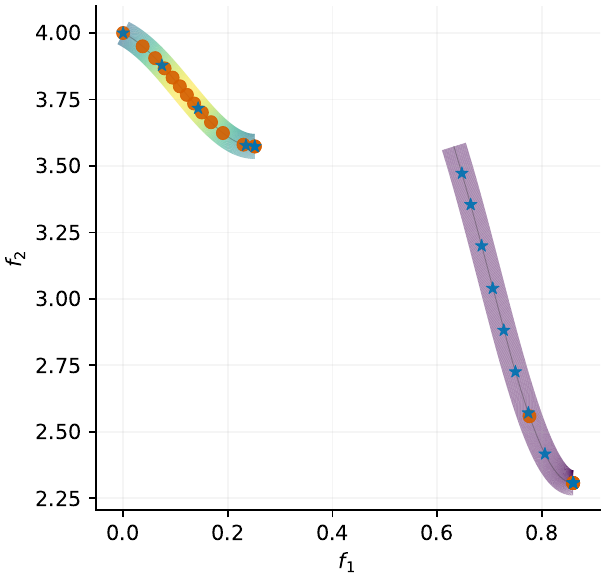}
\includegraphics[width=0.28\linewidth]{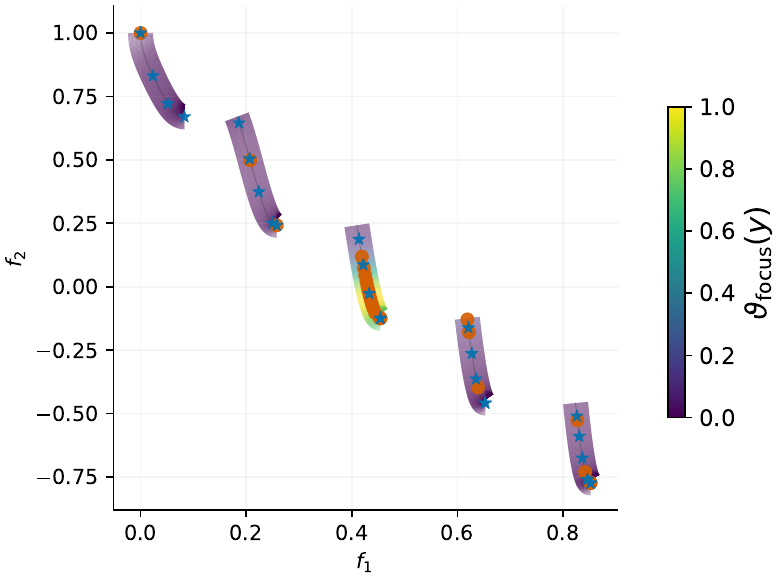}
\vspace{-0.2cm}
\caption{SURFACE with $\vartheta= 1,\vartheta_{\rm focus}$. From left to right: DTLZ2 ($M=3$, $N=66$), DTLZ7 ($M=3$, $N=153$), DTLZ7 ($M=2$, $N=20$), and ZDT3 ($M=2$, $N=25$). Warmer PF colors indicate higher $\vartheta_{\rm focus}$ on a shared $0$--$1$ scale.
}
\label{fig:appendix_preference_allocation}
\end{figure}

Table~\ref{tab:synthetic_main} reports a preview of the results, while complete results for all ten methods are provided in Appendix~\ref{app:exp_complete_tables}.
Across the 11 settings, SURFACE achieves the best mean rank, as visualized in Figure~\ref{fig:synthetic_rank_tradeoff_rewrite}, where the coverage metric averages the first three PF-representation metrics. Compared with Uniform, SURFACE improves all four metrics in every primary comparison, with median reductions of $35.7\%$ in PF-area imbalance, $3.64\%$ in IGD, $14.2\%$ in fill distance, and $7.21\%$ in normalized hypervolume gap. These results show that adapting the scalarization weights to the PF geometry produces a more uniform finite representation while maintaining competitive objective-space quality. 
We further examine how SURFACE allocates solutions under uniform and focused preferences on connected and disconnected PFs. As shown in Figure~\ref{fig:appendix_preference_allocation}, SURFACE with $\vartheta\equiv1$ distributes solutions across the PF, while $\vartheta_{\rm focus}$ concentrates them in preferred regions. On DTLZ2, the focused preference emphasizes balanced tradeoffs, while on disconnected DTLZ7 ($M=2,3$) and ZDT3 fronts, it increases resolution in preferred regions while retaining broader PF coverage. These disconnected-front results extend the empirical evaluation beyond the smooth, connected PF charts covered by our theoretical guarantees.

\begin{table}[t]
\centering
\scriptsize
\renewcommand{\arraystretch}{1.06}
\resizebox{\linewidth}{!}{%
\begin{tabular}{@{}l|r|r|r|r|r|r|r@{}}
\toprule
Metric
& SMS-EMOA
& NSGA-III
& MOEA/D
& MOEA/D-AWA
& UMOD
& Uniform
& SURFACE \\
\midrule

\multicolumn{8}{@{}l}{\textit{Deep Sea Treasure}, $M=2$, $N=14$} \\
$\widehat{\mathcal R}_{\vartheta}\downarrow$
& $.943\pm.379$ & $.838\pm.279$ & $1.296\pm.201$ & $1.156\pm.299$ & $1.138\pm.371$ & $1.069$ & $\mathbf{.249}$\\
$\mathrm{IGD}\downarrow$
& $.346\pm.038$ & $.335\pm.030$ & $.416\pm.056$ & $.387\pm.067$ & $.383\pm.071$ & $.077$ & $\mathbf{.026}$\\
Fill$\downarrow$
& $.822\pm.086$ & $.797\pm.072$ & $.945\pm.079$ & $.895\pm.104$ & $.883\pm.120$ & $.255$ & $\mathbf{.105}$\\
HV$\uparrow$
& $1.031\pm.178$ & $\mathbf{1.093\pm.142}$ & $.812\pm.131$ & $.909\pm.187$ & $.911\pm.208$ & $1.041$ & $1.043$\\

\midrule
\multicolumn{8}{@{}l}{\textit{Resource Gathering}, $M=3$, $N=18$} \\
$\widehat{\mathcal R}_{\vartheta}\downarrow$
& $1.806\pm.055$ & $1.848\pm.052$ & $1.830\pm.042$ & $1.783\pm.117$ & $1.848\pm.053$ & $.454$ & $\mathbf{.392}$\\
$\mathrm{IGD}\downarrow$
& $2.808\pm.054$ & $1.961\pm.732$ & $4.722\pm1.119$ & $1.216\pm.245$ & $4.447\pm1.846$ & $.095$ & $\mathbf{.083}$\\
Fill$\downarrow$
& $3.288\pm.037$ & $2.377\pm.800$ & $5.289\pm1.236$ & $1.623\pm.249$ & $4.956\pm1.891$ & $.200$ & $\mathbf{.154}$\\
HV$\uparrow$
& $1.174\pm.243$ & $.951\pm.331$ & $1.057\pm.154$ & $\mathbf{1.279\pm.084}$ & $1.150\pm.250$ & $.114$ & $.114$\\

\midrule
\multicolumn{8}{@{}l}{\textit{Fruit Tree}, $M=6$, $N=50$} \\
$\widehat{\mathcal R}_{\vartheta}\downarrow$
& $1.485\pm.141$ & $1.258\pm.047$ & $.956\pm.028$ & $1.138\pm.061$ & $1.254\pm.087$ & $1.023$ & $\mathbf{.943}$\\
$\mathrm{IGD}\downarrow$
& $.377\pm.032$ & $.262\pm.014$ & $.263\pm.010$ & $.284\pm.017$ & $.307\pm.015$ & $.255$ & $\mathbf{.249}$\\
Fill$\downarrow$
& $.642\pm.045$ & $.662\pm.008$ & $.787\pm.092$ & $.556\pm.141$ & $.637\pm.064$ & $.392$ & $\mathbf{.381}$\\
HV$\uparrow$
& $.035\pm.006$ & $.069\pm.002$ & $\mathbf{.078\pm.002}$ & $.078\pm.002$ & $.059\pm.005$ & $.068$ & $.069$\\
\bottomrule
\end{tabular}%
}
\vspace{-0.3cm}
\caption{MO-Gym policy results. Stochastic entries are mean $\pm$ sample standard deviation over five seeds. Uniform and SURFACE are deterministic. Bold marks the best mean in each row.}
\label{tab:mogym_combined_rewrite}
\end{table}

\subsection{Multi-Objective Policy Optimization}
\label{sec:exp_policy}

We additionally evaluate SURFACE's compatibility beyond TS by implementing it with weighted-sum scalarization on Deep Sea Treasure, Resource Gathering, and Fruit Tree from MO-Gymnasium~\citep{felten2023toolkit}. 
Deep Sea Treasure ($M=2$) is a navigation task balancing treasure value and travel time, while Resource Gathering ($M=3$) balances gold and gem collection against enemy risk. Fruit Tree ($M=6$) is a tree-navigation task in which different fruits provide different nutritional rewards.
We optimize weighted sums of expected rewards with KL regularization~\citep{roijers2013survey,jiang2026surf}, with theoretical details in Appendix~\ref{sec:linear_specialization}.
With the uniform PF target $\vartheta\equiv1$, we compare SURFACE with Uniform using the same inner solver. We match the solver budget between the two methods. The remaining baselines use their standard evolutionary policy optimizers. As shown in Table~\ref{tab:mogym_combined_rewrite}, SURFACE consistently improves both PF-area residuals and area-weighted IGD over Uniform on all three environments while maintaining comparable hypervolume. Experimental settings, additional resolution studies, and policy diagnostics are provided in Appendix~\ref{app:exp_policy_details}.

\section{Conclusion}\label{sec:conclusion}

This paper introduced SURFACE, a geometry-aware framework for matching a prescribed preference distribution on the PF. The framework estimates the weight-to-PF chart from finitely many scalarized solutions and redistributes scalarization weights via mass equidistribution. Theoretically, we showed that Tchebycheff scalarization induces a smooth, nonfolding, and nondegenerate PF chart under the stated regularity conditions. We also proved that SURFACE contracts the coupled density–solver error geometrically to an $\mathcal{O}(N^{-2/(M-1)})$ finite-resolution floor. The same rate holds for the target-mass imbalance across complete represented PF regions. Experiments across ZDT/DTLZ benchmarks and MO-Gymnasium policy tasks against nine baselines confirm consistent gains in preference matching and Pareto representation quality. SURFACE is compatible with off-the-shelf solvers and inexact warm-started optimization. A natural future direction is to mitigate the rapid growth of the required number of adaptive weights $N$ with the number of objectives $M$. This could be achieved through adaptive sparse weight placement or hierarchical mesh refinement.

\section*{AI Use Statement}
We used generative AI tools to help polish the writing and to double-check parts of the mathematical proofs. All final arguments and proofs were reviewed and verified by the authors.

\section*{Reproducibility Statement}
Implementation details for SURFACE and the experimental settings are provided in Appendix~\ref{app:benchmarks}. The code is provided in the supplementary material.

\section*{Acknowledgements}

This work is partially supported by the Center of Excellence (CoE) in Data Science and Artificial Intelligence at the University of Rochester. We also acknowledge the Center for Integrated Research Computing (CIRC) at the University of Rochester and Empire AI for providing computational resources and support.

\bibliography{main}
\bibliographystyle{iclr2027_conference}

\newpage
\appendix

\section*{Appendix Contents}
\vspace{-0.8ex}
\hrule
\vspace{1.0ex}

{\small
\newcommand{\appsectionline}[2]{%
  \noindent
  \makebox[1.7em][l]{\textbf{\ref{#1}}}%
  \textbf{#2}%
  \leaders\hbox to 0.35em{\hss.\hss}\hfill
  \textbf{\pageref{#1}}\par}

\newcommand{\appsubline}[2]{%
  \noindent
  \hspace*{0.04\linewidth}%
  \makebox[2.9em][l]{\ref{#1}}%
  #2%
  \leaders\hbox to 0.35em{\hss.\hss}\hfill
  \pageref{#1}\par}

\newcommand{\appsubsubline}[2]{%
  \noindent
  \hspace*{0.08\linewidth}%
  \makebox[3.7em][l]{\ref{#1}}%
  #2%
  \leaders\hbox to 0.35em{\hss.\hss}\hfill
  \pageref{#1}\par}

\appsectionline{app:background_notation}
{Related Work and Notation}
\appsubline{app:related_work}
{Detailed Related Work}
\appsubline{app:notation}
{Notation}

\appsectionline{app:geometry_foundations}
{Proofs for Pareto-Front Geometry}
\appsubline{app:prelim}
{Proof of Lemma~\ref{lem:preference_pullback}}
\appsubline{app:proof_pf_geometry}
{Proof of Proposition~\ref{prop:pf_geometry}}
\appsubline{app:tch_sufficient_regimes}
{Sufficient Regular Regimes}
\appsubline{app:proof_linear_geometry}
{Weighted-Sum Specialization}

\appsectionline{app:proofs_theory}
{Details for the SURFACE Algorithm}
\appsubline{app:reconstruction_proof}
{Reconstructing the PF Between Sampled Weights}
\appsubline{app:finite_meq}
{Finite MEQ: Mass Accuracy and Mesh Stability}

\appsectionline{app:proofs_convergence}
{Proofs for the Convergence Analysis}
\appsubline{app:proofs_exact}
{Exact Scalarized Solves}
\appsubline{app:proofs_inexact}
{Inexact Scalarized Solves}

\appsectionline{app:benchmarks}
{Experiment Details}
\appsubline{app:exp_protocol}
{Experimental Protocol and Benchmark Scope}
\appsubline{app:exp_dtlz2_details}
{Controlled Validation}
\appsubline{app:exp_complete_tables}
{Details of ZDT and DTLZ Benchmark Results}
\appsubline{app:exp_policy_details}
{Multi-Objective Policy Optimization}
}
\section{Related Work and Notation}\label{app:background_notation}

This section gives further comparisons with related methods and collects the notation used in the analysis.

\subsection{Detailed Related Work}\label{app:related_work}

\paragraph{Preference specification and Pareto-set learning.}

Preference information in MOO is commonly incorporated either \emph{a priori}, before optimization, or \emph{a posteriori}, after constructing a Pareto representation~\citep{pajasmaa2025group}. A priori methods use specified preferences to target preferred Pareto solutions or regions~\citep{chen2024ferero,chen2025efficient}. Preference-conditioned learning extends this paradigm by learning a map from preferences to Pareto solutions. Pareto HyperNetworks~\citep{navon2021learning} generate models conditioned on preference vectors, while Pareto Set Learning~\citep{lin2022pareto} learns a continuous preference-to-solution map for expensive MOO. Evolutionary Preference Sampling~\citep{ye2024eps} adapts the preference distribution used for training, and shape-agnostic Pareto Set Learning~\citep{ye2024pareto} formulates Pareto-set learning as distribution transformation with a hypervolume objective. A posteriori methods instead construct a finite PF approximation for subsequent selection~\citep{zhang2024umod,boehmer2026picking}. SURFACE connects these two paradigms: it specifies the desired allocation \emph{a priori} through a probability distribution over the PF and returns an \emph{a posteriori} finite representation whose resolution follows that distribution.

\paragraph{Scalarization and geometry-based PF coverage.}

Scalarization generates Pareto solutions by varying preference parameters. Weighted-sum scalarization~\citep{marler2010weighted} covers the full PF under suitable convexity conditions but can miss Pareto solutions on nonconvex fronts~\citep{miettinen1999nonlinear,ehrgott2005multicriteria}. Tchebycheff scalarization~\citep{miettinen1999nonlinear,ehrgott2005multicriteria}, achievement scalarization~\citep{ehrgott2005multicriteria}, and Pascoletti--Serafini scalarization~\citep{eichfelder2009adaptive} provide alternative parameterizations under their respective conditions. Standard scalarization determines which Pareto solution corresponds to a parameter but does not control the distribution of a finite parameter set over the PF. In particular, uniformly spaced parameters can produce nonuniformly spaced Pareto points~\citep{das1998normal,zhang2006min,jiang2026surf}. We therefore use uniform scalarization weights as a baseline for isolating the effect of SURFACE's redistribution.

Several methods adapt scalarization parameters or objective-space constructions to improve PF coverage. \citet{eichfelder2009adaptive} uses local sensitivity information to adapt Pascoletti--Serafini parameters toward nearly equidistant objective-space points. \citet{zhang2006min} adjusts min--max weights to generate more uniformly spaced Pareto points, while \citet{saini2025efficient} removes directions in densely represented regions and adds directions in sparse regions. Normal-boundary intersection~\citep{das1998normal} generates solutions along normal directions from reference points on the convex hull of the individual minima. The normalized normal-constraint method~\citep{messac2003normalized} normalizes the objectives and generates solutions through moving objective-space constraints. Directed Search Domain methods~\citep{erfani2011directed} sequentially restrict objective-space regions to place new solutions in sparsely represented areas, with a higher-dimensional extension given by \citet{yu2024extension}.
These methods primarily target uniform or well-spread representations. Their updates use local sensitivity information~\citep{eichfelder2009adaptive}, sequential parameter selection~\citep{zhang2006min,saini2025efficient,erfani2011directed}, or additional objective-space directions and constraints~\citep{das1998normal,messac2003normalized}. SURFACE uses the weight--objective pairs from a parallel batch of scalarized solves to estimate the PF geometry. It then adjusts the weights for a prescribed PF distribution while retaining the chosen scalarization and inner solver.

\paragraph{Finite Pareto representations: tracing, diversity, and design.}
Continuation methods construct finite PF representations by following local Pareto geometry. \citet{hillermeier2001generalized} models Pareto candidates as a differentiable manifold and uses generalized homotopy to discretize it evenly. \citet{pereyra2009fast,pereyra2013equispaced} incorporate spacing conditions into continuation procedures, including objective-space arc-length spacing for constrained bi-objective problems. \citet{martin2016continuation} follows bi-objective branches with interval validation, \citet{bolten2021tracing} traces locally Pareto-optimal solutions through numerical integration of sensitivity information, and \citet{beltran2020pareto} extends Pareto tracing to inequality-constrained problems. These methods advance sequentially from previously computed solutions using local derivative, tangent, or sensitivity information~\citep{hillermeier2001generalized,pereyra2013equispaced,bolten2021tracing}. SURFACE also uses local PF geometry to place solutions, but reconstructs this geometry from parallel scalarized solves.
Evolutionary decomposition and reference-vector methods construct finite representations by controlling search directions or reference points. MOEA/D~\citep{zhang2007moead} decomposes MOO into scalar subproblems associated with a fixed set of weights. MOEA/D-AWA~\citep{qi2014moea} removes weights associated with crowded or duplicate solutions and adds weights in sparse PF regions. DEA-GNG~\citep{liu2019adapting} adapts preference vectors using a growing neural gas representation of the evolving PF. NSGA-III~\citep{deb2014evolutionary} maintains diversity through reference-point association, while RVEA~\citep{cheng2016reference} combines angle-penalized distance with adaptive reference vectors. These methods allocate solutions using sparsity, occupancy, learned PF structure, or angular diversity.

Other methods define representativeness through an explicit finite-set objective. LMPFE~\citep{tian2022local} approximates the PF using multiple local models, while fixed-cardinality subset selection~\citep{shang2023benchmarking} selects a representative subset from a larger candidate set. UMOD~\citep{zhang2024umod} measures representativeness by fill distance and optimizes a max-packing surrogate with a constant-factor fill-distance guarantee. \citet{borghi2023repulsion} uses binary repulsive potentials for bi- and tri-objective PFs, while \citet{borghi2023adaptive} combines gradient-free search with energy-based parameter dynamics. Riesz energy has been used to generate arbitrary-cardinality reference sets on the unit simplex~\citep{blank2020generating} and to optimize point configurations directly on nonlinear PFs~\citep{nezhadmoghaddam2026approximating}. These methods define the desired representation through approximation, covering, packing, subset selection, or repulsion.

Set-quality indicators provide another, implicit specification of the desired finite representation. Classical indicators combine convergence and spread in different ways~\citep{zitzler2000comparison}. Crowding distance~\citep{deb2002fast} favors locally isolated solutions. IGD~\citep{ishibuchi2015study} measures distances to a chosen reference set, while IGD+~\citep{ishibuchi2015study} modifies this distance to satisfy weak Pareto compliance. Hypervolume~\citep{beume2007smsemoa} measures dominated objective-space volume relative to a reference point, and its optimal fixed-cardinality distribution depends on the PF geometry and reference point~\citep{shang2020survey}. SMS-EMOA~\citep{beume2007smsemoa} selects solutions by hypervolume contribution, while indicator-based subset selection~\citep{shang2023benchmarking} retains the finite set preferred by the selected criterion. 
These indicators induce an allocation through a reference set, reference point, or diversity criterion.
These approaches specify a finite representation through search directions, geometric spacing, or a set-quality objective. SURFACE takes a target PF distribution as input and balances its probability mass across the represented cells. A constant preference density gives uniform PF surface-area allocation.

\paragraph{Mesh redistribution for distribution matching.}

Moving-mesh and geometric-transport methods motivate SURFACE's redistribution step.
Adaptive moving meshes~\citep{huang2010adaptive,budd2009moving,xu2011convergence} distribute a prescribed monitor density equally among cells while controlling mesh quality. The monitor density determines where mesh points concentrate. Prescribed-Jacobian transport~\citep{dacorogna1990jacobian} controls the local volume change of a map to transform one density into another. Moser-type transport on domains with corners~\citep{bruveris2018moser} constructs smooth maps between positive densities. Centroidal Voronoi methods~\citep{du2003constrained} and capacity-constrained sampling~\citep{zhang2016capacity} provide related ways to distribute a finite set of points, called sites. In SURFACE, these sites are scalarization weights. The prescribed PF preference and the weight-to-PF geometry determine the density to redistribute. For two objectives and a constant preference profile, the construction reduces to SURF's chord-length fraction and inverse-CDF update~\citep{jiang2026surf}.
SURF addresses uniform allocation along a one-dimensional PF, while SURFACE supports nonconstant PF distributions and higher-dimensional PFs through face-preserving transport.
For two objectives, a nonconstant profile gives the corresponding preference-weighted arc-length construction.

\subsection{Notation}\label{app:notation}

We use $\|u\|_{C^{k,1}}:=\|u\|_{C^k}+\operatorname{Lip}(D^k u)$, interpreted face-wise where needed. Table~\ref{tab:notation} collects symbols that recur across sections. The last column points to each symbol's main definition or explanation. Each proof introduces its local notation.

\begin{table}[t]
\centering
\caption{Recurring notation guide.}
\label{tab:notation}
\footnotesize
{
\newcommand{\notepage}[1]{\leavevmode\leaders\hbox to 0.35em{\hss.\hss}\hfill p.~\pageref{#1}}
\renewcommand{\arraystretch}{1.12}
\begin{tabular}{@{}p{0.24\linewidth}p{0.56\linewidth}p{0.14\linewidth}@{}}
\toprule
\textbf{Symbol} & \textbf{Meaning} & \textbf{Definition}\\
\midrule

\rowcolor{gray!10}
\multicolumn{3}{@{}l}{\textit{Problem and PF chart}}\\
$M$ & Number of objectives. & \notepage{eq:moo_problem}\\
$\bx$, $\X$ & Decision variable and feasible set. & \notepage{eq:moo_problem}\\
$\f=(f_1,\ldots,f_M)$ & Vector-valued objective. & \notepage{eq:moo_problem}\\
$\bz$ & Utopian Tchebycheff reference, with $z_m<\min_{\bx\in\X}f_m(\bx)$. & \notepage{eq:standard_tchebycheff}\\
$\bw$ & Scalarization weight. & \notepage{eq:standard_tchebycheff}\\

$\bx^\star(\bw)$, $\bx_{\rm TS}^\star(\bw)$ & Generic scalarized optimizer and its Tchebycheff specialization. & \notepage{eq:weight_pf_map}\\
$\fPF(\bw)$, $\FPF$ & Scalarization-induced PF map and Pareto front. & \notepage{eq:weight_pf_map}\\
$\Delta_M$, $\Delta_M^\delta$, $\delta$ & Weight simplex, truncated TS domain, and truncation margin $0<\delta<1/M$. & \notepage{prop:pf_geometry}\\

\addlinespace[0.4ex]
\rowcolor{gray!10}
\multicolumn{3}{@{}l}{\textit{PF target distribution and finite representation}}\\
$J(\bw)$ & PF-area Jacobian factor of the chart $\fPF$. & \notepage{eq:pf_area_density}\\

$\vartheta$, $\rho_\vartheta$ & PF preference density and its normalized pullback to weight space. & \notepage{eq:preference_pullback_density}\\

$W$, $W^{(t)}$, $\mathcal C(W)$ & Weight mesh including fixed corners, its round-$t$ instance, and associated cell partition. & \notepage{lem:preference_pullback}\\
$\mathcal T(W)$ & Straight simplicial reconstruction mesh. & \notepage{app:proofs_theory}\\

$\widehat{\mathcal R}_\vartheta(W)$ & Maximum relative preference-mass imbalance over $\mathcal C(W)$. & \notepage{lem:preference_pullback}\\
$N$ & Number of adaptive weights in SURFACE. & \notepage{alg:surface}\\

\addlinespace[0.4ex]
\rowcolor{gray!10}
\multicolumn{3}{@{}l}{\textit{SURFACE and inexact solves}}\\
$P_t$ & Reconstructed PF chart at outer round $t$. & \notepage{alg:surface}\\
$\rho_t$, $\alpha$ & Outer-round weight density and damping parameter. & \notepage{eq:density_update}\\
$\mathcal E_t$ & Coupled density--solver error at outer round $t$. & \notepage{eq:coupled_metric}\\

\bottomrule
\end{tabular}
}
\end{table}

\subsection{Practical Implementation of SURFACE}
\label{app:practical_surface}

We describe how to implement one outer round of Algorithm~\ref{alg:surface}.
SURFACE requires only a scalarized inner solver, returned objective vectors,
and a prescribed preference density. Reconstruction and redistribution operate
in the $(M-1)$-dimensional weight domain, without differentiating through the
inner optimization.

\paragraph{Initialization.}
Choose a scalarization, a positive preference profile $\vartheta$, the
number $N$ of adaptive weights, the damping $\alpha$, and the number $T$ of
outer rounds. For the regular TS construction, use the weight domain
$\Delta_M^\delta$ with its $M$ fixed chart corners $\{\bw_m^\delta\}_{m=1}^M$.
Construct a shape-regular reference triangulation with $N$ adaptive vertices
and fixed corners, retaining its connectivity across rounds. Initialize
$\rho_0$ as the uniform density on this domain and assign an initial state to
each scalarized solver. Obtain objective values at the fixed chart corners once,
or refine those solutions as needed. These $M$ corner samples are separate from
the $N$ adaptive solutions and are included in the inner-solve accounting if
not supplied.

\paragraph{Scalarized solves and PF reconstruction.}
At round $t$, use $\operatorname{MEQ}_N(\rho_t)$ to obtain the current adaptive
weights $\{\bw_n^{(t)}\}_{n=1}^N$. Solve the associated scalarized problems
independently, warm-starting the $n$th solve from
$\tilde\bx_n^{(t-1)}$ as in~\eqref{eq:surface_inner_solve}. These solves may be
run in parallel. Collect the weight--objective pairs
$\{(\bw_n^{(t)},\f(\tilde\bx_n^{(t)}))\}_{n=1}^N$ together with the fixed
chart-corner pairs. To construct $P_t$, fit a quadratically augmented,
compactly supported Wendland RBF model to each local patch, separately for
each objective coordinate. In a patch, the fitted function has the form
\begin{align}
P_{t,\mathrm{loc}}(\bw)
= q(\bw)+\sum_j c_j\,
\phi\!\left(\frac{\|\bw-\bw_j\|}{r}\right),
\label{eq:practical_local_rbf}
\end{align}
where $q$ is quadratic, $\phi$ is the Wendland kernel, and $r$ is a local
support radius. The coefficients are determined by interpolation at the
neighboring samples with the usual polynomial side constraints. Smooth
partition-of-unity weights blend these overlapping fits into $P_t$.
This gives a differentiable chart between sampled weights and permits
computing $DP_t$ without differentiating the inner solver. Appendix~\ref{app:reconstruction_proof}
specifies the patch and stability conditions.

\paragraph{Density estimation.}
Evaluate $P_t$ and its derivatives at quadrature nodes in the weight domain.
Using the same $M-1$ independent weight coordinates as in
Eq.~\eqref{eq:pf_area_density}, compute
\begin{align}
J_{P_t}(\bw)
=\tfrac{1}{\sqrt M}
\sqrt{\det\!\bigl(DP_t(\bw)^\top DP_t(\bw)\bigr)}.
\label{eq:practical_area_factor}
\end{align}
Evaluate the prescribed preference on the reconstructed objective vectors,
$\vartheta(P_t(\bw))$, and normalize the product
$\vartheta(P_t)J_{P_t}$ by numerical integration over $\Delta_M^\delta$.
Blend this normalized density with $\rho_t$ using Eq.~\eqref{eq:density_update}
to obtain $\rho_{t+1}$. The preference is evaluated through its extension
to a neighborhood of the PF, as in Assumption~\ref{ass:preference_profile}.

\paragraph{Mass equidistribution.}
Use $\rho_{t+1}$ to generate the next adaptive weights. The goal is to make
the associated \emph{complete cells} carry equal probability mass, as in
Eq.~\eqref{eq:meq_target}. For $M=2$, apply inverse-CDF transport to the
reference sites. For $M\ge3$, Appendix~\ref{app:finite_meq} gives two
realizations. The Moser realization integrates a face-preserving flow that
transports the equal-volume reference partition toward the target density.
The equal-capacity realization first selects density-dependent sites, then
balances their power-cell masses by optimizing the dual offsets in
Eq.~\eqref{eq:capacity_candidate_dual}; optional site refinement is restricted
as in Eq.~\eqref{eq:capacity_site_min}. In both cases, the \emph{straight
simplicial mesh} $\mathcal T(W)$, constructed from the resulting sites and
retained connectivity, is used for reconstruction, whereas the complete cells
$\mathcal C(W)$ are used to assess mass balance. The $M$ chart corners remain
fixed. Repeat the scalarized solves at the updated weights in the next round.

\paragraph{Inner-solver interface and reproducibility.}
The inner solver takes a weight, an initialization, and a computational budget
or stopping tolerance, and returns an updated decision state and its objective
vector. For differentiable models it can perform $K$ steps of a standard
optimizer on the scalarized loss; for a finite MDP it can use a convergent
policy solver. Warm starts follow the persistent vertex labels of the mesh.
For reproducibility, report the weight domain and corner samples, $N$, $T$,
$\alpha$, the inner-solver stopping rule, the reconstruction stencil and kernel,
quadrature accuracy, the MEQ realization and tolerances, and measured runtime.
Theoretical guarantees for inexact solves require the solver-accuracy
conditions of Appendix~\ref{app:proofs_inexact}; the implementation interface
itself is agnostic to the chosen optimizer.

\section{Proofs for Pareto-Front Geometry}
\label{app:geometry_foundations}

The geometric analysis has two parts. The area formula relates probability mass in weight space to preference-weighted PF area. The scalarization results establish the smoothness and injectivity needed to use this relation for TS and weighted sums.

\subsection{Proof of Lemma~\ref{lem:preference_pullback}}
\label{app:prelim}

We measure PF surface area using the $(M{-}1)$-dimensional Hausdorff measure $\mathcal H^{M-1}$. For every measurable $\tau\subseteq\Delta_M$, the area formula gives
\begin{align}
\mathcal H^{M-1}\left(\fPF(\tau)\right)
=
\int_\tau J(\bw)\,\mathrm d\operatorname{Vol}_{M-1}(\bw).
\end{align}
Accordingly, $\mathrm d\mathcal H^{M-1}(\by)$ denotes the PF surface-area element at $\by$.

\begin{proof}
For any measurable $A\subseteq\Delta_M$, injectivity and full rank of $\fPF$ give the area formula
\begin{align}
\int_A
\vartheta(\fPF(\bw))J(\bw) \mathrm d\bw
=
\int_{\fPF(A)}
\vartheta(\by) \mathrm d\mathcal H^{M-1}(\by).
\label{eq:preference_area_formula}
\end{align}
Using~\eqref{eq:preference_pullback_density}, we have
\begin{align}
\int_A
\rho_\vartheta(\bw) \mathrm d\bw
=
\frac{
\int_{\fPF(A)}
\vartheta(\by) \mathrm d\mathcal H^{M-1}(\by)
}{
\int_{\fPF(\Delta_M)}
\vartheta(\by) \mathrm d\mathcal H^{M-1}(\by)
}.
\label{eq:preference_mass_identity}
\end{align}

Applying~\eqref{eq:preference_mass_identity} to each
$\tau\in\mathcal C(W)$ shows that
$\widehat{\mathcal R}_\vartheta(W)=0$
if and only if
$\int_\tau\rho_\vartheta(\bw)\,\mathrm d\bw
=1/|\mathcal C(W)|$ for every $\tau\in\mathcal C(W)$,
equivalently, all PF patches $\fPF(\tau)$ carry equal
preference-weighted area.

The same identity describes the distribution of a PF point generated by sampling $\bw$ from $\rho_\vartheta$. For any measurable
$B\subseteq\fPF(\Delta_M)$,
\begin{align}
\Pr \left(\fPF(\bw)\in B\right)
&=
\int_{\fPF^{-1}(B)}
\rho_\vartheta(\bw) \mathrm d\bw
=
\frac{
\int_B
\vartheta(\by) \mathrm d\mathcal H^{M-1}(\by)
}{
\int_{\fPF(\Delta_M)}
\vartheta(\by) \mathrm d\mathcal H^{M-1}(\by)
}.
\label{eq:preference_pushforward_identity}
\end{align}
Thus sampling weights from $\rho_\vartheta$ produces PF points with density proportional to $\vartheta$ with respect to surface area.
\end{proof}

We also use this formula for reconstructed surfaces. For a $C^1$ map $P$, let $J_P$ denote the area factor obtained by replacing $\fPF$ with $P$ in~\eqref{eq:pf_area_density}. If $P$ is injective on a measurable region $A$, then $\mathcal H^{M-1}(P(A))
=
\int_A J_P(\bw) \mathrm d\bw$. Thus $J_P(\bw) \mathrm d\bw$ measures the surface area represented by an infinitesimal weight-space region.

\subsection{Proof of Proposition~\ref{prop:pf_geometry}}
\label{app:proof_pf_geometry}

\begin{proof}
Fix $0<\delta<1/M$. We first show that the TS solution varies smoothly with its weight. On $\Delta_M^\delta$, the objective $\max_{m=1,\ldots,M}(f_m(\bx)-z_m)/w_m$ is jointly continuous in $(\bx,\bw)$. Compactness of $\X$ and uniqueness in Assumption~\ref{ass:tch_route} therefore imply, by Berge's maximum theorem~\citep{berge1963topological}, that $\bx_{\rm TS}^\star(\bw)$ and $s^\star(\bw)$ are continuous. The utopian reference point and active ray constraints give $s^\star(\bw)>0$. Compactness then gives a uniform positive lower bound.

At each $\bw\in\Delta_M^\delta$, LICQ, strict complementarity, and SOSC make the active Karush--Kuhn--Tucker (KKT) system of~\eqref{eq:ray_epigraph} strongly regular~\citep{bonnans2013perturbation}. The $C^{3,1}$ problem data give a $C^{2,1}$ KKT residual. The implicit-function theorem therefore gives a locally unique $C^{2,1}$ mapping from weights to the solution and its multipliers~\citep{bonnans2013perturbation}. Strict complementarity keeps the active set unchanged locally, and continuity and uniqueness identify this local solution with the global TS solution.

Because $\Delta_M^\delta$ is a compact subset of $\operatorname{relint}(\Delta_M)$, finitely many such neighborhoods cover it. This gives uniform $C^{2,1}$ bounds on a neighborhood of $\Delta_M^\delta$ in the affine hyperplane $\mathbf1^\top\bw=1$. Hence $\bx_{\rm TS}^\star$ and $s^\star$ are uniformly Lipschitz, and $\fPF=\f\circ\bx_{\rm TS}^\star$ is $C^{2,1}$ on this neighborhood.

We next use the radial identity $\fPF(\bw)=\bz+s^\star(\bw)\bw$ to show that changes in the weights produce nonzero changes on the PF. For any tangent direction $\bv$ satisfying $\mathbf1^\top\bv=0$,
\begin{align}
D\fPF(\bw)[\bv]
= s^\star(\bw)\bv+Ds^\star(\bw)[\bv]\bw.
\label{eq:radial_pf_derivative}
\end{align}
If $D\fPF(\bw)[\bv]=0$, summing the coordinates in~\eqref{eq:radial_pf_derivative} gives $Ds^\star(\bw)[\bv]=0$. Since $s^\star(\bw)>0$, it follows that $\bv=0$, so $\fPF$ has full tangent rank. Injectivity follows from Proposition~\ref{prop:tch_coverage}.

The uniform $C^1$ bound gives $|\fPF(\bw)-\fPF(\bv)|\le C_P|\bw-\bv|$. For nearby weights, full tangent rank, continuity of $D\fPF$, and compactness give a uniform lower bound on the ratio of PF distance to weight distance. For weights separated by a fixed positive distance, compactness and injectivity imply that $|\fPF(\bw)-\fPF(\bv)|/|\bw-\bv|$ attains a positive minimum. Together, these bounds give~\eqref{eq:pf_chart_bilipschitz}.

Finally, since $\fPF$ is $C^{2,1}$ on a neighborhood of $\Delta_M^\delta$, its derivative is $C^{1,1}$. Full tangent rank and compactness imply that the singular values of the tangent derivative are uniformly bounded above and bounded away from zero. The Gram-determinant map in~\eqref{eq:pf_area_density} is continuous on full-rank matrices, so $J$ is continuous on $\Delta_M^\delta$. Compactness therefore gives constants $0<\underline J\le\overline J<\infty$ such that $\underline J\le J(\bw)\le\overline J$ for all $\bw\in\Delta_M^\delta$. This proves the claim.
\end{proof}

\subsection{Sufficient Regular Regimes}
\label{app:tch_sufficient_regimes}

\begin{proposition}[{Full-ray TS optimality}]
\label{prop:tch_sufficient_regimes}
Suppose Assumption~\ref{ass:common_model} holds, $\X$ is compact, and $\bz$ is utopian. Fix any $0<\delta<1/M$ and assume that, for every $\bw\in\Delta_M^\delta$, the following conditions hold.
\begin{enumerate}[label=\textup{(\roman*)},leftmargin=2.2em]
\item There exists a unique Pareto-optimal $\bx\in\X$ satisfying $\frac{f_1(\bx)-z_1}{w_1} = \cdots = \frac{f_M(\bx)-z_M}{w_M}$.

\item At this point, the gradients $\nabla\left(\frac{f_m(\bx)-z_m}{w_m}-\frac{f_M(\bx)-z_M}{w_M}\right)$ for {$m=1,\ldots,M-1$}, together with the gradients of the active feasibility constraints, are linearly independent.

\item The constrained first-order stationarity condition for~\eqref{eq:standard_tchebycheff} admits strictly positive coefficients on all $M$ normalized objectives.

\item For the same first-order multipliers, the corresponding Lagrangian has positive second variation along every nonzero feasible first-order direction for which all $M$ normalized objectives have the same directional derivative.
\end{enumerate}
{Then~\eqref{eq:ray_epigraph} has a unique fully active solution satisfying LICQ, positive ray multipliers, and SOSC. Strict complementarity of active feasibility inequalities remains the condition stated in Assumption~\ref{ass:tch_route}. In particular, when no feasibility inequality is active, all conditions of that assumption hold and Proposition~\ref{prop:pf_geometry} applies.}
\end{proposition}

\begin{proof}
Fix $\bw\in\Delta_M^\delta$ and let $\bx$ be the unique point in (i). Let $s$ be the common value of the normalized objective gaps. Then
\begin{align}
f_m(\bx)-z_m=s w_m,
\quad m=1,\ldots,M,
\end{align}
so $(\bx,s)$ is feasible for~\eqref{eq:ray_epigraph} and all ray inequalities are active.

We first show global optimality and uniqueness. If another feasible pair had an epigraph value smaller than $s$, then
\begin{align}
f_m(\bx')-z_m
<
f_m(\bx)-z_m,
\quad m=1,\ldots,M,
\end{align}
which would strictly improve every objective and contradict Pareto optimality of $\bx$. Hence $(\bx,s)$ is globally optimal. If another global solution had the same epigraph value, its objective values would be no larger than those of $\bx$. Pareto optimality forces equality in every objective. The normalized objective gaps are then equal at that solution as well, and uniqueness in (i) gives the same decision point. Thus~\eqref{eq:ray_epigraph} has a unique solution.

We next verify LICQ. Consider any linear dependence among the gradients of the $M$ active ray constraints and the active feasibility constraints in the variables $(\bx,s)$. The $s$ component implies that the coefficients of the ray gradients have weighted sum zero. Eliminating the coefficient of the $M$th ray constraint reduces the $\bx$ component to a linear dependence among the $M-1$ normalized-gap gradients in part~\textup{(ii)} and the active feasibility-constraint gradients. Part~\textup{(ii)} forces all coefficients to vanish. Hence LICQ holds for~\eqref{eq:ray_epigraph}.

Part~\textup{(iii)} gives strictly positive coefficients on the gradients $\nabla f_m(\bx)/w_m$ in the max-form stationarity condition. Dividing the coefficient associated with objective $m$ by $w_m>0$ gives a strictly positive multiplier for the corresponding ray inequality. Since the max-form coefficients sum to one, these ray multipliers satisfy the stationarity condition with respect to $s$. The stationarity condition with respect to $\bx$ is unchanged. Hence every active ray inequality has a positive multiplier.

Finally, full activity and positive multipliers imply that every critical direction of~\eqref{eq:ray_epigraph} satisfies the linearized equalities of all $M$ ray constraints. Eliminating the epigraph component leaves a feasible decision direction along which all normalized objectives have the same directional derivative. If this decision component is zero, the linearized ray equalities also force the epigraph component to be zero. Part~\textup{(iv)} therefore gives strictly positive second variation along every nonzero critical direction, proving SOSC. When the feasibility inequalities are inactive, the positive ray multipliers also establish strict complementarity for the full problem.
\end{proof}

\begin{proposition}[DTLZ2 with $\delta\downarrow 0$]
\label{rem:dtlz2_regular_ts}
Consider standard DTLZ2~\citep{deb2002scalable} with $M\ge2$ objectives.
For every $0<\delta<1/M$, choosing a strictly utopian reference $\bz=-\epsilon\mathbf1$ with $0<\epsilon\le\delta/2$ ensures that the TS problem satisfies the regular full-ray conditions of Proposition~\ref{prop:tch_sufficient_regimes} uniformly on $\Delta_M^\delta$.
In particular, $\delta$ can be made arbitrarily small, provided $\epsilon=O(\delta)$, allowing the represented PF to approach the entire DTLZ2 PF with approximation error $O(\delta)$.
\end{proposition}

\begin{proof}
The DTLZ2 PF is the nonnegative unit sphere, characterized by
$\|\f(\bx)\|_2=1$, with all tail variables equal to $1/2$.
For any $\bw\in\Delta_M^\delta$, the ray from
$\bz=-\epsilon\mathbf1$ in direction $\bw$ intersects this sphere
uniquely.
Indeed, $\|\bw-\epsilon\mathbf1\|_2<1$, and the norm increases
strictly along the ray thereafter.
Since $\epsilon\le\delta/2$, every coordinate of the intersection
is at least $\delta/2>0$.
Thus, the intersection corresponds to a unique decision point with
interior angular coordinates and tail variables equal to $1/2$.
No smaller TS objective value is feasible, since it would require
an objective vector strictly dominated by this PF point.
This establishes part~\textup{(i)}.

The DTLZ2 objective map has full tangent rank on the strictly
positive PF, and $\fPF(\bw)^\top\bw>0$ ensures that the ray is
transverse to the sphere, verifying part~\textup{(ii)}.
The stationarity multipliers are strictly positive and proportional
to $w_m f_m(\bx_{\rm TS}^\star(\bw))$, establishing
part~\textup{(iii)}.
For part~\textup{(iv)}, the linearized active constraints
eliminate all nonzero angular critical directions, while the
remaining tail directions have strictly positive quadratic curvature.
These conditions hold uniformly on each fixed $\Delta_M^\delta$
by smoothness and compactness.

Finally, the ray intersection satisfies $\left\|\fPF(\bw)
-\frac{\bw}{\|\bw\|_2}\right\|_2
\le 2\sqrt{M}\epsilon$.
As $\delta\downarrow0$, $\Delta_M^\delta$ approaches the full
simplex, and the normalization map
$\bw\mapsto\bw/\|\bw\|_2$ is Lipschitz on the simplex.
Consequently, every point on the DTLZ2 PF can be approximated
by the represented PF with error $O(\delta+\epsilon)=O(\delta)$.
Hence, $\delta$ can be made arbitrarily small while preserving
the full-ray conditions on each $\Delta_M^\delta$.
\end{proof}

\subsection{Weighted-Sum Specialization}
\label{sec:linear_specialization}
\label{app:proof_linear_geometry}

Weighted-sum scalarization gives another regular PF chart when the objectives become strongly convex after a smooth change of variables. We consider
\begin{align}
\bx_{\rm LS}^\star(\bw)\in\argmin_{\bx\in\X}\sum_{m=1}^M w_m f_m(\bx), \quad \bw\in\Delta_M.
\label{eq:linear_scalarization}
\end{align}

Although optimization problems may be nonconvex in their original parameterization, some admit an equivalent convex representation through a change of variables, as studied in hidden-convex optimization~\citep{fatkhullin2025stochastic}.
Building on this perspective, we impose hidden strong convexity and additional regularity conditions to establish a smooth, nondegenerate PF chart for weighted-sum scalarization.

\begin{assumption}[Hidden strong convexity]
\label{ass:linear_route}
There exist a compact convex set $\mathcal Y$ and a $C^{3,1}$ diffeomorphism $\Psi$ between neighborhoods of $\mathcal Y$ and $\X$, with $\Psi(\mathcal Y)=\X$, such that each transformed objective $f_m\circ\Psi$ is $\mu$-strongly convex on $\mathcal Y$ for a common $\mu>0$.
The set $\mathcal Y$ is defined by $C^{3,1}$ convex inequalities and affine equalities.
For every $\bw\in\Delta_M$, the transformed weighted-sum problem satisfies LICQ and strict complementarity at its minimizer, and the objective gradients projected onto the tangent space of the active constraints have rank $M-1$.
\end{assumption}

\begin{proposition}[Regular PF chart from weighted sums]
\label{prop:linear_specialization}
Suppose Assumptions~\ref{ass:common_model} and~\ref{ass:linear_route} hold. Then~\eqref{eq:linear_scalarization} has a unique solution for every $\bw\in\Delta_M$, weighted-sum scalarization covers the PF, and $\fPF(\bw):=\f(\bx_{\rm LS}^\star(\bw))$ is injective with full tangent rank. Moreover, $\fPF$ admits a $C^{2,1}$ extension to a neighborhood of $\Delta_M$ in the affine hyperplane $\mathbf1^\top\bw=1$. There exist constants $0<c_P\le C_P<\infty$,
\begin{align}
\text{s.t.}\quad c_P\|\bw-\bv\|\le\|\fPF(\bw)-\fPF(\bv)\|\le C_P\|\bw-\bv\|,\quad \bw,\bv\in\Delta_M.
\label{eq:linear_pf_bilipschitz}
\end{align}
The area factor $J$ is $C^{1,1}$ on $\Delta_M$ and satisfies $0<\underline J\le J(\bw)\le\overline J<\infty$.
\end{proposition}
For regularized policy objectives, the proposition applies when a smooth invertible policy parameterization satisfies Assumption~\ref{ass:linear_route}, including the condition that weight changes produce nonzero first-order changes on the PF.

\begin{proof}
Under $\bx=\Psi(\by)$, problem~\eqref{eq:linear_scalarization} minimizes $\sum_{m=1}^M w_m(f_m\circ\Psi)(\by)$ over $\mathcal Y$.
Since the transformed objectives are $\mu$-strongly convex and $\bw\in\Delta_M$, each weighted-sum problem has a unique minimizer.
Convexity also ensures that weighted-sum scalarization covers the entire PF~\citep{ehrgott2005multicriteria}.
Conversely, every weighted-sum minimizer is Pareto optimal: any dominating point would either improve the weighted objective or contradict uniqueness.
The bijection $\Psi$ preserves these properties in the original variables.

We next establish smoothness.
Strong convexity and convexity of the constraints imply SOSC.
Together with LICQ and strict complementarity, these conditions ensure nonsingularity of the KKT system~\citep{bonnans2013perturbation}.
Since the problem data are $C^{3,1}$, the implicit-function theorem yields a locally unique $C^{2,1}$ solution mapping.
Strict complementarity preserves the active set locally, while uniqueness makes these local mappings consistent.
They therefore extend to a neighborhood of the compact simplex $\Delta_M$, giving a $C^{2,1}$ extension of $\bx_{\rm LS}^\star$ and $\fPF$.

To prove full tangent rank, fix $\bw\in\Delta_M$ and a nonzero $\bv$ with $\mathbf1^\top\bv=0$.
Let $\dot\by$ denote the derivative of the transformed minimizer in direction $\bv$.
Suppose $D\fPF(\bw)[\bv]=0$, so $\nabla(f_m\circ\Psi)^\top\dot\by=0$ for every $m$.
Differentiating the KKT conditions and taking the inner product with $\dot\by$ gives
\begin{align}
\dot\by^\top H\dot\by
=-\dot\by^\top\sum_{m=1}^M v_m\nabla(f_m\circ\Psi)=0,
\end{align}
where $H$ is the Lagrangian Hessian.
Since $\dot\by$ lies in the active-constraint tangent space, SOSC implies $\dot\by=0$.
The differentiated stationarity condition then shows that $\bv$ lies in the nullspace of the projected objective-gradient matrix.
By KKT stationarity and the rank-$M-1$ assumption, this nullspace is spanned by $\bw$.
Since $\mathbf1^\top\bv=0$, we obtain $\bv=0$, a contradiction.
Hence $\fPF$ has full tangent rank.

For injectivity, suppose $\fPF(\bw)=\fPF(\bv)$.
The corresponding minimizers attain the same objective vector and therefore minimize both weighted-sum problems.
Uniqueness gives the same decision point.
Subtracting their KKT stationarity conditions shows that $\bw-\bv$ lies in the nullspace of the projected objective-gradient matrix.
The rank condition and $\mathbf1^\top(\bw-\bv)=0$ imply $\bw=\bv$.

Finally, the $C^{2,1}$ regularity, full tangent rank, injectivity, and compactness of $\Delta_M$ yield the bi-Lipschitz bounds in~\eqref{eq:linear_pf_bilipschitz}.
The Gram-determinant expression~\eqref{eq:pf_area_density} then implies that $J$ is $C^{1,1}$ and satisfies $0<\underline J\le J(\bw)\le\overline J<\infty$ uniformly on $\Delta_M$.
\end{proof}

\section{Details for the SURFACE Algorithm}
\label{app:proofs_theory}

We analyze PF reconstruction and weight redistribution on the fixed domain $\Delta_M^\delta$ from Proposition~\ref{prop:pf_geometry}.
Throughout, $C$ denotes constants independent of $N$, $T$, $t$, and $\alpha$; unless specified otherwise, they may depend on fixed regularity parameters, including $\kappa_\vartheta$.
The round-dependent coefficient $C_t$ captures the largest PF-patch diameter on the current reconstruction mesh.
For PF reconstruction, let $\mathcal T(W)$ denote the straight simplicial mesh through the returned vertices.
To obtain exactly $N$ adaptive vertices, we map the finest standard edgewise grid with at most $N$ noncorner vertices onto $\Delta_M^\delta$, keeping its $M$ corners fixed.
Any remaining vertices are inserted at the barycenters of distinct cells, each subdivided into $M$ simplices.
Since consecutive edgewise grids differ by $O(N^{(M-2)/(M-1)})$ vertices and contain $\Theta(N)$ cells, these insertions suffice for all sufficiently large $N$.
In the Moser realization, complete transported reference cells define $\mathcal C(W)$. In the capacity realization, complete power cells define $\mathcal C(W)$.
The retained reference connectivity defines $\mathcal T(W)$ in either case.
The straight reconstruction mesh $\mathcal T(W)$ has $\Theta(N)$ uniformly shape-regular simplices of diameter comparable to $N^{-1/(M-1)}$, with corresponding mesh bounds on each face. Quasi-uniformity means that cell sizes remain comparable across the domain. Shape regularity means that the simplices avoid becoming arbitrarily thin relative to their diameter. These geometric conditions apply to the reconstruction mesh. The power cells are used to measure mass and need not satisfy them.
The two realizations retain the same reconstruction connectivity and vertex labels across rounds.
For $M=2$, no additional insertions are needed.

Reconstruction also uses fixed TS chart-corner samples at $\bw_m^\delta$.
Their initial solver states satisfy
\begin{align}
\max_{m\in[M]}
\left\|\tilde\bx_{m,\delta}^{(0)}
-\bx_{\rm TS}^\star(\bw_m^\delta)\right\|
\le c_x N^{-3/(M-1)},
\label{eq:corner_initial_accuracy}
\end{align}
for some $\Oc(1)$ constant $c_x>0$.
These states are retained or refined without increasing their error, providing corner interpolation data of the same accuracy in objective space at every round.
If not supplied, the corner samples require $M$ additional initial solves. Subsequent refinements are counted in the inner-solver cost.

\subsection{Reconstructing the PF Between Sampled Weights}
\label{app:reconstruction_proof}

We use the quadratically augmented local RBF interpolation from Section~\ref{sec:algorithm}, with strictly positive-definite $C^4$ Wendland kernels and a smooth partition of unity~\citep{wendland2005scattered}. Each local model combines an RBF fit with quadratic terms, and the partition of unity blends the overlapping models smoothly.
Each local fit uses all mesh vertices within its partition-weight support.

The overlapping patches have diameter $\Theta(N^{-1/(M-1)})$, bounded overlap, and kernel support radii proportional to their diameters.
They are enlarged sufficiently to ensure uniform quadratic unisolvency, including at the boundary of $\Delta_M^\delta$. This means that the sample locations distinguish quadratic polynomials uniquely and with uniform stability after each patch is rescaled to unit size.
The partition weights have derivatives of order $j\le3$ bounded by $CN^{j/(M-1)}$.
For $\kappa_\vartheta$-independent reconstruction constants, we additionally require the following bounds to be uniform across the considered preference profiles: after rescaling $\vartheta$ so that $\int_{\Delta_M^\delta}\vartheta(\fPF)J=1$, its $C^{2,1}$ norm; the chart's $C^{2,1}$ norm and positive tangent-rank margin; and the stability of the rescaled local interpolation systems.
These conditions hold for a fixed regular reference mesh and a fixed chart with uniformly bounded normalized preference smoothness, but require separate verification for a density-adapted mesh. They do not require a uniform positive lower bound on $\vartheta$.

Under the $C^{2,1}$ chart regularity of Proposition~\ref{prop:pf_geometry}, we define the round-dependent reconstruction factor
\begin{align}
C_t
:=
1+
N^{2/(M-1)}
\max_{\tau\in\mathcal T(W^{(t)})}
\operatorname{diam}\left(\fPF(\tau)\right)^2.
\label{eq:round_reconstruction_factor}
\end{align}
This factor captures the largest PF-patch diameter relative to the nominal mesh spacing $N^{-1/(M-1)}$.
The mesh and chart regularity bounds ensure that $C_t$ remains bounded across rounds for each fixed preference, with a bound that may depend on $\kappa_\vartheta$.
For capacity MEQ, the density-selected sites and their admissible refinements preserve a regular reconstruction mesh, so $C_t=\Oc(1)$ across rounds for each fixed regular problem.
For adaptive Moser MEQ, a $\kappa_\vartheta$-independent bound after some round follows if $\max_{\tau\in\mathcal T(W^{(t)})}\operatorname{diam}(\fPF(\tau))\le D N^{-1/(M-1)}$ holds from that round onward with $D$ independent of $\kappa_\vartheta$; then $C_t\le1+D^2$.
This extra PF-diameter bound is not implied by equal cell mass alone.
For example, when $M=2$ and $\vartheta\equiv1$, exact inverse-CDF MEQ gives equal PF arc length to the transported intervals, so the bound follows from a uniform bound on total PF arc length.
For strongly nonuniform preferences, low-preference regions may instead contain large PF intervals.

\begin{lemma}[PF reconstruction accuracy]
\label{lem:chart_reconstruction}
Suppose Assumptions~\ref{ass:tch_route} and~\ref{ass:common_model} hold, and the pinned-corner data satisfy the initialization and retention rule in~\eqref{eq:corner_initial_accuracy}. Suppose the current weight mesh is quasi-uniform and shape regular on $\Delta_M^\delta$ and on each relative face, with mesh spacing comparable to $N^{-1/(M-1)}$. If the adaptive scalarized solutions used to construct $P_t$ are exact, then, for all sufficiently large $N$, the restriction of $P_t$ to $\Delta_M^\delta$ satisfies
\begin{align}
\|P_t-\fPF\|_{C^1}
&\le C C_t N^{-2/(M-1)}, ~
\|P_t-\fPF\|_{C^2}\le C C_t^{1/2} N^{-1/(M-1)}, ~
\|P_t\|_{C^{2,1}} \le C.
\label{eq:chart_reconstruction_local}
\end{align}
If the scalarized solutions are inexact, then
\begin{align}
\|P_t-\fPF\|_{C^1}
\le{}&
C C_t N^{-2/(M-1)}
+
C N^{1/(M-1)}
\max_{n\in[N]}
\left\|
\tilde\bx_n^{(t)}-
\bx_{\rm TS}^\star(\bw_n^{(t)})
\right\|,
\label{eq:inexact_chart_c1}\\
\|P_t-\fPF\|_{C^2}
\le{}&
C C_t^{1/2} N^{-1/(M-1)}
+
C N^{2/(M-1)}
\max_{n\in[N]}
\left\|
\tilde\bx_n^{(t)}-
\bx_{\rm TS}^\star(\bw_n^{(t)})
\right\|.
\label{eq:inexact_chart_c2}
\end{align}
Moreover, if the maximum decision error is $\Oc(N^{-3/(M-1)})$, then $P_t$ has a uniform $C^{2,1}$ bound on $\Delta_M^\delta$. The fixed chart-corner samples are included in the reconstruction mesh and satisfy~\eqref{eq:corner_initial_accuracy}.
\end{lemma}

\begin{proof}
We first establish uniform interpolation stability.
After rescaling, quasi-uniformity gives bounded vertex counts and positive
separation, while patch construction ensures quadratic unisolvency,
including at the boundary.
Strict positive definiteness of the Wendland kernel then yields uniformly
bounded inverses, with stability preserved by the partition of unity.
Under the stated additional uniformity conditions, these constants are
independent of $\kappa_\vartheta$. 

Let $h=N^{-1/(M-1)}$.
By the bi-Lipschitz property of Proposition~\ref{prop:pf_geometry} and~\eqref{eq:round_reconstruction_factor}, the local interpolation scales are bounded by $C C_t^{1/2}h$.
Since $\fPF$ is $C^{2,1}$, its quadratic Taylor remainder is cubic in the local patch diameter.
Quadratic reproduction and interpolation stability therefore give errors of order $C C_t h^2$ in $C^1$ and $C C_t^{1/2}h$ in $C^2$.
The cubic remainder also controls the Lipschitz second derivative, yielding a uniform $C^{2,1}$ bound.
The partition of unity preserves these estimates, including on boundary patches by the one-sided chart regularity.

For inexact data, linearity of the interpolation operator implies that nodal perturbations are amplified by at most $Ch^{-j}$ in derivatives of order $j\le3$.
Since $\f$ is uniformly Lipschitz near the TS solution path, objective-value errors are bounded by the corresponding decision errors.
Adding these perturbations to the exact-data estimates proves~\eqref{eq:inexact_chart_c1}--\eqref{eq:inexact_chart_c2}.
If the maximum adaptive decision error is $O(h^3)$, the third-order amplification remains bounded, preserving the $C^{2,1}$ estimate.

Finally, the corner errors in~\eqref{eq:corner_initial_accuracy} contribute $O(h^2)$ in $C^1$, $O(h)$ in $C^2$, and $O(1)$ in $C^{2,1}$, all absorbed into the stated bounds.
\end{proof}
The next result transfers chart error to the density used in~\eqref{eq:density_update}.

\begin{lemma}[Preference-density stability]
\label{lem:area_density_stability}
Suppose Assumptions~\ref{ass:tch_route}--\ref{ass:preference_profile} hold. If $P_t$ is sufficiently close to $\fPF$ in $C^1$, then $P_t$ has full tangent rank and
\begin{align}
\left\|
\frac{\vartheta(P_t)J_{P_t}}
{\int_{\Delta_M^\delta}\vartheta(P_t)J_{P_t}}
-
\rho_\vartheta
\right\|_\infty
&\le
C\|P_t-\fPF\|_{C^1},\quad
\left\|
\frac{\vartheta(P_t)J_{P_t}}
{\int_{\Delta_M^\delta}\vartheta(P_t)J_{P_t}}
-
\rho_\vartheta
\right\|_{C^1}\le
C\|P_t-\fPF\|_{C^2}.
\label{eq:density_stability_c1}
\end{align}
If $P_t$ has a uniform $C^{2,1}$ bound, then the normalized density on the left is strictly positive and has a uniform $C^{1,1}$ bound.
\end{lemma}

\begin{proof}
Proposition~\ref{prop:pf_geometry} gives a uniform positive lower bound on the tangent singular values of $D\fPF$. A sufficiently small $C^1$ perturbation preserves this bound for $DP_t$. The Gram-determinant expression in~\eqref{eq:pf_area_density} is smooth on the resulting compact full-rank set. Hence
\begin{align}
\|J_{P_t}-J\|_\infty
&\le C\|P_t-\fPF\|_{C^1},\quad \text{and}\quad
\|J_{P_t}-J\|_{C^1}
\le C\|P_t-\fPF\|_{C^2}.
\end{align}
Assumption~\ref{ass:preference_profile} and the chain rule give the corresponding estimates for $\vartheta(P_t)-\vartheta(\fPF)$. The product rule gives the same orders for $\vartheta(P_t)J_{P_t}-\vartheta(\fPF)J$.
For the $\kappa_\vartheta$-uniform variant, normalize $\vartheta$ so that the exact integral equals one. The uniform upper smoothness bounds then make the integrand perturbation at most $C\|P_t-\fPF\|_{C^1}$ with $C$ independent of $\kappa_\vartheta$. For sufficiently small error, the reconstructed normalizer is at least $1/2$, so the $L^\infty$ density-stability constant is also independent of $\kappa_\vartheta$.
For the $L^\infty$ estimate, the area factor is the norm of an exterior product of chart derivatives and is Lipschitz on bounded sets without a positive lower bound on $J$. For the $C^1$ estimate, a uniform tangent-rank margin additionally controls derivatives of this norm. The final regularity claim follows from the uniform $C^{2,1}$ bounds.
Without the additional uniformity conditions, the original estimates hold with constants depending on the fixed preference and chart bounds.
\end{proof}

Combining Lemmas~\ref{lem:chart_reconstruction} and~\ref{lem:area_density_stability} gives
\begin{align}
\left\|
\frac{\vartheta(P_t)J_{P_t}}
{\int_{\Delta_M^\delta}\vartheta(P_t)J_{P_t}}
-
\rho_\vartheta
\right\|_\infty
\le
C C_t N^{-2/(M-1)}
+
C N^{1/(M-1)}
\max_{n\in[N]}
\left\|
\tilde\bx_n^{(t)}-
\bx_{\rm TS}^\star(\bw_n^{(t)})
\right\|.
\label{eq:main_reconstruction_inexact}
\end{align}
This is the reconstruction bound stated in Lemma~\ref{lem:main_reconstruction}. Under the additional uniformity conditions above, the constants $C$ in this reconstruction bound are independent of $\kappa_\vartheta$, while $C_t$ retains its dependence on the current PF-patch geometry.
Decision errors of order $N^{-3/(M-1)}$ contribute at the same $N^{-2/(M-1)}$ scale after differentiation.

\subsection{Finite MEQ: Mass Accuracy and Mesh Stability}
\label{app:finite_meq}

Finite MEQ redistributes the weights for the normalized density $\rho$ in~\eqref{eq:density_update}.
The complete represented cells $\mathcal C(W)$ measure target mass, whereas the straight simplicial mesh $\mathcal T(W)$ through the returned weights and fixed corners is used for PF reconstruction.
We give two finite realizations with the same cell-mass accuracy and reconstruction-mesh guarantees.

\subsubsection{Face-Preserving Moser Transport}
\label{app:redistribution_proof}
\label{app:numerical_meq}
\label{app:mesh_quality}

On the regular reference mesh, locally average $\rho$ over overlapping patches of diameter $\Theta(N^{-1/(2(M-1))})$.
Following~\citet{bruveris2018moser}, construct two face-tangent divergence fields: the first transports the uniform density $\rho_0$ to the local average, and the second transports that average to $\rho$.
For each stage, linearly blend the source and destination densities over time. Dividing the corresponding divergence field by this positive density gives a velocity field. Integrating that velocity moves the weights from the source distribution to the destination distribution.
The composition $\Phi$ preserves relative faces and corners and satisfies
\begin{align}
\rho(\Phi(\bw))\det D\Phi(\bw)=\rho_0(\bw).
\label{eq:moser_pullback}
\end{align}
For $M=2$, this is implemented by inverse-CDF sampling~\citep{jiang2026surf}.

\begin{proposition}[Finite face-preserving Moser realization]
\label{prop:finite_moser_realization}
Let $\rho$ be a positive normalized $C^{1,1}$ density with common positive bounds.
Suppose it admits a regular component $g$ with a common $C^{1,1}$ bound such that
\begin{align}
&\|\rho-g\|_\infty+N^{-1/(M-1)}\|\rho-g\|_{C^1}
+N^{-2/(M-1)}\|\rho-g\|_{C^{1,1}}\nonumber\\
&\hspace{31mm}\le cN^{-1/(2(M-1))},
\label{eq:moser_scale_condition}
\end{align}
for sufficiently small $c$.
Then the selected $\Phi$ is uniformly bi-Lipschitz, face preserving, and satisfies
\begin{align}
N^{-1/(M-1)}\operatorname{Lip}(D\Phi)
\le C\bigl(N^{-1/(2(M-1))}+c\bigr).
\label{eq:moser_scaled_regular}
\end{align}
Using the same reference and divergence operators for every input, the selected transport is Lipschitz in the density in the uniform norm, with a constant independent of $N$.
A finite realization attains
\begin{align}
\max_{\tau\in\mathcal C(W)}\left|
|\mathcal C(W)|\int_\tau\rho(\bw)\,\mathrm d\bw-1\right|
\le CN^{-2/(M-1)}.
\label{eq:finite_meq_error}
\end{align}
\end{proposition}

\begin{proof}
Cover the simplex by overlapping interior, face, and corner patches of diameter $\Theta(N^{-1/(2(M-1))})$, with bounded overlap and smooth partition weights. Subtracting patch averages before differentiation shows that local averaging preserves positive bounds and normalization, has bounded $C^1$ norm, and has $C^{1,1}$ norm $\Oc(N^{1/(2(M-1))})$ under~\eqref{eq:moser_scale_condition}.

Apply a fixed face-tangent divergence operator~\citep{bruveris2018moser} to the difference between $\rho_0$ and the local average. Its coordinate integrals and cutoffs give bounded $C^1$ velocity after division by the positive interpolated density. The velocity's $C^{1,1}$ norm is $\Oc(N^{1/(2(M-1))})$. The second flow corrects the averaging difference. This difference decomposes into patch-supported zero-integral terms. Rescaling the same divergence formulas and extending by zero gives a globally face-tangent correction field. The rescaled estimates and~\eqref{eq:moser_scale_condition} bound its first derivative and mesh-scaled derivative variation by $C(N^{-1/(2(M-1))}+c)$. The denominators in both flows remain positive. The variational equations and Gronwall's inequality yield uniform bi-Lipschitz bounds and~\eqref{eq:moser_scaled_regular}. Face tangency preserves relative faces and corners, and mass conservation proves~\eqref{eq:moser_pullback}.
For two input densities, local averaging is nonexpansive in the uniform norm, and the local divergence correction has uniform operator norm $\Oc(N^{-1/(2(M-1))})$ in that norm.
The two Moser velocities consequently differ by at most $C$ times the uniform density difference.
Their common spatial Lipschitz bounds and Gronwall's inequality give the claimed stability of the selected transports.

Approximate the fixed local divergence fields and integrate both flows with face-preserving quadrature and time steps. Refine until the map-value error is $\Oc(\alpha N^{-2/(M-1)})$, the derivative error is $\Oc(N^{-2/(M-1)})$, and the defect in~\eqref{eq:moser_pullback} is $\Oc(N^{-2/(M-1)})$. These tolerances are attainable by convergence of the numerical fields and flows for each fixed $N$. Integrating over each equal-volume reference cell, of volume $\Theta(N^{-1})$, gives~\eqref{eq:finite_meq_error}. Independent cell quadrature to absolute accuracy $\Oc(N^{-1-2/(M-1)})$ certifies the estimate.
\end{proof}

\begin{lemma}[Moser mesh stability]
\label{lem:mesh_invariance}
Under Proposition~\ref{prop:finite_moser_realization}, the straight mesh $\mathcal T(W)$ is quasi-uniform and shape regular at scale $N^{-1/(M-1)}$ on $\Delta_M^\delta$ and every relative face, uniformly over the rounds.
\end{lemma}
\begin{proof}
The piecewise-affine interpolation of $\Phi$ has derivative error at most $CN^{-1/(M-1)}\operatorname{Lip}(D\Phi)$, while finite nodal errors add $\Oc(\alpha N^{-1/(M-1)})$.
The bi-Lipschitz bound for the exact transport gives a positive lower bound on the ratio of transported distance to original distance. By~\eqref{eq:moser_scaled_regular}, choose $c$ sufficiently small and then $N$ sufficiently large so that the total interpolation error is smaller than half this lower bound. The required resolution depends on the density and transport bounds.
The interpolant is thus globally bi-Lipschitz. Face preservation and degree one ensure full coverage without overlapping straight simplices.
The resulting singular-value bounds prove shape regularity and quasi-uniformity, including on relative faces. Their constants may depend on the transport's density bounds, and hence on $\kappa_\vartheta$.
\end{proof}

\subsubsection{Equal-Capacity Minimization}
\label{app:capacity_meq}

Alternatively, equal-capacity MEQ directly partitions $\Delta_M^\delta$
into cells of equal $\rho$-mass, using the density in~\eqref{eq:density_update}.
Here the sites are the scalarization weights. A positive smooth selection density determines where they are placed. Select their coordinates successively by inverting the conditional CDF of each coordinate given the preceding coordinates. Retain the reference mesh connectivity and fixed corners.
For exact solves, the input density itself has the required regularity.
For inexact solves, apply a fixed, positive, smooth averaging rule to obtain the site-selection density. Evaluate the cell capacities against the original $\rho$.
Both choices give uniformly bi-Lipschitz, face-preserving sites, and the fixed averaging rule is Lipschitz in the input density in the uniform norm.
{Identify $\Delta_M^\delta$ affinely with the standard simplex and use barycentric stick-breaking coordinates $z_i=w_i/(1-\sum_{j<i}w_j)$. The uniform measure has beta-type conditional densities, which a positive $C^{1,1}$ target modifies by bounded positive factors. Successive conditional-CDF matching fixes $z_i=0,1$, preserving each face and corner. The conditional-quantile identities and common density bounds control the derivatives of the map and its inverse, including face limits, and its stability under uniform density perturbations. Standard piecewise-affine interpolation then preserves the regularity of the selected mesh for sufficiently large $N$.}
The density-selected sites may be refined by minimizing the equal-capacity transport cost over $W$ alone:
\begin{align}
\min_{W}\quad &
\sum_{n=1}^N\int_{\tau_n(W)}
\|\bw-\bw_n\|^2\rho(\bw)\,\mathrm d\bw,
\label{eq:capacity_site_min}
\end{align}
subject to each site remaining on its original relative face and within an $\Oc(\alpha N^{-2/(M-1)})$ perturbation of its selected position. For every admissible collection of distinct sites $W$, the cells $\tau_n(W)$ are the capacity-balanced power cells
\begin{align}
\tau_n(W)=\{\bw\in\Delta_M^\delta:\;
\|\bw-\bw_n\|^2-\lambda_n
\le \|\bw-\bw_j\|^2-\lambda_j,\ \forall j\},
\label{eq:capacity_candidate_cells}
\end{align}
whose offsets maximize
\begin{align}
\frac1N\sum_{n=1}^N\lambda_n
+\int_{\Delta_M^\delta}
\min_n\{\|\bw-\bw_n\|^2-\lambda_n\}
\rho(\bw)\,\mathrm d\bw.
\label{eq:capacity_candidate_dual}
\end{align}
The maximizing offsets in~\eqref{eq:capacity_candidate_dual} enforce $\int_{\tau_n(W)}\rho(\bw)\,\mathrm d\bw=1/N$. Thus, the cells are determined by $W$ and $\rho$, and the objective in~\eqref{eq:capacity_site_min} optimizes the returned scalarization sites. Here $\mathcal C(W)$ comprises these complete power cells, while $\mathcal T(W)$ retains the reference connectivity through the returned sites and pinned corners.

\begin{proposition}[Finite equal-capacity realization]
\label{prop:capacity_meq_proof}
For the positive normalized densities generated under Assumptions~\ref{ass:tch_route}--\ref{ass:preference_profile}, select the density-dependent sites above and optimize the offsets in~\eqref{eq:capacity_candidate_dual}. This gives a finite capacity-balanced realization satisfying~\eqref{eq:finite_meq_error}. The selected sites, as well as any minimizer of~\eqref{eq:capacity_site_min} within the prescribed displacement bounds, define a quasi-uniform, shape-regular reconstruction mesh $\mathcal T(W)$ for sufficiently large $N$, including on relative faces.
\end{proposition}
\begin{proof}
Fix the density-selected noncorner sites and the pinned corners. The smooth, uniformly bi-Lipschitz selection map and sufficiently fine reference triangulation give a quasi-uniform, shape-regular straight mesh, including on relative faces, with distinct sites. For these fixed sites, the concave dual~\eqref{eq:capacity_candidate_dual} attains its maximum modulo a common offset: positivity of $\rho$ and the prescribed positive masses prevent an optimizing sequence of normalized offsets from diverging.
Its derivative with respect to $\lambda_n$ is $1/N-\int_{\tau_n(W)}\rho$, because power-cell interfaces have zero volume. Hence all complete power cells have mass $1/N$ at a maximizer. Finite dual optimization and quadrature to absolute mass accuracy $\Oc(N^{-1-2/(M-1)})$ per cell yield~\eqref{eq:finite_meq_error}.
The selected sites themselves satisfy the displacement constraints in~\eqref{eq:capacity_site_min}. The admissible site set is compact and preserves separation for sufficiently large $N$, so the continuous reduced objective attains a minimum. The allowed $\Oc(\alpha N^{-2/(M-1)})$ nodal perturbations change the affine element derivatives by $\Oc(\alpha N^{-1/(M-1)})$, preserving the selected mesh regularity. Site positions vary with the selection density, while dual offsets enforce capacities for each selected or refined $W$.
\end{proof}

For each selected or refined $W$, optimizing the power-cell offsets balances cell masses without moving the sites; redistribution of $W$ comes from density-dependent selection and optional constrained site refinement.

\begin{lemma}[Round-to-round admissibility]
\label{lem:round_admissibility}
Under Assumptions~\ref{ass:tch_route}--\ref{ass:preference_profile}, both MEQ realizations preserve a regular reconstruction mesh and positive, uniformly $C^{1,1}$ densities at every exact-solve round.
\end{lemma}
\begin{proof}
The uniform initial density and regular reference mesh start a joint induction.
Fix the positive density and mesh regularity bounds, with slack around the convex segment joining $\rho_0$ and $\rho_\vartheta$, before choosing the reconstruction accuracy and $N$.
For a regular current mesh, Lemmas~\ref{lem:chart_reconstruction} and~\ref{lem:area_density_stability} give full tangent rank, a uniformly $C^{2,1}$ reconstructed chart, and a positive normalized reconstructed density with uniform $C^{1,1}$ bounds for exact solves.
Damping in~\eqref{eq:density_update} preserves these bounds.
Proposition~\ref{prop:finite_moser_realization} with $g=\rho_{t+1}$ and Lemma~\ref{lem:mesh_invariance} give the next Moser mesh.
For the capacity realization, the density-selected sites and their admissible refinements provide the next regular reconstruction mesh.
The fresh-density $C^1$ error is $\Oc(N^{-1/(M-1)})$ by Lemmas~\ref{lem:chart_reconstruction} and~\ref{lem:area_density_stability}. Unrolling the damped update keeps $\rho_{t+1}$ within the fixed density class, while the preceding mesh bounds keep the next mesh in its fixed regularity class. This closes the exact-solve induction.
For the capacity realization, the selected-site mesh remains regular under the stated displacement restriction, so $N$ must exceed the interpolation, site-selection mesh, and tangent-rank thresholds. These thresholds depend on the chart and preference regularity.
Theorem~\ref{thm:surface_inexact} requires $N$ to exceed these fixed-problem thresholds, including the Moser mesh-stability threshold in Lemma~\ref{lem:mesh_invariance}.
For inexact solves, the joint density--mesh--solver induction is proved in Lemma~\ref{lem:inexact_admissibility}.
The capacity realization uses the regularity of the site-selection map and the positive density bounds for its power-cell dual, whereas the Moser step uses its scale-dependent condition.
\end{proof}

\begin{proof}[Proof of Proposition~\ref{prop:main_meq}]
Lemma~\ref{lem:round_admissibility} supplies the admissible densities and meshes at each round.
Proposition~\ref{prop:finite_moser_realization} and Lemma~\ref{lem:mesh_invariance} establish the finite Moser realization.
Proposition~\ref{prop:capacity_meq_proof} establishes the capacity realization with the same mass accuracy and mesh guarantees.
\end{proof}

\begin{proposition}[Target-mass error after MEQ]
\label{prop:target_mass_residual}
For either realization, let $\rho$ be the normalized density passed to MEQ and $W$ its returned weights and fixed corners.
Then
\begin{align}
\widehat{\mathcal R}_\vartheta(W)
\le C\|\rho-\rho_\vartheta\|_\infty
+C N^{-2/(M-1)}.
\label{eq:curved_target_mass_bound}
\end{align}
\end{proposition}
\begin{proof}
By~\eqref{eq:finite_meq_error}, each represented cell $\tau$ has $\rho$-mass at most $[1+CN^{-2/(M-1)}]/|\mathcal C(W)|$.
The positive lower bound on $\rho$ therefore gives $|\mathcal C(W)|\operatorname{Vol}_{M-1}(\tau)\le C$.
Consequently,
\begin{align}
|\mathcal C(W)|\left|
\int_\tau(\rho_\vartheta(\bw)-\rho(\bw))\,\mathrm d\bw
\right|\le C\|\rho-\rho_\vartheta\|_\infty.
\end{align}
Adding the finite cell-mass error proves~\eqref{eq:curved_target_mass_bound}. The mass-residual constant may depend on the positive lower bound on $\rho$.
\end{proof}

For moving-weight Moser MEQ, the complete transported cells and the corresponding straight reconstruction simplices have comparable diameters by Lemma~\ref{lem:mesh_invariance}. The definition of $\widehat{\mathcal R}_\vartheta$ bounds the preference-weighted PF area of each cell by $(1+\widehat{\mathcal R}_\vartheta)/|\mathcal C(W^{(t)})|$ times the total preference mass. Positivity of $\vartheta$ and the transported-cell shape regularity convert this area bound into a PF-diameter bound. Since $|\mathcal C(W^{(t)})|=\Theta(N)$,~\eqref{eq:round_reconstruction_factor} yields
\begin{align}
C_t=\Oc\left(1+\left(1+\widehat{\mathcal R}_\vartheta(W^{(t)})\right)^{2/(M-1)}\right),
\label{eq:round_resolution_from_mass}
\end{align}
For capacity MEQ, the density-selected mesh likewise gives a round-uniform bound on $C_t$.

\section{Proofs for the Convergence Analysis}
\label{app:proofs_convergence}

The reconstruction and MEQ bounds now give the convergence results in Section~\ref{sec:theory}. With exact scalarized solves, the density update contracts toward the target up to reconstruction error. With inexact solves, we track solver error together with density error and show that the required density and mesh bounds hold at every round. Throughout, $N$ is taken large enough to meet the reconstruction and mesh-stability thresholds above, which may depend on the fixed regularity parameters, e.g., $\kappa_\vartheta$.

\subsection{Exact Scalarized Solves}
\label{app:proofs_exact}

\begin{theorem}[Exact-solve convergence of SURFACE]
\label{thm:surface}
Suppose Assumptions~\ref{ass:tch_route}--\ref{ass:preference_profile} hold, take $N$ sufficiently large for the reconstruction and MEQ stability conditions in Appendix~\ref{app:finite_meq}, and suppose all scalarized solutions used to construct $P_t$ are exact. Use the finite MEQ construction and numerical accuracy targets in Appendix~\ref{app:finite_meq}. Then, at every outer round from initialization through the last returned round of Algorithm~\ref{alg:surface},
\begin{align}
\|\rho_t-\rho_\vartheta\|_\infty
\le{}&
(1-\alpha)^t
\|\rho_0-\rho_\vartheta\|_\infty
+
C\alpha N^{-2/(M-1)}
\sum_{j=0}^{t-1}
(1-\alpha)^{t-1-j}C_j
\nonumber\\
\le{}&
(1-\alpha)^t
\|\rho_0-\rho_\vartheta\|_\infty
+
C N^{-2/(M-1)}.
\label{eq:density_rate}
\end{align}
Moreover, there is 
\begin{align}
\widehat{\mathcal R}_\vartheta(W^{(t)})
&\le
C(1-\alpha)^t
\|\rho_0-\rho_\vartheta\|_\infty
+
C N^{-2/(M-1)},
\label{eq:target_residual_rate}\\
\|P_t-\fPF\|_{C^1}
&\le
C C_t N^{-2/(M-1)}.
\label{eq:chart_rate}
\end{align}
\end{theorem}

\begin{proof}
Lemma~\ref{lem:round_admissibility} shows that all exact rounds remain in the regular reconstruction and MEQ regime. Lemmas~\ref{lem:chart_reconstruction} and~\ref{lem:area_density_stability} give $\left\|
\frac{\vartheta(P_t)J_{P_t}}
{\int_{\Delta_M^\delta}\vartheta(P_t)J_{P_t}}
-
\rho_\vartheta
\right\|_\infty
\le
C C_t N^{-2/(M-1)}$. Using~\eqref{eq:density_update},
\begin{align}
\|\rho_{t+1}-\rho_\vartheta\|_\infty
\le
(1-\alpha)\|\rho_t-\rho_\vartheta\|_\infty
+
C\alpha C_tN^{-2/(M-1)}.
\label{eq:exact_outer_recursion}
\end{align}
Iterating this recursion gives the first inequality in~\eqref{eq:density_rate}. Lemma~\ref{lem:round_admissibility} gives $C_j\le C$, so the geometrically weighted sum is uniformly bounded and yields the second inequality.

The sum in~\eqref{eq:density_rate} shows how early reconstruction errors affect later rounds. An initially large $C_j$ receives a geometrically decreasing weight, while~\eqref{eq:round_resolution_from_mass} bounds subsequent coefficients for each fixed preference. Under the additional PF-diameter condition following~\eqref{eq:round_reconstruction_factor}, later $C_j$ have a $\kappa_\vartheta$-independent bound; otherwise a dependence through the mesh may remain. Equation~\eqref{eq:chart_rate} follows from Lemma~\ref{lem:chart_reconstruction}, and Proposition~\ref{prop:target_mass_residual} combined with~\eqref{eq:density_rate} gives~\eqref{eq:target_residual_rate}.
\end{proof}

\subsection{Inexact Scalarized Solves}
\label{app:proofs_inexact}

Following the finite-step inner-solver analysis of SURF~\citep{jiang2026surf}, we characterize the work of $\textsc{InnerSolver}_K$ by its contraction factor $\eta_K$. When the solve at round $t+1$ is warm-started from the returned state at round $t$, we assume
\begin{align} \left\|\tilde\bx_n^{(t+1)}-\bx_{\rm TS}^\star(\bw_n^{(t+1)})\right\| \le \eta_K\left\|\tilde\bx_n^{(t)}-\bx_{\rm TS}^\star(\bw_n^{(t+1)})\right\|,\quad n\in[N]. \label{eq:warm_start_contraction} \end{align}
The factor $\eta_K$ measures how much $K$ inner steps reduce decision error at a fixed weight. We require~\eqref{eq:warm_start_contraction} at every warm start encountered by SURFACE. For linearly convergent methods, $\eta_K$ decreases geometrically with $K$. Sublinear methods fit this analysis when their guarantee yields the same multiplicative decision-error bound. Projected gradient provides one example in the hidden strongly convex weighted-sum setting~\citep{jiang2026surf}. For TS, the local strong regularity and quadratic growth in Assumption~\ref{ass:tch_route} support locally contractive methods when the initialization lies in their convergence neighborhood~\citep{bonnans2013perturbation}.

{Damping limits how far the weights move between rounds and therefore how much the exact scalarized solutions change. Reconstruction then converts errors in the returned solutions into errors in the estimated density. By Lemma~\ref{lem:chart_reconstruction}, one, two, and three inverse powers of mesh spacing amplify sample error in $C^1$, $C^2$, and $C^{2,1}$, respectively. We choose the solver accuracy to keep the $C^1$ error small enough to preserve tangent rank and the accumulated density error small enough for the two-stage transport. The initial solves satisfy}
\begin{align} \max_{n\in[N]}\left\|\tilde\bx_n^{(0)}-\bx_{\rm TS}^\star(\bw_n^{(0)})\right\| \le C_xN^{-3/(M-1)}, \label{eq:initial_inner_accuracy} \end{align}
and we choose the inner budget so that
{\begin{align}
\eta_K=\Oc\left(\min\left\{(1+\alpha N^{1/(M-1)})^{-1},\,\alpha^{-2}N^{-3/(M-1)}\right\}\right).
\label{eq:warm_start_accuracy}
\end{align}}
Here $C_x>0$ is any fixed constant, independent of $N$, $T$, and $\alpha$. Choose $K$ to meet the fixed transport and contraction bounds below. In particular, $\bar\alpha^2\eta_K=\Oc(N^{-3/(M-1)})$ with an admissible constant implies~\eqref{eq:warm_start_accuracy} when $\bar\alpha=\Omega(N^{-3/(2(M-1))})$. Exact solves correspond to $\eta_K=0$.

The next lemma shows that this initialization and inner budget keep the solver error, density bounds, and mesh regularity under control at every round.

\begin{lemma}[Invariant inexact trajectory]
\label{lem:inexact_admissibility}
Suppose Assumptions~\ref{ass:tch_route}--\ref{ass:preference_profile} hold, $0<\alpha\le\bar\alpha<1$ with $\bar\alpha=\Omega(N^{-3/(2(M-1))})$, and~\eqref{eq:initial_inner_accuracy}--\eqref{eq:warm_start_accuracy} hold. Initialize and retain or refine the pinned-corner data as in~\eqref{eq:corner_initial_accuracy}. Use the moving-weight Moser realization and numerical targets in Appendix~\ref{app:finite_meq}.
For every fixed $C_x$, choose the constants in the inner-solver and numerical MEQ accuracy bounds independently of $N$, $T$, and $\alpha$. Then, for all sufficiently large $N$ and every outer round of Algorithm~\ref{alg:surface},
\begin{align}
\max_{n\in[N]}\big\|\tilde\bx_n^{(t)}-\bx_{\rm TS}^\star(\bw_n^{(t)})\big\|
\le C_*\min\{N^{-1/(M-1)},\,\alpha^{-1}N^{-3/(M-1)}\},
\quad C_t\le C.
\label{eq:inexact_invariants}
\end{align}
Moreover, $P_t$ has full tangent rank, $\rho_t$ has common positive lower and upper bounds and satisfies the scale-dependent condition in Proposition~\ref{prop:finite_moser_realization}, and $W^{(t)}$ satisfies Lemma~\ref{lem:mesh_invariance}.
\end{lemma}

\begin{proof}
Write $h=N^{-1/(M-1)}$ and let $e_t$ denote the maximum adaptive decision error at round $t$. Fix the density and mesh bounds as in Lemma~\ref{lem:round_admissibility}. Take $N$ sufficiently large for the fixed initialization constant $C_x$ and choose $K$ and numerical MEQ tolerances to meet the transport and contraction estimates.
At each round, compare the actual chart with the reconstruction from exact objective values at the same adaptive weights, retaining the current pinned-corner data. Starting at $g_0=\rho_0$, apply the damped update to these comparison charts to define $g_t$. Lemmas~\ref{lem:chart_reconstruction} and~\ref{lem:area_density_stability} give the fresh comparison densities a common $C^{1,1}$ bound and $\Oc(h)$ proximity to $\rho_\vartheta$ in $C^1$.
Hence $g_t$ remains in the fixed regular class.

We jointly maintain the stated decision-error invariant and $e_t\le q^t e_0+C\eta_K\alpha$ for a fixed $q<1/2$. Both hold initially, with $\rho_0=g_0$ and the regular reference mesh. Suppose they hold through round $t$. Reconstruction gives
\begin{align}
\|P_t-\fPF\|_{C^1}\le Ch^2+Ce_t/h,\quad\text{and},\quad
\|P_t\|_{C^{2,1}}\le C(1+e_t/h^3).
\label{eq:inexact_chart_c21_quantitative}
\end{align}
Since $e_t\le 2^{-t}C_xh^3+C\eta_K\alpha$ and~\eqref{eq:warm_start_accuracy} gives $\eta_K\alpha/h\le C\min\{\alpha/h,h^2/\alpha\}\le C h^{1/2}$, we have $e_t/h=\Oc(h^{1/2})$. Thus, the reconstructed density retains full tangent rank and positive bounds for all sufficiently large $N$, for any fixed $C_x$.

The scaled interpolation estimates and the chain and product rules bound the difference between actual and comparison densities by $Ce_t/h$, $Ce_t/h^2$, and $Ce_t/h^3$ in $C^0$, $C^1$, and $C^{1,1}$, respectively.
The quadratic second-derivative term is absorbed using $e_t^2/h^4\le C_*e_t/h^3$. Unrolling gives
\begin{align}
&\|\rho_{t+1}-g_{t+1}\|_\infty+h\|\rho_{t+1}-g_{t+1}\|_{C^1}
+h^2\|\rho_{t+1}-g_{t+1}\|_{C^{1,1}}\nonumber\\
&\qquad\le \frac{C\alpha}{h}\sum_{s=0}^t(1-\alpha)^{t-s}e_s
\le\frac{C}{h}(\alpha e_0+\eta_K\alpha)=\Oc(\sqrt h).
\label{eq:damped_density_regularity}
\end{align}
Here $\alpha e_0\le C_xh^3$ and $\eta_K\alpha\le C h^{3/2}$ follow from~\eqref{eq:initial_inner_accuracy}--\eqref{eq:warm_start_accuracy}. The initialization contributes $\Oc(C_xh^2)$ to the scaled density error, which is $o(h^{1/2})$. The fixed inner-solver accuracy is chosen to meet~\eqref{eq:moser_scale_condition}, with regular component $g_{t+1}$.

Proposition~\ref{prop:finite_moser_realization} and Lemma~\ref{lem:mesh_invariance} therefore give the moving next mesh in the fixed mesh class, so $C_{t+1}\le C$. Damping preserves the positive lower and upper bounds of the reconstructed densities. Retain the same reference vertices, cover, and divergence operators at every round. The stability of the selected exact Moser map and the $\Oc(\alpha h^2)$ numerical map-value tolerance give
\begin{align}
\max_{n\in[N]}\|\bw_n^{(t+1)}-\bw_n^{(t)}\|
\le C\|\rho_{t+1}-\rho_t\|_\infty+C\alpha h^2.
\label{eq:weight_motion}
\end{align}
The density update and reconstruction bound give
\begin{align}
\|\rho_{t+1}-\rho_t\|_\infty
\le C\alpha\bigl(\|\rho_t-\rho_\vartheta\|_\infty+C_th^2+e_t/h\bigr).
\label{eq:density_motion}
\end{align}
By the uniform density bounds, Lipschitz sensitivity of $\bx_{\rm TS}^\star$, and the warm-start contraction~\eqref{eq:warm_start_contraction},
\begin{align}
e_{t+1}\le\eta_K\bigl[(1+C\alpha/h)e_t+C\alpha+C\alpha C_th^2\bigr]
\le q e_t+C\eta_K\alpha.
\label{eq:invariant_inner_recursion}
\end{align}
Choose the fixed inner-solver accuracy in~\eqref{eq:warm_start_accuracy} so that $q\le1/2$. Then $e_t\le2^{-t}e_0+C\eta_K\alpha$. If $\alpha\le h^2$, then $\eta_K\alpha=\Oc(h)$; otherwise the second condition gives $\eta_K\alpha=\Oc(h^3/\alpha)$. Therefore $e_t\le C_*\min\{h,\alpha^{-1}h^3\}$ for a fixed $C_*$, closing the joint induction.
\end{proof}

Lemma~\ref{lem:inexact_admissibility} gives mesh regularity and the weight-motion bound along the moving Moser trajectory. The same bounds control the coupled recursion below.

\begin{lemma}[One-step coupled contraction]
\label{lem:main_one_step}
Under the conditions of Lemma~\ref{lem:inexact_admissibility}, there exist constants $C_0,C>0$, independent of $N$, $T$, and $\alpha$, such that, for all sufficiently large $N$,
\begin{align} \mathcal E_{t+1}\le(1-\alpha/4)\mathcal E_t+C\alpha C_tN^{-2/(M-1)}. \label{eq:main_one_step} \end{align}
\end{lemma}

\begin{proof}
Equation~\eqref{eq:main_reconstruction_inexact} and the density update give
\begin{align} 
\|\rho_{t+1}-\rho_\vartheta\|_\infty
\le &(1-\alpha)\|\rho_t-\rho_\vartheta\|_\infty+C\alpha C_tN^{-2/(M-1)} \nonumber\\
& +C\alpha N^{1/(M-1)}\max_{n\in[N]}\big\|\tilde\bx_n^{(t)}-\bx_{\rm TS}^\star(\bw_n^{(t)})\big\|. \label{eq:inexact_outer_recursion} \end{align}
{Equations~\eqref{eq:weight_motion}--\eqref{eq:density_motion}, Lipschitz continuity of $\bx_{\rm TS}^\star$, and~\eqref{eq:warm_start_contraction} give the inner recursion below. The numerical motion $C\alpha N^{-2/(M-1)}$ is absorbed into its last term because $C_t\ge1$.}
\begin{align} \max_{n\in[N]}\left\|\tilde\bx_n^{(t+1)}-\bx_{\rm TS}^\star(\bw_n^{(t+1)})\right\| \le \eta_K\Big(
\big(1+C\alpha N^{1/(M-1)}\big)\max_{n\in[N]}\big\|\tilde\bx_n^{(t)}-\bx_{\rm TS}^\star(\bw_n^{(t)})\big\|\nonumber\\
+C\alpha\|\rho_t-\rho_\vartheta\|_\infty+C\alpha C_tN^{-2/(M-1)}\Big). \label{eq:inner_recursion} \end{align}
Multiply~\eqref{eq:inner_recursion} by $C_0(1+\alpha N^{1/(M-1)})$ and add it to~\eqref{eq:inexact_outer_recursion}. Choose $C_0$ to absorb the direct solver-error term and $K$ to meet the coupled-contraction margin in~\eqref{eq:warm_start_accuracy}. The density component then has coefficient at most $1-\alpha/2$, and the solver component has coefficient at most $1-\bar\alpha/4\le1-\alpha/4$. The remaining terms are bounded by $C\alpha C_tN^{-2/(M-1)}$, which proves~\eqref{eq:main_one_step}.
\end{proof}

\begin{proof}[Proof of Theorem~\ref{thm:surface_inexact}]
Use the moving-weight Moser realization in Proposition~\ref{prop:finite_moser_realization}.
Lemma~\ref{lem:inexact_admissibility} gives $C_j\le\overline C_{\mathrm{mesh}}$ for all rounds, where $\overline C_{\mathrm{mesh}}$ is independent of $j$ and $N$ for each fixed regular problem.
Take $N=\Omega(\kappa_\vartheta^{3(M-1)/2})$ sufficiently large for the reconstruction and Moser mesh-stability thresholds in Appendix~\ref{app:finite_meq}. Iterating~\eqref{eq:main_one_step} gives
\begin{align}
\mathcal E_t\le &(1-\alpha/4)^t\mathcal E_0+C\alpha N^{-2/(M-1)}\sum_{j=0}^{t-1}(1-\alpha/4)^{t-1-j}C_j\nonumber\\
\leq & (1-\alpha/4)^t\mathcal E_0+4C\overline C_{\mathrm{mesh}}N^{-2/(M-1)}.
\end{align}
This is the claimed bound on the coupled error. Proposition~\ref{prop:target_mass_residual} then gives
\begin{align} \widehat{\mathcal R}_\vartheta(W^{(t)})\le C\|\rho_t-\rho_\vartheta\|_\infty+C N^{-2/(M-1)}\le C\mathcal E_t+C N^{-2/(M-1)}, \end{align}
Combining this residual bound with~\eqref{eq:round_resolution_from_mass} shows that the residual-dependent part of the upper bound on $C_t$ contracts at the linear rate $(1-\alpha/4)^t$. After $t=\Oc(\alpha^{-1}\log N)$ rounds, $\widehat{\mathcal R}_\vartheta(W^{(t)})=\Oc(N^{-2/(M-1)})$ and hence $C_t=\Oc(1+N^{-2/(M-1)})=\Oc(1)$ uniformly over subsequent rounds.
This proves the target-mass bound.
\end{proof}

{For fixed $\kappa_\vartheta$, choose $N=\Theta(\epsilon^{-(M-1)/2})$ sufficiently large for the regularity thresholds. Under the initialization in~\eqref{eq:initial_inner_accuracy}, Lemmas~\ref{lem:inexact_admissibility} and~\ref{lem:main_one_step} give an $\Oc(\epsilon)$ error floor after $T=\Oc(\alpha^{-1}\log\epsilon^{-1})$ outer rounds.} For a linearly convergent $\textsc{InnerSolver}_K$ with $\eta_K\le q^K$, $q\in(0,1)$, the three damping regimes in Table~\ref{tab:outer_inner_regimes} yield repeated work per adaptive weight of $\Oc(\log^2\epsilon^{-1})$, $\Oc(\epsilon^{-1/2}\log^2\epsilon^{-1})$, and $\Oc(\epsilon^{-3/4}\log\epsilon^{-1})$, respectively, favoring constant damping.
For $\eta_K=\Oc(K^{-p})$, the corresponding repeated-work bounds are $\widetilde{\Oc}(\epsilon^{-3/(2p)})$, $\widetilde{\Oc}(\epsilon^{-1/2-1/(2p)})$, and $\widetilde{\Oc}(\epsilon^{-3/4})$. The leading powers favor constant damping for $p>2$ and fixed inner work for $p<2$, with a tie at $p=2$.
The theorem's $\Oc(\epsilon^{3/2})$ initialization requires $\Oc(\epsilon^{-3/(2p)})$ one-time work per weight in every regime. Including warmup, the repeated-work advantage of smaller damping disappears in the leading exponent for $p\le2$.
These bounds apply to the $N$ adaptive weights. The $M$ fixed corners additionally require $\Oc(M\log\epsilon^{-1})$ initial work for linear convergence or $\Oc(M\epsilon^{-3/(2p)})$ for sublinear convergence if accurate states are not supplied, plus subsequent refinements. Aggregate adaptive inner work scales with $N$, while reconstruction, quadrature, and transport costs are counted separately.

\section{Experiment Details}
\label{app:benchmarks}

This section gives the additional experimental details for Section~\ref{sec:experiments}.

\subsection{Experimental Protocol and Benchmark Scope}
\label{app:exp_final_recipe}
\label{app:exp_protocol}
\label{app:exp_baselines}

\paragraph{Baselines.}
Uniform weighting uses the same scalarized solver and returned cardinality as SURFACE, providing a direct comparison between fixed and geometry-adaptive scalarization weights. We further compare with several classes of finite-PF methods. \emph{MOEA/D}~\citep{zhang2007moead} decomposes the MOO problem using a fixed set of weight vectors, while \emph{MOEA/D-AWA}~\citep{qi2014moea} adaptively removes weights in crowded regions and adds weights toward sparse PF regions. \emph{DEA-GNG}~\citep{liu2019adapting} also adapts preference vectors, using a growing neural gas model to capture the evolving PF geometry. \emph{NSGA-III}~\citep{deb2014evolutionary} maintains population diversity through reference-point association, whereas \emph{SMS-EMOA}~\citep{beume2007smsemoa} uses hypervolume contribution for environmental selection. \emph{LMPFE}~\citep{tian2022local} approximates the PF through multiple local models, and \emph{Subset selection}~\citep{shang2023benchmarking} selects a fixed-cardinality representation from a larger candidate set. Finally, \emph{UMOD}~\citep{zhang2024umod} directly targets representativeness through a max-packing design, which serves as a surrogate for minimizing PF fill distance.
All methods use matched returned cardinalities, with five seeds for stochastic baselines.

\paragraph{Benchmarks.}
We evaluate SURFACE on problems with $M\in\{2,3,4,6\}$ objectives. The synthetic suite includes ZDT1/2/4~\citep{zitzler2000comparison} and DTLZ2/3/4~\citep{deb2002scalable}. ZDT1/2/4 have smooth, connected bi-objective PFs on the represented interior branch. ZDT2 changes the PF curvature, while ZDT4 adds multiple local optima in decision space. DTLZ2/3/4 share the same smooth spherical PF of dimension $M-1$. DTLZ3 adds multiple local optima, and DTLZ4 biases the map from decisions to PF points. Appendix~\ref{app:tch_sufficient_regimes} verifies the regular full-ray TS conditions for DTLZ2.
We also evaluate the engineering benchmark RE21~\citep{tanabe2020easy} using its empirical reference front, and the policy problems Deep Sea Treasure, Resource Gathering, Fruit Tree, and Minecart from MO-Gymnasium~\citep{felten2023toolkit}. These tasks test the method on analytic fronts, empirical reference fronts, and optimized policies.

\paragraph{Metrics.}
We measure how evenly the returned solutions allocate target mass, how closely they cover the PF, and how much objective-space volume they dominate. The primary metric is the \textbf{finite target-mass residual} $\widehat{\mathcal R}_\vartheta(W)$ $(\downarrow)$ from Lemma~\ref{lem:preference_pullback}, evaluated on the selected cells associated with each returned set. We use connected mesh cells for transport diagnostics and PF-area-weighted Voronoi cells for cross-method comparisons. 
For $\vartheta=1$, it measures imbalance in PF surface-area allocation.
\textbf{IGD} $(\downarrow)$ measures the average distance from the PF to the nearest returned objective vector. The \textbf{fill distance} $(\downarrow)$ measures the largest such distance~\citep{ishibuchi2015study,zhang2024umod}. We also report \textbf{hypervolume} HV $(\uparrow)$~\citep{beume2007smsemoa,shang2020survey}, the dominated objective-space volume relative to a fixed task-specific reference point. Its normalized \textbf{hypervolume gap} $\Delta\mathrm{HV}:=1-\mathrm{HV}/\mathrm{HV}_{\rm ref}$ $(\downarrow)$ uses the reference PF hypervolume $\mathrm{HV}_{\rm ref}$ computed with the same task-specific reference point.

\paragraph{Inner solvers.}
Within each task, SURFACE and Uniform use the same inner solver and return the same number of solutions. The controlled DTLZ2 resolution studies use exact scalarized solutions. The quadratic experiment uses warm-started gradient descent to measure the effect of inner-solver error. For Deep Sea Treasure, Resource Gathering, and Fruit Tree, we use deterministic KL-regularized soft-value iteration with tolerance $10^{-11}$. For Minecart, both methods use a common frozen GPI-LS model and $2{,}016$ evaluation episodes. We run each stochastic baseline with five seeds and the same returned-set size.

\paragraph{Computational budgets.}
All methods return the same number of Pareto solutions. SURFACE and Uniform use identical inner solvers with matched iteration budgets: SURFACE performs $K$ inner steps per weight over $T$ rounds, while Uniform performs $KT$ steps per fixed weight, with auxiliary corner solves counted consistently. For analytic or high-accuracy inner solvers, we account for numerical solve costs and SURFACE's reconstruction and transport overhead. For evolutionary and other stochastic baselines with different optimization procedures, we compare end-to-end runtime and evaluation costs, including wall-clock-budget comparisons on Resource Gathering.

\subsection{Controlled Validation}
\label{app:exp_dtlz2_details}
\label{app:exp_transport_certificates}
\label{app:exp_transport_limit}
\label{app:exp_reconstruction_diagnostics}

\paragraph{Controlled PF distortion.}
We first use the three-objective radial problem in~\eqref{eq:radial_moo_problem} to isolate the effect of scalarization-induced PF distortion. We vary
$\kappa\in\{0,0.5,1,1.5,2,2.5,3\}$ and normalize the analytic density using four independently scrambled Sobol rules with $8{,}192$ points each. The mesh contains 66 returned vertices, including three pinned corners, and 100 cells. SURFACE uses $T=5$ outer rounds and $\alpha=0.2$. Across the sweep, the measured distortion ratio $\overline J/\underline J$ increases from $1$ to $299.53$. Uniform weighting reaches $\widehat{\mathcal R}_\vartheta=8.901$, whereas SURFACE remains below $0.785$ throughout the sweep.
Supplying the analytic PF-area density to the same transport solver gives $\widehat{\mathcal R}_\vartheta=0.0337$, {providing an analytic-density reference.}

\paragraph{Finite-resolution error.}
Figure~\ref{fig:controlled_rates} compares the measured errors with the predicted $\Oc(N^{-2/(M-1)})$ scale. In the figure, free probes are the $N$ adaptive weights, and curved-mass residual measures target-mass imbalance on complete transported cells. For DTLZ2 with $M=3$, exact scalarized solves and one undamped update reduce the chart error $\|P_T-\fPF\|_{C^1}$, density error $\mathcal E_T$, and residual $\widehat{\mathcal R}_\vartheta(W_T)$ from $(0.2368,0.3300,0.04909)$ at $N=12$ to $(0.02912,0.01934,0.00652)$ at $N=150$. This decrease is consistent with the $\Oc(N^{-1})$ scale in Lemma~\ref{lem:main_reconstruction} and the exact-solve result, Theorem~\ref{thm:surface}. For $M=6$, the errors also decrease with $N$, and the chart-error slope is close to the predicted $\Oc(N^{-2/5})$ scale.
The analytic quadratic experiment separately measures the effect of inexact solves. Exact solutions give $\widehat{\mathcal R}_\vartheta=0.00253$, compared with $0.320$ for Uniform. With 16 warm-started gradient steps, the decision error falls to $1.24\times10^{-9}$ and $\widehat{\mathcal R}_\vartheta<0.00254$, approaching the exact-solve result.

\paragraph{Finite MEQ validation.}
We check numerical MEQ by independently evaluating transported cell masses.
{For $M=3$, the refined Bernstein--Moser configuration has worst cell-mass error $0.0395$, compared with $0.0575$ for the standard configuration.} For $M=4$, all tested transports at $N\in\{6,16,31,52,80\}$ adaptive weights remain nondegenerate. At the finest resolution, the cell-mass error is $0.0354$, the minimum orientation ratio is $0.535$, and the maximum condition number is $7.07$. The orientation ratio is the signed volume of a transported straight simplex divided by that of its reference simplex. A positive ratio indicates that the simplex has retained its orientation. The condition number is the ratio of the largest to smallest singular value of the affine map from the reference simplex to the transported simplex. The corresponding PF-area residuals on complete transported cells are $0.160$, $0.155$, $0.185$, $0.207$, and $0.207$.
These measurements check the mass accuracy and mesh geometry in Proposition~\ref{prop:main_meq}. Appendix~\ref{app:numerical_meq} specifies the numerical tolerances.

\paragraph{Reconstruction accuracy.}
To examine how PF reconstruction affects density estimation, we reuse the saved samples at $N\in\{16,31,52\}$. We measure chart and density errors by root-mean-square (RMS) values over evaluation points. Radial reconstruction fits the scalar multiplier of each reference direction and combines it with that direction and the reference point to reconstruct the objective vector. At $N=31$, radial reconstruction reduces the PF reconstruction error from $0.0646$ to $0.0267$ and the density error from $0.671$ to $0.127$. At $N=52$, it reduces the density error from $0.377$ to $0.0942$ and the independently evaluated cell-mass error from $0.0479$ to $0.0196$. These results illustrate the reconstruction-to-density relation in Lemma~\ref{lem:main_reconstruction}. A fixed piecewise-linear density approximation adds interpolation error at finer resolutions, so we evaluate the reconstructed density directly.

\subsection{Details of ZDT and DTLZ Benchmark Results}
\label{app:exp_complete_tables}

The ZDT1/2/4 and DTLZ2/3/4 comparison contains 11 tasks with $M=2,3,4$. Every method returns $N+M$ solutions. SURFACE and Uniform share the same scalarized solver, so their comparison isolates weight redistribution. The other baselines use their respective optimization procedures, with five seeds for each stochastic method.  Table~\ref{tab:synthetic_mean_ranks_main} gives the mean ranks, Table~\ref{tab:synthetic_full_results_rewrite} gives the per-task results, and Figure~\ref{fig:synthetic_rank_tradeoff_rewrite} compares PF-coverage and hypervolume ranks.

\begin{wraptable}{r}{0.46\textwidth}
\vspace{-0.5cm}
\centering
\small
\setlength{\tabcolsep}{2.2pt}
\renewcommand{\arraystretch}{1.03}
\begin{tabular}{@{}lrrrrr@{}}
\toprule
Method &
$\widehat{\mathcal R}_\vartheta$ &
$\mathrm{IGD}$ &
Fill &
$\Delta\mathrm{HV}$ &
Coverage \\
\midrule
SURFACE & \textbf{1.18} & \textbf{1.18} & \textbf{1.64} & \textbf{2.55} & \textbf{1.33} \\
Uniform & 3.18 & 2.27 & 2.45 & 4.73 & 2.64 \\
NSGA-III & 5.32 & 4.91 & 4.45 & 5.00 & 4.89 \\
LMPFE & 4.91 & 5.00 & 5.09 & 6.18 & 5.00 \\
SMS-EMOA & 5.00 & 7.00 & 7.09 & 3.91 & 6.36 \\
MOEA/D-AWA & 7.27 & 6.82 & 6.55 & 7.27 & 6.88 \\
Subset & 6.18 & 6.45 & 7.00 & 2.82 & 6.55 \\
MOEA/D & 7.09 & 6.82 & 6.00 & 6.73 & 6.64 \\
DEA-GNG & 6.64 & 6.73 & 7.64 & 8.00 & 7.00 \\
UMOD & 8.23 & 7.82 & 7.09 & 7.82 & 7.71 \\
\bottomrule
\end{tabular}
\vspace{-0.3cm}
\caption{Mean ranks over 11 ZDT/DTLZ task-dimension groups. Coverage averages the first three PF-representation metrics and excludes hypervolume. Lower is better.}
\label{tab:synthetic_mean_ranks_main}
\vspace{-0.4cm}
\end{wraptable}%

SURFACE has the best mean ranks in PF-area imbalance, IGD, fill distance, and aggregate PF coverage over the 11 tasks in Table~\ref{tab:synthetic_mean_ranks_main}. These measures capture both the balance of allocated PF area and the distances from the PF to the returned solutions. Relative to Uniform, SURFACE improves PF-area imbalance, IGD, and hypervolume gap on all listed tasks. It also improves fill distance except on the two $M=4$ tasks, where Uniform gives $0.380$ and SURFACE gives $0.381$. The best individual method varies by task. On DTLZ2 with $M=4$, LMPFE and DEA-GNG have lower PF-area imbalance than SURFACE, and DEA-GNG has lower IGD.

SURFACE also has the best mean $\Delta\mathrm{HV}$ rank over the 11 groups. Other methods achieve better hypervolume on individual tasks, as shown in Table~\ref{tab:synthetic_full_results_rewrite}. Hypervolume rewards dominated objective-space volume, while uniform PF coverage rewards balanced representation of the surface. Figure~\ref{fig:synthetic_rank_tradeoff_rewrite} shows the resulting rank comparison, with SURFACE combining the lowest mean PF-coverage rank and the lowest mean hypervolume-gap rank.

\begin{table}[t]
\centering
\tiny
\renewcommand{\arraystretch}{1.02}
\resizebox{\linewidth}{!}{%
\begin{tabular}{@{}l|r|r|r|r|r|r|r|r|r|r@{}}
\toprule
Metric
& SMS-EMOA
& NSGA-III
& LMPFE
& Subset
& MOEA/D-AWA
& MOEA/D
& DEA-GNG
& UMOD
& Uniform
& SURFACE \\
\midrule

\multicolumn{11}{@{}l}{\textit{ZDT1, $M=2$, $N=9$}} \\
$\widehat{\mathcal R}_\vartheta\downarrow$ & $.431 \pm .015$ & $.608 \pm .000$ & $.563 \pm .060$ & $.769 \pm .001$ & $.434 \pm .076$ & $.897 \pm .239$ & $.516 \pm .122$ & $.483 \pm .065$ & $.505 $ & $\mathbf{.316}$\\
$\mathrm{IGD}\downarrow$ & $.037 \pm .000$ & $.042 \pm .000$ & $.039 \pm .003$ & $.040 \pm .000$ & $.039 \pm .000$ & $.049 \pm .005$ & $.041 \pm .002$ & $.037 \pm .000$ & $.037 $ & $\mathbf{.035}$\\
Fill$\downarrow$ & $.092 \pm .002$ & $.150 \pm .000$ & $.095 \pm .016$ & $.174 \pm .000$ & $.105 \pm .003$ & $.189 \pm .032$ & $.119 \pm .009$ & $\mathbf{.085 \pm .008}$ & $.106 $ & $.096 $\\
$\Delta\mathrm{HV}\downarrow$ & $.048 \pm .000$ & $.055 \pm .000$ & $.053 \pm .004$ & $.049 \pm .000$ & $.052 \pm .000$ & $.060 \pm .004$ & $.055 \pm .003$ & $.049 \pm .001$ & $.048 $ & $\mathbf{.045}$\\
Coverage$\downarrow$
& $2.67 \pm .38$ & $8.33 \pm .38$ & $5.33 \pm 1.69$ & $8.33 \pm .53$ & $4.33 \pm .62$ & $10.00 \pm .73$ & $7.00 \pm 1.22$ & $2.67 \pm .96$ & $4.33$ & $\mathbf{2.00}$\\

\midrule
\multicolumn{11}{@{}l}{\textit{ZDT2, $M=2$, $N=9$}} \\
$\widehat{\mathcal R}_\vartheta\downarrow$ & $.480 \pm .018$ & $.589 \pm .000$ & $.881 \pm .374$ & $.491 \pm .000$ & $.580 \pm .044$ & $.917 \pm .120$ & $.599 \pm .070$ & $.812 \pm .047$ & $.514 $ & $\mathbf{.397}$\\
$\mathrm{IGD}\downarrow$ & $.039 \pm .000$ & $.038 \pm .000$ & $.048 \pm .013$ & $.039 \pm .000$ & $.038 \pm .000$ & $.040 \pm .000$ & $.039 \pm .001$ & $.041 \pm .000$ & $.037 $ & $\mathbf{.036}$\\
Fill$\downarrow$ & $.126 \pm .003$ & $.089 \pm .000$ & $.156 \pm .068$ & $.122 \pm .000$ & $.088 \pm .000$ & $.102 \pm .012$ & $.106 \pm .011$ & $.101 \pm .003$ & $.087 $ & $\mathbf{.073}$\\
$\Delta\mathrm{HV}\downarrow$ & $\mathbf{.069 \pm .000}$ & $.076 \pm .000$ & $.097 \pm .025$ & $.075 \pm .000$ & $.076 \pm .001$ & $.095 \pm .003$ & $.076 \pm .002$ & $.084 \pm .001$ & $.078 $ & $.074 $\\
Coverage$\downarrow$
& $5.67 \pm .55$ & $4.33 \pm .43$ & $9.67 \pm 2.83$ & $5.33 \pm .61$ & $4.00 \pm .43$ & $8.00 \pm 1.09$ & $7.00 \pm 1.51$ & $7.33 \pm .33$ & $2.67$ & $\mathbf{1.00}$\\

\midrule
\multicolumn{11}{@{}l}{\textit{ZDT4, $M=2$, $N=9$}} \\
$\widehat{\mathcal R}_\vartheta\downarrow$ & $.418 \pm .006$ & $.611 \pm .003$ & $.572 \pm .033$ & $.760 \pm .005$ & $.612 \pm .167$ & $1.618 \pm .044$ & $.862 \pm .739$ & $2.124 \pm .972$ & $.505 $ & $\mathbf{.316}$\\
$\mathrm{IGD}\downarrow$ & $.037 \pm .000$ & $.043 \pm .000$ & $.038 \pm .000$ & $.040 \pm .000$ & $.040 \pm .001$ & $.056 \pm .000$ & $.049 \pm .012$ & $.108 \pm .060$ & $.037 $ & $\mathbf{.035}$\\
Fill$\downarrow$ & $.090 \pm .001$ & $.150 \pm .000$ & $\mathbf{.088 \pm .004}$ & $.173 \pm .001$ & $.111 \pm .006$ & $.281 \pm .007$ & $.178 \pm .103$ & $.324 \pm .155$ & $.106 $ & $.096 $\\
$\Delta\mathrm{HV}\downarrow$ & $.049 \pm .000$ & $.056 \pm .001$ & $.053 \pm .002$ & $.050 \pm .000$ & $.055 \pm .003$ & $.075 \pm .007$ & $.065 \pm .012$ & $.196 \pm .122$ & $.048 $ & $\mathbf{.045}$\\
Coverage$\downarrow$
& $2.33 \pm .15$ & $6.00 \pm .49$ & $3.00 \pm .51$ & $6.67 \pm .60$ & $5.33 \pm .90$ & $9.00 \pm .69$ & $8.00 \pm 1.59$ & $10.00 \pm .33$ & $3.00$ & $\mathbf{1.67}$\\

\midrule
\multicolumn{11}{@{}l}{\textit{DTLZ2, $M=2$, $N=9$}} \\
$\widehat{\mathcal R}_\vartheta\downarrow$ & $.503 \pm .008$ & $.613 \pm .000$ & $.587 \pm .065$ & $.623 \pm .001$ & $.673 \pm .078$ & $.613 \pm .000$ & $.691 \pm .068$ & $.810 \pm .092$ & $.543 $ & $\mathbf{.402}$\\
$\mathrm{IGD}\downarrow$ & $.045 \pm .000$ & $.041 \pm .000$ & $.041 \pm .001$ & $.048 \pm .000$ & $.041 \pm .000$ & $.041 \pm .000$ & $.043 \pm .003$ & $.043 \pm .002$ & $.040 $ & $\mathbf{.038}$\\
Fill$\downarrow$ & $.141 \pm .001$ & $.099 \pm .000$ & $.106 \pm .012$ & $.148 \pm .000$ & $.099 \pm .000$ & $.099 \pm .000$ & $.121 \pm .021$ & $.102 \pm .017$ & $.097 $ & $\mathbf{.078}$\\
$\Delta\mathrm{HV}\downarrow$ & $\mathbf{.061 \pm .000}$ & $.082 \pm .000$ & $.079 \pm .004$ & $.063 \pm .000$ & $.081 \pm .001$ & $.082 \pm .000$ & $.085 \pm .011$ & $.081 \pm .003$ & $.085 $ & $.074 $\\
Coverage$\downarrow$
& $6.67 \pm .30$ & $4.00 \pm .76$ & $4.67 \pm 1.96$ & $9.00 \pm .28$ & $6.33 \pm .68$ & $5.00 \pm .71$ & $8.33 \pm 1.12$ & $7.67 \pm 1.24$ & $2.33$ & $\mathbf{1.00}$\\

\midrule
\multicolumn{11}{@{}l}{\textit{DTLZ2, $M=3$, $N=18$}} \\
$\widehat{\mathcal R}_\vartheta\downarrow$ & $1.312 \pm .038$ & $1.178 \pm .001$ & $.913 \pm .086$ & $1.588 \pm .001$ & $4.469 \pm 1.499$ & $1.180 \pm .000$ & $.896 \pm .067$ & $1.457 \pm .174$ & $1.078 $ & $\mathbf{.693}$\\
$\mathrm{IGD}\downarrow$ & $.164 \pm .001$ & $.137 \pm .000$ & $.134 \pm .003$ & $.162 \pm .000$ & $.334 \pm .117$ & $.137 \pm .000$ & $.132 \pm .003$ & $.142 \pm .004$ & $.132 $ & $\mathbf{.129}$\\
Fill$\downarrow$ & $.334 \pm .008$ & $.272 \pm .000$ & $.296 \pm .027$ & $.333 \pm .000$ & $.796 \pm .306$ & $.273 \pm .000$ & $.319 \pm .015$ & $.309 \pm .032$ & $.265 $ & $\mathbf{.227}$\\
$\Delta\mathrm{HV}\downarrow$ & $\mathbf{.133 \pm .001}$ & $.162 \pm .000$ & $.169 \pm .006$ & $.138 \pm .000$ & $.475 \pm .093$ & $.162 \pm .000$ & $.200 \pm .015$ & $.176 \pm .003$ & $.175 $ & $.162 $\\
Coverage$\downarrow$
& $8.33 \pm .53$ & $4.33 \pm .28$ & $4.00 \pm 1.01$ & $8.33 \pm .37$ & $10.00 \pm .00$ & $5.33 \pm .24$ & $3.67 \pm .71$ & $7.00 \pm .87$ & $3.00$ & $\mathbf{1.00}$\\

\midrule
\multicolumn{11}{@{}l}{\textit{DTLZ2 , $M=4$, $N=31$}} \\
$\widehat{\mathcal R}_\vartheta\downarrow$ & $5.547 \pm 2.009$ & $3.436 \pm .001$ & $\mathbf{1.767 \pm .327}$ & $3.771 \pm .002$ & $9.975 \pm 3.034$ & $3.434 \pm .002$ & $1.827 \pm .478$ & $4.523 \pm .509$ & $3.042 $ & $2.498 $\\ $\mathrm{IGD}\downarrow$ & $.348 \pm .043$ & $.228 \pm .000$ & $.216 \pm .003$ & $.246 \pm .000$ & $.510 \pm .091$ & $.228 \pm .000$ & $\mathbf{.208 \pm .014}$ & $.245 \pm .006$ & $.218 $ & $.216 $\\
Fill$\downarrow$ & $.723 \pm .091$ & $.391 \pm .000$ & $.420 \pm .021$ & $.467 \pm .000$ & $.975 \pm .225$ & $.391 \pm .000$ & $.447 \pm .023$ & $.467 \pm .035$ & $\mathbf{.380}$ & $.381 $\\ $\Delta\mathrm{HV}\downarrow$ & $.493 \pm .050$ & $.183 \pm .000$ & $.196 \pm .004$ & $\mathbf{.166 \pm .000}$ & $.581 \pm .069$ & $.183 \pm .000$ & $.266 \pm .008$ & $.209 \pm .005$ & $.208 $ & $.196 $\\
Coverage$\downarrow$
& $9.00 \pm .94$ & $5.00 \pm .25$ & $\mathbf{2.67 \pm .61}$ & $7.67 \pm .41$ & $10.00 \pm .15$ & $4.67 \pm .48$ & $3.00 \pm 1.26$ & $7.33 \pm .41$ & $3.00$ & $\mathbf{2.67}$\\

\midrule
\multicolumn{11}{@{}l}{\textit{DTLZ3, $M=2$, $N=9$}} \\
$\widehat{\mathcal R}_\vartheta\downarrow$ & $.509 \pm .017$ & $.616 \pm .006$ & $.492 \pm .016$ & $.624 \pm .001$ & $.652 \pm .042$ & $.784 \pm .200$ & $3.273 \pm 3.130$ & $1.471 \pm .399$ & $.543 $ & $\mathbf{.402}$\\
$\mathrm{IGD}\downarrow$ & $.046 \pm .000$ & $.041 \pm .000$ & $.041 \pm .001$ & $.048 \pm .001$ & $.043 \pm .002$ & $.047 \pm .005$ & $.189 \pm .185$ & $.085 \pm .022$ & $.040 $ & $\mathbf{.038}$\\
Fill$\downarrow$ & $.143 \pm .001$ & $.099 \pm .000$ & $.096 \pm .006$ & $.148 \pm .001$ & $.100 \pm .002$ & $.107 \pm .010$ & $.525 \pm .444$ & $.233 \pm .052$ & $.097 $ & $\mathbf{.078}$\\
$\Delta\mathrm{HV}\downarrow$ & $\mathbf{.070 \pm .006}$ & $.088 \pm .003$ & $.092 \pm .007$ & $.079 \pm .013$ & $.106 \pm .025$ & $.151 \pm .040$ & $.307 \pm .231$ & $.289 \pm .113$ & $.085 $ & $.074 $\\
Coverage$\downarrow$
& $5.33 \pm .43$ & $4.33 \pm .40$ & $2.33 \pm .78$ & $7.33 \pm .77$ & $5.67 \pm .88$ & $7.00 \pm 1.09$ & $10.00 \pm 2.37$ & $9.00 \pm .55$ & $3.00$ & $\mathbf{1.00}$\\

\midrule
\multicolumn{11}{@{}l}{\textit{DTLZ3, $M=3$, $N=18$}} \\
$\widehat{\mathcal R}_\vartheta\downarrow$ & $1.490 \pm .402$ & $1.179 \pm .001$ & $2.546 \pm 1.447$ & $1.588 \pm .001$ & $20.000 \pm .000$ & $1.180 \pm .002$ & $4.833 \pm 1.872$ & $3.120 \pm .202$ & $1.078 $ & $\mathbf{.693}$\\
$\mathrm{IGD}\downarrow$ & $.166 \pm .005$ & $.137 \pm .000$ & $.210 \pm .062$ & $.162 \pm .000$ & $143.225 \pm 159.716$ & $.138 \pm .000$ & $.359 \pm .123$ & $.204 \pm .022$ & $.132 $ & $\mathbf{.129}$\\
Fill$\downarrow$ & $.342 \pm .020$ & $.272 \pm .000$ & $.479 \pm .137$ & $.333 \pm .000$ & $143.717 \pm 159.735$ & $.273 \pm .001$ & $.828 \pm .284$ & $.476 \pm .050$ & $.265 $ & $\mathbf{.227}$\\
$\Delta\mathrm{HV}\downarrow$ & $\mathbf{.134 \pm .001}$ & $.162 \pm .000$ & $.258 \pm .071$ & $.139 \pm .002$ & $.939 \pm .083$ & $.172 \pm .010$ & $.432 \pm .129$ & $.266 \pm .021$ & $.175 $ & $.162 $\\
Coverage$\downarrow$
& $5.67 \pm .37$ & $3.00 \pm .15$ & $7.67 \pm 1.10$ & $5.33 \pm .37$ & $10.00 \pm .00$ & $4.00 \pm .15$ & $9.00 \pm .28$ & $7.33 \pm .55$ & $2.00$ & $\mathbf{1.00}$\\

\midrule
\multicolumn{11}{@{}l}{\textit{DTLZ3, $M=4$, $N=31$}} \\
$\widehat{\mathcal R}_\vartheta\downarrow$ & $6.121 \pm 2.177$ & $3.436 \pm .001$ & $3.977 \pm 2.356$ & $3.781 \pm .025$ & $34.00 \pm .000$ & $3.436 \pm .004$ & $7.726 \pm 2.906$ & $4.766 \pm 1.414$ & $3.042 $ & $\mathbf{2.498}$\\ $\mathrm{IGD}\downarrow$ & $.439 \pm .079$ & $.228 \pm .000$ & $.294 \pm .059$ & $.247 \pm .001$ & $73.139 \pm 161.229$ & $.228 \pm .000$ & $.456 \pm .089$ & $.309 \pm .032$ & $.218 $ & $\mathbf{.216}$\\
Fill$\downarrow$ & $.967 \pm .186$ & $.391 \pm .000$ & $.616 \pm .059$ & $.467 \pm .000$ & $73.537 \pm 161.271$ & $.391 \pm .001$ & $.972 \pm .244$ & $.653 \pm .085$ & $\mathbf{.380}$ & $.381 $\\ $\Delta\mathrm{HV}\downarrow$ & $.529 \pm .044$ & $.184 \pm .000$ & $.267 \pm .061$ & $\mathbf{.166 \pm .000}$ & $.908 \pm .051$ & $.184 \pm .000$ & $.498 \pm .127$ & $.267 \pm .030$ & $.208 $ & $.196 $\\
Coverage$\downarrow$
& $8.00 \pm .62$ & $3.17 \pm .47$ & $6.00 \pm 1.01$ & $5.00 \pm .33$ & $10.00 \pm .00$ & $3.83 \pm .30$ & $9.00 \pm 1.64$ & $7.00 \pm .64$ & $1.67$ & $\mathbf{1.33}$\\

\midrule
\multicolumn{11}{@{}l}{\textit{DTLZ4, $M=2$, $N=9$}} \\
$\widehat{\mathcal R}_\vartheta\downarrow$ & $8.101 \pm 4.247$ & $4.368 \pm 5.142$ & $8.112 \pm 4.222$ & $6.249 \pm 5.136$ & $8.140 \pm 4.159$ & $10.000 \pm .000$ & $8.086 \pm 4.279$ & $10.000 \pm .000$ & $.543 $ & $\mathbf{.402}$\\
$\mathrm{IGD}\downarrow$ & $.606 \pm .313$ & $.323 \pm .386$ & $.605 \pm .316$ & $.467 \pm .382$ & $.605 \pm .315$ & $.746 \pm .000$ & $.605 \pm .315$ & $.746 \pm .000$ & $.040 $ & $\mathbf{.038}$\\
Fill$\downarrow$ & $1.160 \pm .569$ & $.625 \pm .721$ & $1.149 \pm .592$ & $.908 \pm .694$ & $1.151 \pm .588$ & $1.414 \pm .000$ & $1.152 \pm .586$ & $1.414 \pm .000$ & $.097 $ & $\mathbf{.078}$\\
$\Delta\mathrm{HV}\downarrow$ & $.602 \pm .302$ & $.344 \pm .359$ & $.604 \pm .296$ & $.468 \pm .368$ & $.606 \pm .293$ & $.737 \pm .000$ & $.604 \pm .297$ & $.737 \pm .000$ & $.085 $ & $\mathbf{.074}$\\
Coverage$\downarrow$
& $7.33 \pm 1.84$ & $3.00 \pm 1.45$ & $5.67 \pm 2.65$ & $4.00 \pm 1.47$ & $7.00 \pm 1.75$ & $9.17 \pm 1.22$ & $6.00 \pm 2.52$ & $9.83 \pm .52$ & $2.00$ & $\mathbf{1.00}$\\

\midrule
\multicolumn{11}{@{}l}{\textit{DTLZ4, $M=3$, $N=18$}} \\
$\widehat{\mathcal R}_\vartheta\downarrow$ & $20.000 \pm .000$ & $20.000 \pm .000$ & $4.698 \pm 8.556$ & $8.954 \pm 10.083$ & $1.253 \pm .549$ & $16.236 \pm 8.417$ & $12.282 \pm 10.568$ & $20.000 \pm .000$ & $1.078 $ & $\mathbf{.693}$\\
$\mathrm{IGD}\downarrow$ & $.943 \pm .000$ & $.943 \pm .000$ & $.296 \pm .361$ & $.475 \pm .427$ & $.150 \pm .019$ & $.782 \pm .360$ & $.615 \pm .449$ & $.943 \pm .000$ & $.132 $ & $\mathbf{.129}$\\
Fill$\downarrow$ & $1.414 \pm .000$ & $1.414 \pm .000$ & $.523 \pm .498$ & $.765 \pm .592$ & $.345 \pm .048$ & $1.186 \pm .511$ & $.959 \pm .623$ & $1.414 \pm .000$ & $.265 $ & $\mathbf{.227}$\\
$\Delta\mathrm{HV}\downarrow$ & $.848 \pm .000$ & $.848 \pm .000$ & $.307 \pm .302$ & $.422 \pm .389$ & $.187 \pm .026$ & $.711 \pm .307$ & $.586 \pm .358$ & $.848 \pm .000$ & $.175 $ & $\mathbf{.162}$\\
Coverage$\downarrow$
& $9.00 \pm 2.11$ & $8.33 \pm 1.03$ & $4.00 \pm 1.80$ & $5.00 \pm 1.09$ & $3.00 \pm 1.32$ & $7.00 \pm 1.98$ & $6.00 \pm 3.19$ & $9.67 \pm .63$ & $2.00$ & $\mathbf{1.00}$\\

\bottomrule
\end{tabular}%
}
\caption{
Complete results for SURFACE and 9 baselines. Stochastic-method entries are mean $\pm$ sample standard deviation over five seeds. Uniform and SURFACE are deterministic and have no standard-deviation terms. Each Coverage row reports the mean rank over $\widehat{\mathcal R}_\vartheta$, IGD, and fill distance.
}
\label{tab:synthetic_full_results_rewrite}

\end{table}

\subsection{Multi-Objective Policy Optimization}
\label{app:exp_policy_details}

We evaluate SURFACE on Deep Sea Treasure, Resource Gathering, and Fruit Tree from MO-Gymnasium~\citep{felten2023toolkit}. For each preference $\bw$, we optimize
\begin{align}
F_{\bw}(\pi)
=
-\sum_{m=1}^M w_mJ_m(\pi)
+
\beta R_{\rm KL}(\pi).
\label{eq:policy_objective_rewrite}
\end{align}
Here $J_m(\pi)=\mathbb E_\pi[\sum_{t\ge0}\gamma^t r_m(s_t,a_t)]$ is the unregularized discounted return, and
\begin{align*}
R_{\rm KL}(\pi)
&:=\mathbb E_\pi\!\left[\sum_{t\ge0}\gamma^t
\operatorname{KL}\bigl(\pi(\cdot\mid s_t)\,\|\,\pi_{\rm ref}(\cdot\mid s_t)\bigr)\right]
=\frac{1}{1-\gamma}\sum_s d_\pi^\gamma(s)
\operatorname{KL}\bigl(\pi(\cdot\mid s)\,\|\,\pi_{\rm ref}(\cdot\mid s)\bigr).
\end{align*}
Here $d_\pi^\gamma(s):=(1-\gamma)\sum_{t\ge0}\gamma^t\Pr_\pi(s_t=s)$ is the discounted state-occupancy distribution, starting from the environment's initial-state distribution. The reference policy $\pi_{\rm ref}$ assigns positive probability to every available action. For the standard entropy-regularized soft-value solver, it is uniform over those actions.
As $\sum_m w_m=1$,~\eqref{eq:policy_objective_rewrite} is the weighted sum of $f_m^\beta(\pi)=-J_m(\pi)+\beta R_{\rm KL}(\pi)$. We evaluate PF-area allocation, IGD, and HV using the unregularized return vectors $(J_m(\pi))_{m=1}^M$.
We use discount factor $0.999$, with $\beta=1.5$ for Deep Sea Treasure and $\beta=0.5$ for Resource Gathering and Fruit Tree. Soft value iteration stops at tolerance $10^{-11}$. SURFACE uses $\alpha=0.3$. Each adaptive weight retains its label across rounds and warm-starts from the preceding solution with that label. We reuse the fixed corner solutions and perform a final solve at the returned weights.

Deep Sea Treasure and Resource Gathering use 30 outer rounds, while Fruit Tree uses ten. They return 16, 21, and 56 policies, respectively. SURFACE and Uniform share the deterministic soft-value solver. The other baselines use an evolutionary optimizer over policy logits, run with five seeds. Table~\ref{tab:mogym_combined_rewrite} reports the comparison.

We further study the effect of returned-set resolution on the $M=6$ Fruit Tree task~\citep{felten2023toolkit}. Policy sensitivities describe how the optimized policies change with the scalarization weights. We use these derivatives to evaluate the return-space geometry, and compare all resolutions using independently sampled, surface-area-weighted return-space Voronoi cells.
Site selection uses nearest-site assignments on a separate fixed weighted oracle reference. SURFACE achieves decreasing target-mass residuals $\widehat{\mathcal R}_\vartheta=0.2523,0.1955,0.1694$ as the returned cardinality increases from $N=126$ to $252$ and $462$.
On this six-objective task, increasing the number of returned policies reduces the measured target-mass imbalance.

Minecart tests SURFACE with a different policy-optimization backend. This continuous-state, discrete-action benchmark is provided by MO-Gymnasium~\citep{felten2023toolkit} and was introduced by~\citet{abels2019dynamic}. Both methods use the same frozen GPI-LS policy set~\citep{alegre2022optimistic}, with $2{,}016$ evaluation episodes per method at temperature $0.03$. The selected weights condition GPI-LS action selection over this fixed policy set, without retraining its members.
 Relative to Uniform, SURFACE reduces the return-space coverage metric from $3.55$ to $2.50$ and IGD from $0.429$ to $0.421$, while increasing hypervolume from $507.7$ to $516.0$.

\end{document}